\documentclass[runningheads]{llncs}
\usepackage[mobile]{eccv}

\usepackage{eccvabbrv}

\usepackage{graphicx}
\usepackage{booktabs}
\usepackage{multirow}

\usepackage[accsupp]{axessibility}

\usepackage{hyperref}

\usepackage{orcidlink}

\begin{document}

\title{Anchor then Polish for Low-light Enhancement}

\titlerunning{ATP}

\author{
    Tianle Du\textsuperscript{$\dagger$} \and
    Mingjia Li\textsuperscript{$\dagger$} \and
    Hainuo Wang \and
    Xiaojie Guo\textsuperscript{*}
}

\authorrunning{Du et al.}

\institute{School of Software, Tianjin University, Tianjin, China\\
\email{\{dutianle, mingjiali, hainuo\}@tju.edu.cn, xj.max.guo@gmail.com \\
\url{https://tianledu.github.io/ATP/}}}

\maketitle
\insert\footins{\noindent\footnotesize\textsuperscript{$\dagger$}Equal Contribution.}
\insert\footins{\noindent\footnotesize\textsuperscript{*}Corresponding Author.}

\begin{figure}[tb]
  \centering
  \includegraphics[width=\linewidth,height=6.5cm,keepaspectratio]{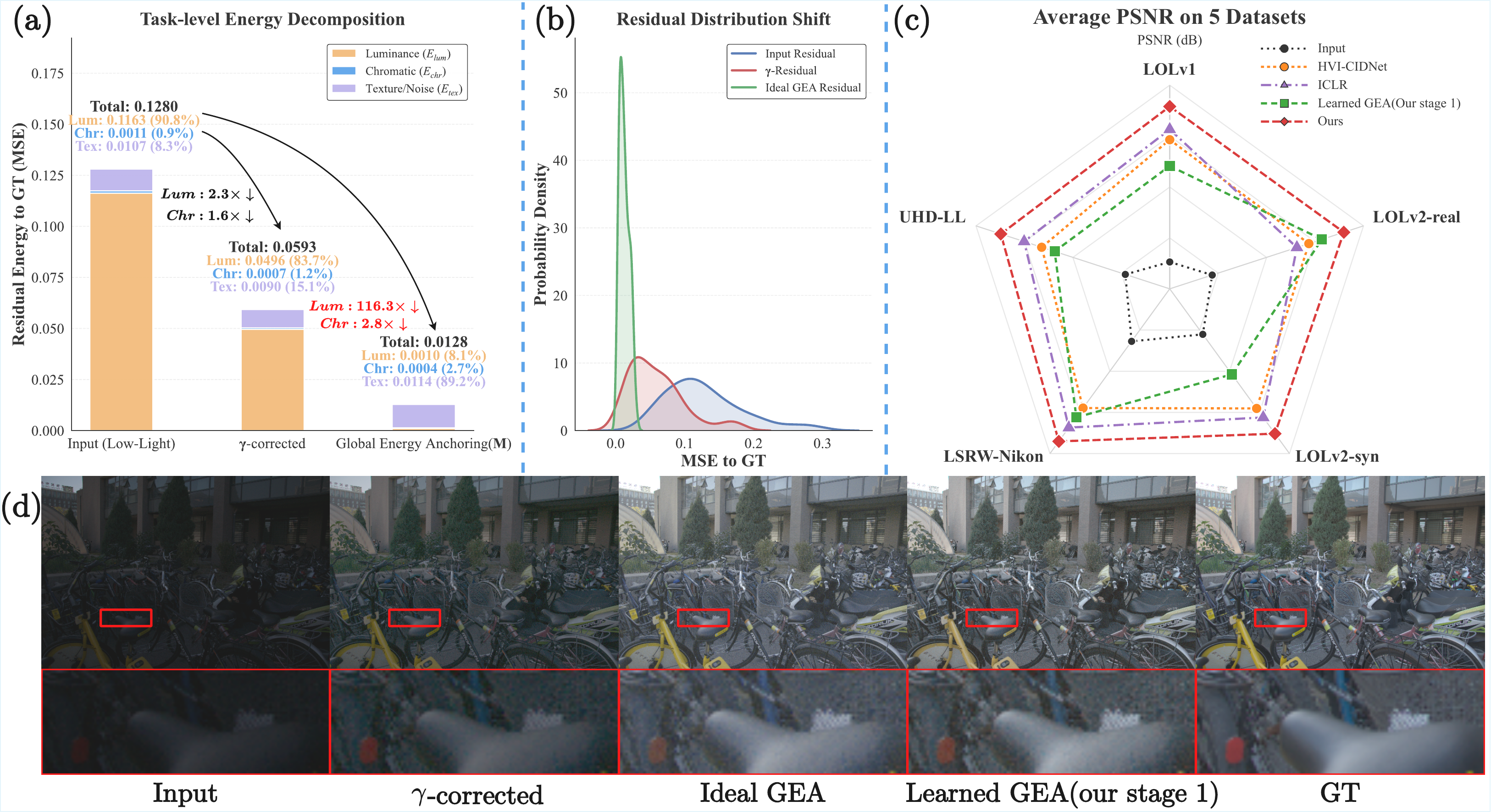}
  \caption{(a) Task-level energy decomposition of low-light degradations. (b) Residual distribution shift after global energy anchoring (GEA), illustrating error compression and re-centering. (c) PSNR comparison across five benchmarks. (d) Visual validation of different global energy alignment manners, where the Ideal GEA uses the optimal matrix specific to the given image. }
  \label{fig:motivation}
\end{figure}

\begin{abstract}
Low-light image enhancement is challenging due to entangled degradations, mainly including poor illumination, color shifts, and texture interference. Existing methods often rely on complex architectures to address these issues jointly but may overfit simple physical constraints, leading to global distortions. This work proposes a novel anchor-then-polish (ATP) framework to fundamentally decouple global energy alignment from local detail refinement. First, macro anchoring is customized to (greatly) stabilize luminance distribution and correct color by learning a scene-adaptive projection matrix with merely 12 degrees of freedom, revealing that a simple linear operator can effectively align global energy. The macro anchoring then reduces the task to micro polishing, which further refines details in the wavelet domain and chrominance space under matrix guidance. A constrained luminance update strategy is designed to ensure global consistency while directing the network to concentrate on fine-grained polishing. Extensive experiments on multiple benchmarks show that our method achieves state-of-the-art performance, producing visually natural and quantitatively superior low-light enhancements.

  \keywords{Low-light Image Enhancement \and Global Alignment \and Detail Refinement}
\end{abstract}

\section{Introduction}
\label{sec:intro}
Low-light image enhancement (LLIE) aims to recover visual clarity from images captured under suboptimal lighting conditions. The central challenge lies in simultaneously addressing coupled degradations, \emph{e.g.}, unsatisfactory illumination, color distortion, and texture contamination, which can be broadly categorized into global energy shifts and local detail artifacts. Global energy shifts distort luminance and chrominance distributions, whereas local perturbations primarily manifest as amplified noise and corrupted textures. Hence, achieving a principled balance between these two distinct aspects is key to producing high-quality enhancement results.

Recent deep learning approaches~\cite{guo2023low, jiang2021enlightengan, she2024mpc, liu2024ntire, yan2024causality, zhang2024dmfourllie, cai2023retinexformer} have significantly advanced LLIE. Most existing methods~\cite{jiang2021enlightengan,wang2022uformer, zamir2020learning} employ high-capacity networks~\cite{lecun2002gradient, vaswani2017attention} to learn end-to-end mappings that uniformly treat global energy shifts and local distortions. Despite their empirical success, such unified modeling overlooks that these two distinct objectives are inherently imbalanced in both physical characteristics and optimization behavior. Consequently, during joint optimization, networks tend to prioritize (low-frequency) global correction, as it dominates the loss. This induces an optimization imbalance where dominant gradients from low-frequency energy restoration suppress the learning signals necessary for fine-grained texture recovery and noise attenuation.

To better understand this phenomenon, we perform task-level energy decomposition of low-light degradations in terms of luminance, chrominance, and texture/noise, as illustrated in Fig.~\ref{fig:motivation}~(a). We observe a striking energy imbalance: over 90\% of the residual energy (measured by mean squared error, MSE) originates from global illumination shifts. Inspired by classical image signal processing (ISP) pipelines, we find that these dominant energy discrepancies can be effectively aligned using a simple $3 \times 4$ affine transformation matrix with only 12 degrees of freedom. Compared with conventional non-linear $\gamma$ correction, this linear alignment reduces luminance errors by 116.3$\times$, revealing that the PSNR degradation in LLIE primarily stems from linear global shifts rather than complex non-linear detail distortions.
As further evidenced by the residual distribution shift shown in \cref{fig:motivation}~(b), linear alignment effectively compresses the residual distribution and recenters it around zero, producing images whose luminance and color closely match the ground truth. The remaining discrepancies are predominantly noise-like and high-frequency in nature, as visually illustrated in \cref{fig:motivation}~(d). This observation provides crucial physical insight, say by pre-aligning global energy with a linear transformation operator, the enhancement task can be simplified as a constrained residual refinement problem. Such decoupling mitigates the optimization bias induced by dominant linear components and enables the network to devote its modeling capacity primarily to the recovery of non-linear details under the guidance of globally aligned energy.

Based on the above analysis, we propose an {anchor-then-polish} framework. ATP first employs a {global energy anchoring module} to predict a scene-adaptive projection matrix for coarse-grained global luminance and color correction. By explicitly modeling the dominant linear energy shift, the global energy anchoring provides a stable global prior and significantly simplifies the subsequent restoration task. After global anchoring, the aligned image is transformed into the YUV color space, where a {detail polishing module}  performs fine-grained texture reconstruction and noise suppression under the guidance of the predicted projection matrix. To maintain global energy consistency during local refinement, we further design a {constrained luminance update} strategy, which restricts undesired luminance drift and prevents low-frequency errors from dominating the optimization process, thereby encouraging the network to concentrate on high-frequency detail recovery within a decoupled representation.

The main contributions of this work are summarized as follows:
\begin{itemize}
    \item We uncover a pronounced energy imbalance in LLIE, showing that over 90\% of the residual MSE arises from linear global illumination shifts, which can be effectively corrected using a simple global anchoring operator with only 12 degrees of freedom. 
    \item We design a Global Energy Anchoring Module (GEAM) that explicitly captures the underlying linear physical prior of low-light degradation through scene-adaptive matrix estimation, providing a stable luminance and color foundation for subsequent refinement.
    \item We propose an Anchor-then-Polish (ATP) framework to decouple global energy alignment from local detail restoration. GEAM performs macro-scale global correction, while the Detail Polishing Module (DPM) together with the {Constrained Luminance Update} (CLU) strategy and the transformation prediction refines micro-scale textures.
\end{itemize}
Extensive ablation studies verify the rationality and effectiveness of the proposed design, while comprehensive experiments on multiple benchmarks show that ATP achieves state-of-the-art performance, delivering superior luminance consistency, accurate color reproduction, and high-fidelity texture restoration.

\section{Related Work}

\subsection{Unified Low-light Image Enhancement}
Unified methods formulate low-light enhancement as a direct image-to-image translation task, aiming to learn a holistic mapping from degraded input to normal-light image. Earlier methods primarily relied on global pixel-level transformations, such as Histogram Equalization (HE)~\cite{abdullah2007dynamic, celik2011contextual} and Gamma Correction (GC)~\cite{huang2012efficient, wang2009real}, to redistribute pixel intensities. With the advent of deep learning, this paradigm has evolved into more sophisticated end-to-end architectures~\cite{zamir2020learning, yan2025hvi, jiang2021enlightengan, moran2020deeplpf, zhou2022lednet}. Initial neural-based approaches like LLNet~\cite{lore2017llnet} employed stacked autoencoders to jointly handle denoising and brightening. More recently, advanced models such as SNR-Aware~\cite{xu2022snr} and LLFormer~\cite{wang2023ultra} have been developed to address poor illumination, noise, and color shifts within an integrated framework. Although these unified methods avoid the complexity of manual decomposition, they often struggle with entangled degradations, leading to global color distortions and artifacts in complex real-world scenes. Additionally, these methods typically employ complex Transformer-based~\cite{vaswani2017attention} architectures that require extensive attention computations, resulting in slow inference speeds and demanding costly computational resources.

\subsection{Decoupled Low-Light Image Enhancement}
Decoupled strategies provide a robust alternative by decomposing low-light images into distinct components. Most decoupled methods~\cite{li2025interpretable, bai2024retinexmamba, yi2023diff, wei2025illumflow, fu2023you, zhang2019kindling} follow the Retinex theory~\cite{land1977retinex} which decomposes an image into reflectance and illumination. For example, RetinexNet~\cite{wei2018deep} and Retinexformer~\cite{cai2023retinexformer} utilize specific sub-networks to estimate these components separately. Recently, frequency-based decomposition~\cite{wang2023fourllie, lv2024fourier} is also appealing for low-light restoration because illumination deviations are predominantly low-frequency, whereas textures and noise concentrate in high-frequency bands~\cite{wang2020high}. Wavelet-based approaches~\cite{zou2024wave, zou2022joint, jiang2023low, zhang2025cwnet, tan2024wavelet} introduce localized multi-scale sub-bands that enable more selective refinement. Although these methods enhance the different components by decoupling, the joint optimization process couples the errors together, preventing the potential of each subnetwork from being realized. Multi-stage decomposition networks~\cite{jiang2024lightendiffusion, liu2023low, zhang2021beyond} have been developed to address these limitations through progressive restoration. For instance, Bread~\cite{guo2023low} and Kind~\cite{zhang2019kindling} decouples luminance from chrominance and adopts a stage-wise enhancement strategy to alleviate color shift caused by brightening. However, these methods employ complex architectures to fit the virtually linear global energy, leading to overfitting and distortion of luminance. Furthermore, they ignore the uneven contribution of luminance and structure to the optimization objective, even a small luminance error can seriously affect the optimization direction of local details.

\section{ATP: Proposed Anchor then Polish Framework}

\begin{figure}[t]
  \centering
  \includegraphics[width=\linewidth,height=7.5cm,keepaspectratio]{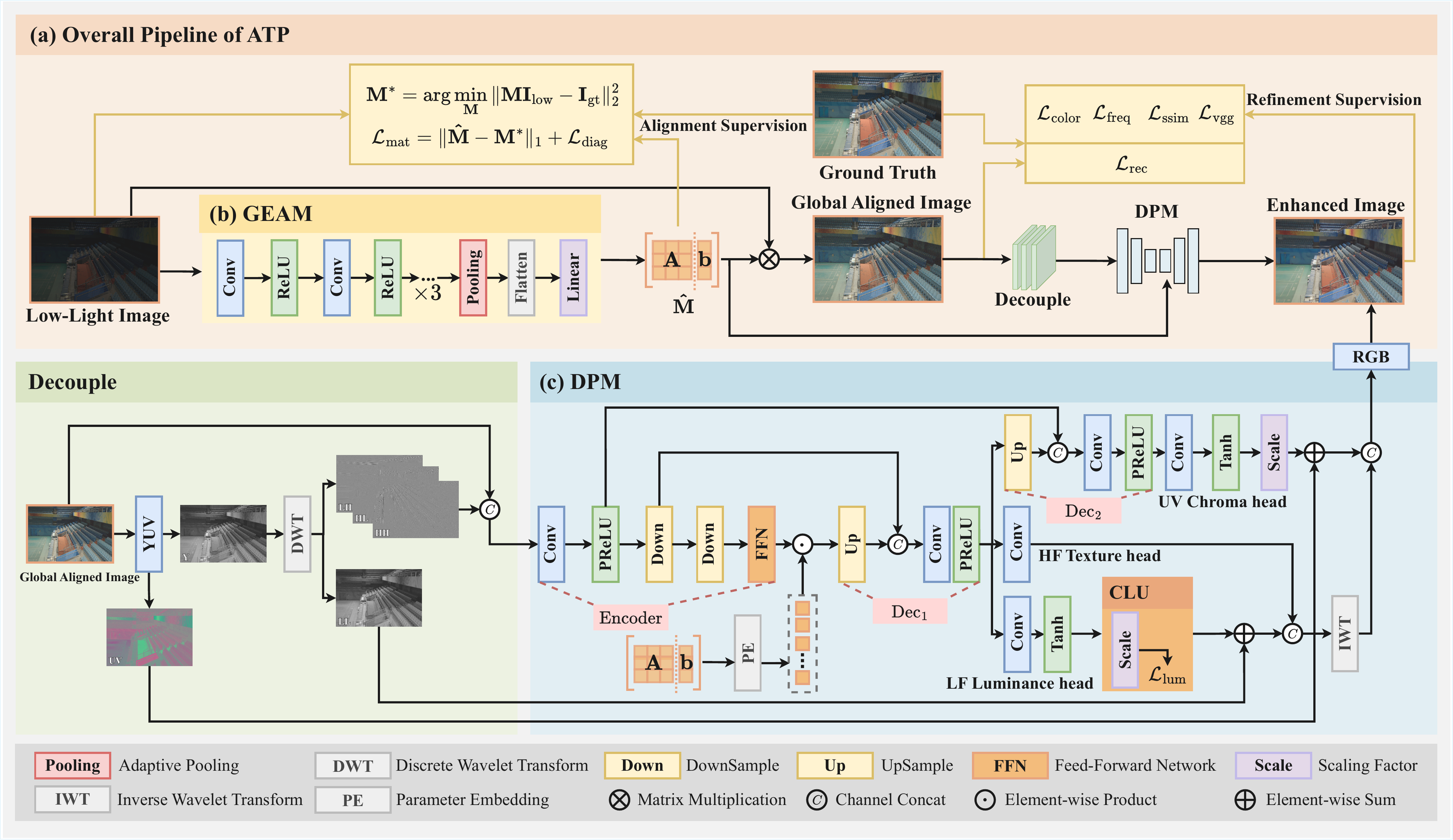}
  \caption{(a) Architecture of the proposed ATP framework. (b) Global Energy Anchoring Module (GEAM). (c) Detail Polishing Module (DPM).}
  \label{fig:ATP_network}
\end{figure}

As schematically depicted in \cref{fig:ATP_network}~(a), the proposed framework follows a macro-to-micro pipeline that explicitly decouples global energy alignment from local detail refinement. Specifically, given a low-light input $\mathbf{I}_{\text{low}}$, the Global Energy Anchoring Module (GEAM) first predicts a scene-adaptive energy anchoring matrix $\mathbf{M}$, which is supervised by an explicit optimal reference matrix. The estimated matrix is then applied pixel-wise to $\mathbf{I}_{\text{low}}$ through a linear transformation to obtain a globally aligned image $\mathbf{I}_\text{align}$, thereby anchoring/correcting the dominant linear energy shifts. Subsequently, $\mathbf{I}_\text{align}$ is decomposed into three complementary components, including low-frequency luminance, UV chrominance, and high-frequency texture. These decoupled representations are then fed into the Detail Polishing Module (DPM), which performs fine-grained refinement under the guidance of the global matrix to produce the final enhanced output $\mathbf{\hat{I}}$.

\subsection{Global Energy Anchoring Module}
 In conventional ISP pipelines, digital gain applies channel-wise scaling to compensate for under-exposure, while the color correction matrix (CCM) acts cross-channel mixing to calibrate white balance and sensor-specific color bias. These sequential stages can be formulated as a unified affine photometric model:
\begin{equation}
    \mathbf{I}_{\text{ISP}} = \mathbf{C}\,(\mathrm{Diag}(\mathbf{g})\, \mathbf{I}) + \mathbf{b},
    \label{eq:isp}
\end{equation}
where $\mathbf{g}\in\mathbb{R}^{3\times 1}=[g_r,g_g,g_b]^{\top}$ denotes per-channel gains, $\mathbf{C}\in\mathbb{R}^{3\times3}$ is the CCM, and $\mathbf{b}\in\mathbb{R}^{3\times 1}$ represents the black-level shift. Combining them yields a single affine matrix $\mathbf{M}\in\mathbb{R}^{3\times4} =[\mathbf{A}\,|\,\mathbf{b}]$ with $\mathbf{A}=\mathbf{C}\,\mathrm{Diag}(\mathbf{g})$, which characterizes global luminance amplification and color correction as a low-DoF linear operator.

Inspired by Eq.~\eqref{eq:isp}, GEAM directly regresses the scene-adaptive $\mathbf{\hat{M}}=[\mathbf{\hat{A}}\,|\,\mathbf{\hat{b}}]$. As illustrated in \cref{fig:ATP_network}~(b), GEAM adopts a straightforward architecture based on the premise that regressing the 12-parameter affine matrix $\mathbf{\hat{M}}$ does not require a complex backbone. Since the $\mathbf{\hat{M}}$ represents a global energy mapping, the network is encouraged to prioritize large-scale semantic context while suppressing interference from fine-grained local textures. To this end, GEAM employs five $3\times3$ convolutional blocks with progressive downsampling to extract hierarchical features with expanding receptive fields. An adaptive average pooling layer aggregates these features into a compact global descriptor, effectively filtering out local spatial variations. Finally, a linear layer regresses the 12 elements of $\mathbf{\hat{M}}$. Then the globally aligned image $\mathbf{I}_\text{align}$ is computed via:
\begin{equation}
    \mathbf{I}_\text{align} = \mathbf{\hat{A}} \mathbf{I}_\text{low} + \mathbf{\hat{b}}.
\end{equation}
This explicit affine transform jointly enhances luminance, corrects white balance, and aligns cross-channel color responses, providing a stable and physically interpretable global photometric anchor for subsequent non-linear refinement.

\subsection{Detail Polishing Module}
\label{sec:dpm}

To better decouple and refine individual components, the aligned image $\mathbf{I}_{\text{align}}$ is first converted into the YUV space to decouple luminance from chrominance. Subsequently, a wavelet transform is applied to the luminance channel $\mathbf{Y}$ to further decouple low-frequency luminance $\mathbf{Y}_{\text{LL}}$ from high-frequency textures $\mathcal{Y}_{\text{HF}}$ via the decomposition $\{\mathbf{Y}_{\text{LL}}, \mathbf{Y}_{\text{LH}}, \mathbf{Y}_{\text{HL}}, \mathbf{Y}_{\text{HH}}\} = \text{DWT}(\mathbf{Y})$, where $\mathcal{Y}_{\text{HF}} = \{\mathbf{Y}_{\text{LH}}, \mathbf{Y}_{\text{HL}}, \mathbf{Y}_{\text{HH}}\}$. Finally, as shown in~\cref{fig:ATP_network}~(c), these decoupled components are fed into Detail Polishing Module~(DPM). This module employs an encoder-decoder U-Net architecture to perform specifically for luminance, color and texture, yielding the new representations $\{\hat{\mathbf{Y}}, \hat{\mathbf{U}}, \hat{\mathbf{V}}\}$ in the YUV space, which are then converted back to the RGB space to obtain the final result $\mathbf{\hat{I}}$.

Since the GEA operation is applied globally to all pixels, the strength of local restoration should be inherently coupled with the magnitude of global photometric transformation. Intuitively, scenes requiring substantial global energy correction often suffer from heavier noise amplification or texture degradation. To reflect this dependency, we incorporate the global transformation matrix $\mathbf{\hat{M}}$ predicted by GEAM as a conditional prior to guide the detail refinement, yielding the modulated representation as:
\begin{equation}
    \mathbf{F}_{\text{mod}} = \text{Enc}([\mathbf{I}_{\text{align}};\mathcal{Y}_{\text{HF}}]) \odot \text{PE}(\mathbf{\hat{M}}),
\end{equation}
where $[\cdot;\cdot]$ denotes channel-wise concatenation and $\text{PE}(\cdot)$ executes parameter embedding operation. This mechanism enables the DPM to dynamically adjust its restoration behavior according to the scene-level photometric state. When $\mathbf{\hat{M}}$ deviates significantly from the identity matrix (\ie, the scene takes substantial energy shift), the modulation emphasizes denoising-oriented features. Conversely, when only mild global adjustment is required, the network focuses more on fine texture preservation.

To preserve the photometric stability established by GEAM, we introduce the Constrained Luminance Update (CLU) strategy to restrict unnecessary modifications to $\mathbf{Y}_{\text{LL}}$. Architecturally, we use scaling factors and tanh activation to limit the low-frequency residual components. Optimization-wise, we use luminance preservation loss to make the low-frequency luminance residual as close to zero as possible:
\begin{equation}
    \mathbf{\hat{Y}_{\text{LL}}}=\mathbf{Y}_{\text{LL}}+\gamma_{\text{L}} \cdot \tanh\big(\mathcal{H}_{\text{LL}}(\text{Dec}_{\text{1}}(\mathbf{F}_{\text{mod}}))\big),
    \mathcal{L}_{\text{lum}}=\| \mathbf{\hat{Y}_{\text{LL}}}-\mathbf{{Y}_{\text{LL}}} \|_{1},
\end{equation}
where $\mathcal{H}_{\text{LL}}$ denotes the low-frequency luminance prediction head. This strategy effectively steers the DPM to prioritize high-frequency denoising rather than redundant global illumination adjustments.

\subsection{Learning Objective}
\label{sec:loss}

Mirroring the architecture, the learning objectives for GEAM and DPM are alignment supervision and decoupled refinement supervision, respectively.

\noindent\textbf{Alignment Supervision.}
GEAM is optimized via a reconstruction loss $\mathcal{L}_{\text{rec}} = \|\hat{\mathbf{I}}_{\text{align}} - \mathbf{I}_{\text{gt}}\|_1$ alongside a matrix supervision loss that provides explicit guidance. During the training process, the optimal matrix $\mathbf{M}^{*}$ is computed via least-squares fitting between the input $\mathbf{I}_{\text{low}}$ and ground truth $\mathbf{I}_{\text{gt}}$:
\begin{equation}
    \mathbf{M}^{*} = \arg\min_{\mathbf{M}} \| \mathbf{M}\mathbf{I}_{\text{low}} - \mathbf{I}_{\text{gt}} \|_2^2 = \mathbf{I}_{\text{gt}}\mathbf{I}_{\text{low}}^{\top}(\mathbf{I}_{\text{low}}\mathbf{I}_{\text{low}}^{\top})^{-1}.
\end{equation}
The optimal matrix supervision $\mathcal{L}_{\text{mat}}$ penalizes the deviation of the predicted $\hat{\mathbf{M}}$ from $\mathbf{M}^{*}$ while regularizing the $3 \times 3$ sub-matrix $\hat{\mathbf{A}}$ toward identity matrix $\mathbf{I}_\text{iden}$ to prevent degeneracy and stabilize early training:
\begin{equation}
    \mathcal{L}_{\text{mat}} = \|\hat{\mathbf{M}} - \mathbf{M}^{*}\|_1 + \lambda_{\text{diag}}\|\hat{\mathbf{A}} - \mathbf{I}_\text{iden}\|_\text{F}^2.
\end{equation}
In summary, the total alignment objective is $\mathcal{L}_{\text{align}} = \lambda_{\text{rec}}\mathcal{L}_{\text{rec}} + \lambda_{\text{mat}}\mathcal{L}_{\text{mat}}$, with $[\lambda_{\text{rec}}, \lambda_{\text{mat}}, \lambda_{\text{diag}}] = [1.0, 1.0, 0.1]$. And the joint effect of $\mathcal{L}_{\text{rec}}$ and $\mathcal{L}_{\text{mat}}$ mitigates overfitting, because supervision with $\mathbf{M}^{*}$ alone may lead the network to memorize training-set matrices, while $\mathcal{L}_{\text{rec}}$ enforces image-level consistency and encourages the network to explore alignment matrices that surpass the performance of the least-squares baseline.

\noindent\textbf{Refinement Supervision.}

With GEAM frozen, to supervise the learning of distinct image components, the objective for DPM is formulated as follows:
\begin{equation}
    \mathcal{L}_{\text{dec}} = \underbrace{\lambda_{\text{freq}}\mathcal{L}_{\text{freq}} + \lambda_{\text{color}}\mathcal{L}_{\text{color}} + \lambda_{\text{lum}}\mathcal{L}_{\text{lum}}}_{\text{Decoupled Fidelity}} + \underbrace{\lambda_{\text{ssim}}\mathcal{L}_{\text{ssim}} + \lambda_{\text{vgg}}\mathcal{L}_{\text{vgg}}}_{\text{Perceptual Quality}}.
\end{equation}
The decoupled fidelity group enforces restoration in the decomposed representation. The wavelet frequency loss is computed exclusively on the high-frequency sub-bands of the luminance channel to preserve texture fidelity:
\begin{equation}
    \mathcal{L}_{\text{freq}} = \|\text{DWT}(\mathbf{\hat{Y}})_{\text{HF}} - \text{DWT}(\mathbf{Y}_{\text{gt}})_{\text{HF}}\|_1,
\end{equation}
where $\text{DWT}(\cdot)_{\text{HF}}$ extracts the three high-frequency sub-bands. The chrominance loss constrains color fidelity in the YUV space as follows:
\begin{equation}
    \mathcal{L}_{\text{color}} = \|\mathbf{\hat{U}} - \mathbf{U}_{\text{gt}}\|_1+ \|\mathbf{\hat{V}} - \mathbf{V}_{\text{gt}}\|_1.
\end{equation}
Setting $[\lambda_{\text{freq}},\allowbreak \lambda_{\text{color}},\allowbreak \lambda_{\text{lum}},\allowbreak \lambda_{\text{ssim}},\allowbreak \lambda_{\text{vgg}}]
= [1.0,\allowbreak 1.0,\allowbreak 0.01,\allowbreak 0.5,\allowbreak 0.01]$ performs empirically well, better results could be obtained by further tuning these weights.

\begin{table*}[t]
\centering
\caption{Quantitative comparison on LOLv1\cite{wei2018deep} and LOLv2\cite{yang2021sparse} datasets. Best results are in \textbf{bold} and second best are \underline{underlined}. All results are reported using the checkpoint that achieves the highest PSNR without using the GT mean strategy.}
\label{tab:lol_with_params}
\setlength{\tabcolsep}{3.5pt}
\renewcommand{\arraystretch}{1.15}
\footnotesize
\resizebox{\textwidth}{!}{%
\begin{tabular}{c c c ccc ccc ccc}
\toprule
\multirow{2}{*}{Method} & 
\multirow{2}{*}{\begin{tabular}[c]{@{}c@{}}FPS$\uparrow$\\ (600$\times$400)\end{tabular}} & 
\multirow{2}{*}{\begin{tabular}[c]{@{}c@{}}Params$\downarrow$\\ (M)\end{tabular}} &
\multicolumn{3}{c}{LOLv1} &
\multicolumn{3}{c}{LOLv2-Real} &
\multicolumn{3}{c}{LOLv2-Synthetic} \\
\cmidrule(lr){4-6}\cmidrule(lr){7-9}\cmidrule(lr){10-12}
& & &
PSNR$\uparrow$ & SSIM$\uparrow$ & LPIPS$\downarrow$ &
PSNR$\uparrow$ & SSIM$\uparrow$ & LPIPS$\downarrow$ &
PSNR$\uparrow$ & SSIM$\uparrow$ & LPIPS$\downarrow$ \\
\midrule
KinD\cite{zhang2019kindling}              &  2.75  & 8.02 & 19.427 & 0.7135 & 0.2391 & 17.544 & 0.6690 & 0.3750 & 17.137 & 0.7620 & 0.2550 \\
ZeroDCE\cite{guo2020zero}           & 629.79 & 0.075 & 14.861 & 0.5588 & 0.3351 & 18.059 & 0.5736 & 0.3125 & 17.756 & 0.8163 & 0.1681 \\
MIRNet\cite{zamir2020learning}            & 6.50 & 31.79 & 20.566 & 0.7722 & 0.2542 & 21.281 & 0.7920 & 0.3538 & 21.741 & 0.8780 & 0.1383 \\
RUAS\cite{liu2021retinex}              & 159.45 & 0.003 & 16.405 & 0.4996 & 0.2701 & 15.326 & 0.4876 & 0.3097 & 13.402 & 0.6444 & 0.3640 \\
SNR-Aware\cite{xu2022snr}         & 89.74 & 4.01 & 23.005 & 0.8240 & 0.1640 & 21.103 & 0.8390 & 0.1690 & 24.173 & 0.9240 & 0.0640 \\
Restormer\cite{zamir2022restormer}         & 9.55 & 25.31 & 23.231 & 0.8257 & 0.1434 & 21.109 & 0.8341 & 0.1681 & 25.318 & 0.9310 & 0.0560 \\
UFormer\cite{wang2022uformer}           & 8.12 & 50.88 & 18.851 & 0.7508 & 0.2879 & 19.495 & 0.6583 & 0.3743 & 21.501 & 0.8839 & 0.1197 \\
Retformer\cite{cai2023retinexformer}    & 28.45 & 1.53 & 23.399 & 0.8215 & 0.1522 & 21.693 & 0.8461 & 0.1573 & 25.480 & 0.9298 & 0.0605 \\
FourLLIE\cite{wang2023fourllie}          & 55.96   & 0.12 & 20.211 & 0.7785 & 0.2223 & 22.347 & 0.8472 & \underline{0.1141} & 24.643 & 0.9205 & 0.0656 \\
LDiff\cite{jiang2024lightendiffusion} & 1.49 & 27.83 & 19.990 & 0.8131 & 0.1800 & 22.835 & 0.8546 & 0.1670 & 21.804 & 0.8688 & 0.1545 \\
CIDNet\cite{yan2025hvi}            & 25.20 & 1.88 & 23.973 & 0.8490 & 0.1039 & 23.904 & \underline{0.8656} & 0.1219 & 25.269 & 0.9354 & 0.0481 \\
ICLR\cite{xu2025iclr}              & 14.75 & 4.27 & \underline{24.063} & \underline{0.8618} & \underline{0.0840} & 23.669 & 0.8631 & 0.1300 & \underline{25.658} & \underline{0.9389} & \textbf{0.0425} \\
\midrule
\textbf{Ours-GEAM}  & 532.65 & 1.25 & 22.147 & 0.6986 & 0.3024 & \underline{24.001} & 0.7279 & 0.2490 & 22.004 & 0.8761 & 0.1348 \\
\textbf{Ours-Full}    & 127.18 & 1.83 &
\textbf{24.946} & \textbf{0.8710} & \textbf{0.0769} &
\textbf{25.140} & \textbf{0.8851} & \textbf{0.0959} &
\textbf{26.301} & \textbf{0.9398} & \underline{0.0478} \\
\bottomrule
\end{tabular}%
}
\end{table*}

\begin{figure*}[tb]
  \centering
  \setlength{\tabcolsep}{0.5pt}
  \renewcommand{\arraystretch}{0.3}

  \begin{tabular}{ccccccc}

    \includegraphics[width=0.14\linewidth]{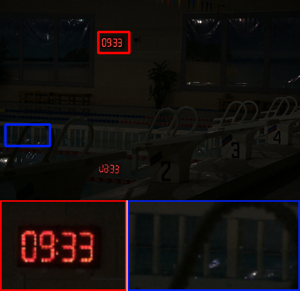} &
    \includegraphics[width=0.14\linewidth]{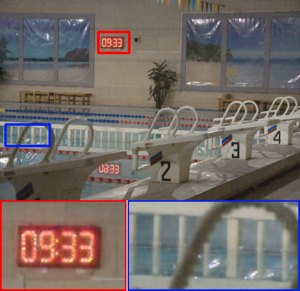} &
    \includegraphics[width=0.14\linewidth]{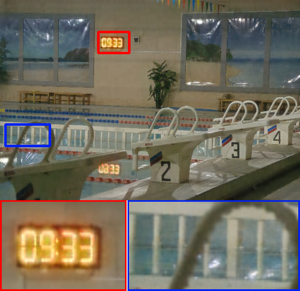} &
    \includegraphics[width=0.14\linewidth]{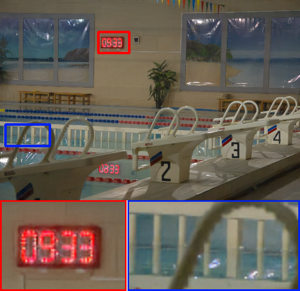} &
    \includegraphics[width=0.14\linewidth]{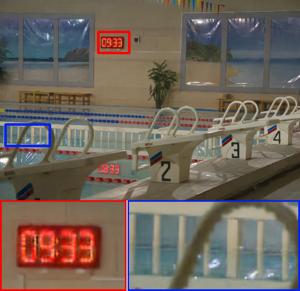} &
    \includegraphics[width=0.14\linewidth]{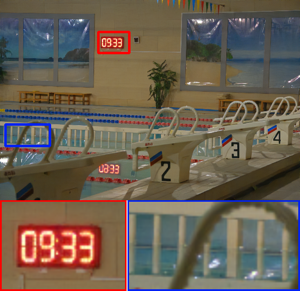} &
    \includegraphics[width=0.14\linewidth]{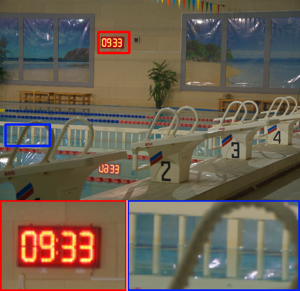} \\

    \includegraphics[width=0.14\linewidth]{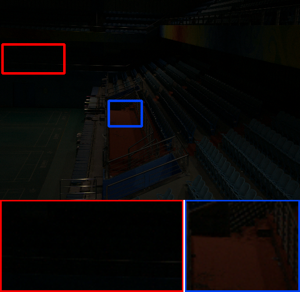} &
    \includegraphics[width=0.14\linewidth]{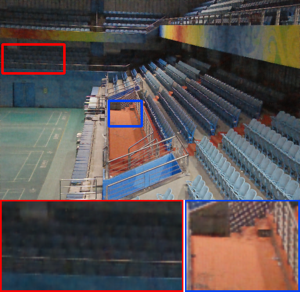} &
    \includegraphics[width=0.14\linewidth]{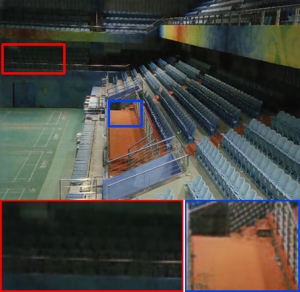} &
    \includegraphics[width=0.14\linewidth]{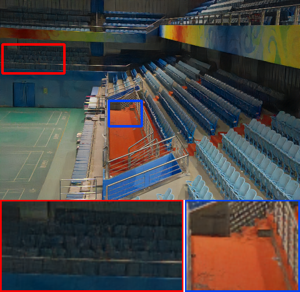} &
    \includegraphics[width=0.14\linewidth]{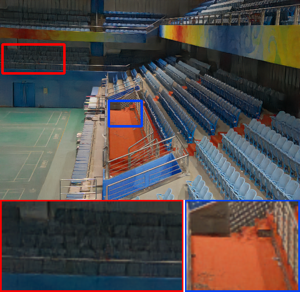} &
    \includegraphics[width=0.14\linewidth]{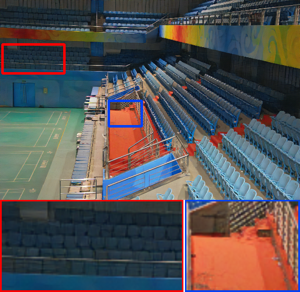} &
    \includegraphics[width=0.14\linewidth]{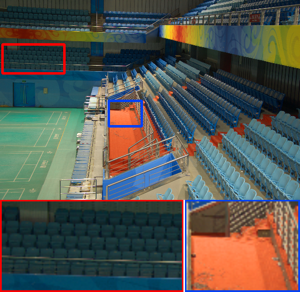} \\

    \includegraphics[width=0.14\linewidth]{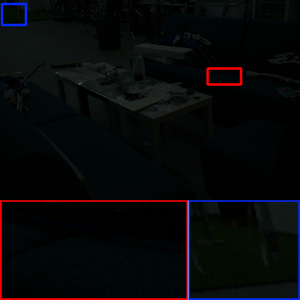} &
    \includegraphics[width=0.14\linewidth]{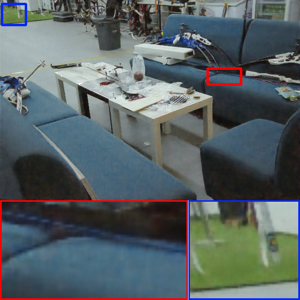} &
    \includegraphics[width=0.14\linewidth]{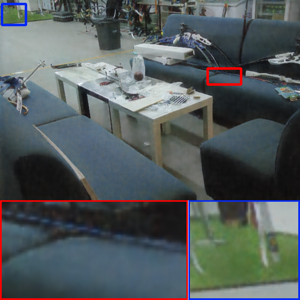} &
    \includegraphics[width=0.14\linewidth]{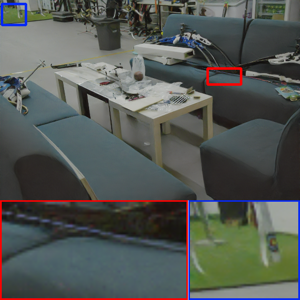} &
    \includegraphics[width=0.14\linewidth]{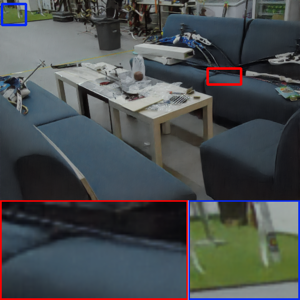} &
    \includegraphics[width=0.14\linewidth]{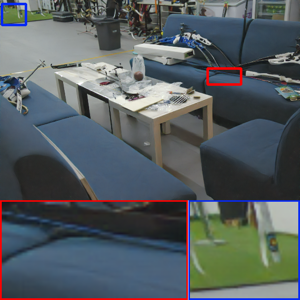} &
    \includegraphics[width=0.14\linewidth]{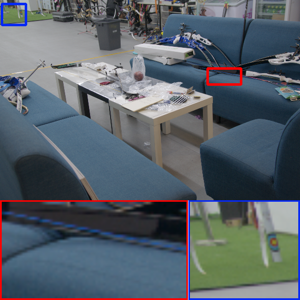} \\

    \includegraphics[width=0.14\linewidth]{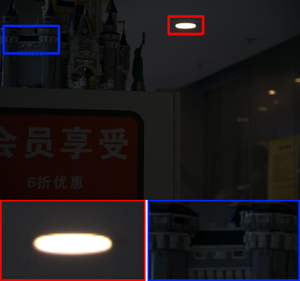} &
    \includegraphics[width=0.14\linewidth]{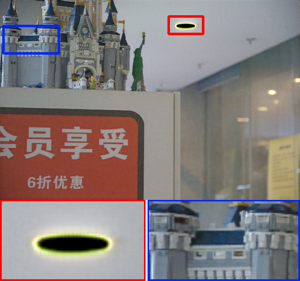} &
    \includegraphics[width=0.14\linewidth]{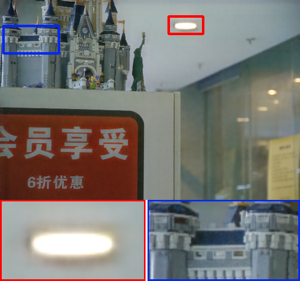} &
    \includegraphics[width=0.14\linewidth]{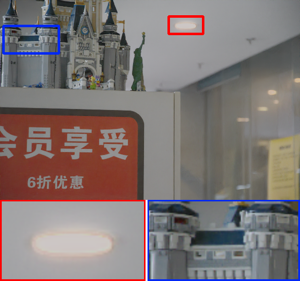} &
    \includegraphics[width=0.14\linewidth]{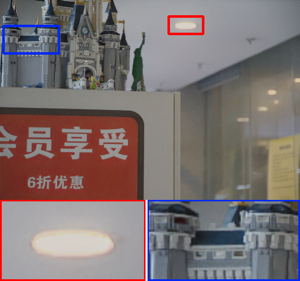} &
    \includegraphics[width=0.14\linewidth]{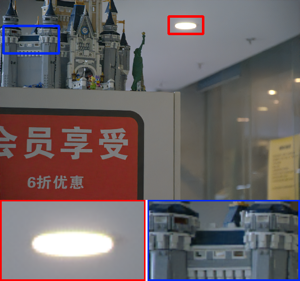} &
    \includegraphics[width=0.14\linewidth]{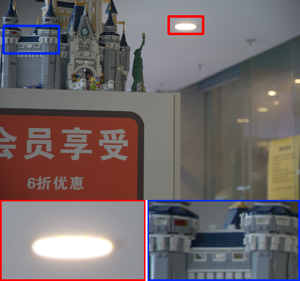} \\

    \includegraphics[width=0.14\linewidth]{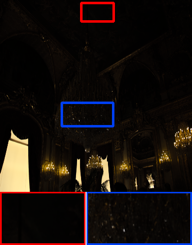} &
    \includegraphics[width=0.14\linewidth]{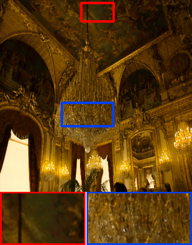} &
    \includegraphics[width=0.14\linewidth]{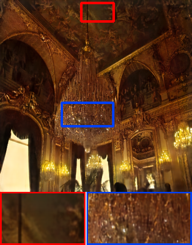} &
    \includegraphics[width=0.14\linewidth]{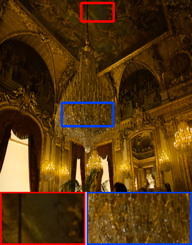} &
    \includegraphics[width=0.14\linewidth]{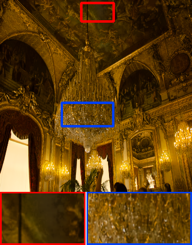} &
    \includegraphics[width=0.14\linewidth]{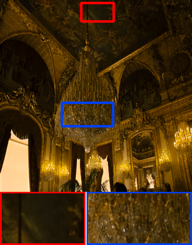} &
    \includegraphics[width=0.14\linewidth]{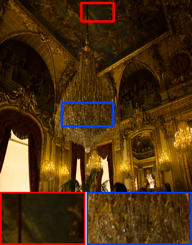} \\

    \\[-1pt]
    {\scriptsize{Input}} &
    {\scriptsize{Retformer}\cite{cai2023retinexformer}} &
    {\scriptsize{LDiff}\cite{jiang2024lightendiffusion}} &
    {\scriptsize{CIDNet}\cite{yan2025hvi}} &
    {\scriptsize{ICLR}\cite{xu2025iclr}} &
    {\scriptsize{Ours}} &
    {\scriptsize{GT}} \\
  \end{tabular}

  \caption{Visual comparisons of the enhanced results by different methods on LOLv1~\cite{wei2018deep}, LOLv2-Real~\cite{yang2021sparse} and LOLv2-Synthetic~\cite{yang2021sparse}. Zoom in for the best view.}
  \label{fig:lol_visual}
\end{figure*}

\section{Experimental Validation}
\subsection{Datasets and Metrics}

\noindent\textbf{Paired datasets.}
For paired evaluation, we employ five benchmarks spanning diverse capture conditions and resolutions. LOLv1~\cite{wei2018deep} comprises 485 training and 15 testing low/normal-light image pairs from real scenes. Its successor, LOLv2~\cite{yang2021sparse}, provides two subsets: LOLv2-Real with 689 training and 100 testing pairs from real captures, and LOLv2-Synthetic with 900 training and 100 testing pairs generated through synthetic degradation pipelines. LSRW-Nikon~\cite{hai2023r2rnet} offers 3,150 training and 20 testing real-world pairs captured with Nikon cameras under diverse indoor and outdoor lighting. UHD-LL~\cite{li2023embedding} is a large-scale 4K-resolution benchmark of 2,000 training and 150 testing pairs, designed to evaluate enhancement methods in ultra-high-definition settings.

\noindent\textbf{Unpaired datasets.}
For generalization evaluation under no-reference settings, we adopt five widely used benchmarks: DICM~\cite{lee2013contrast}, LIME~\cite{guo2016lime}, MEF~\cite{ma2015perceptual}, NPE~\cite{wang2013naturalness}, and VV~\cite{vonikakis2018evaluation}, which collectively cover diverse real-world low-light scenarios without ground-truth references.

\begin{table*}[t]
\centering
\caption{Quantitative comparison on LSRW-Nikon\cite{hai2023r2rnet} and UHD-LL\cite{li2023embedding} datasets. Best results are in \textbf{bold} and second best are \underline{underlined}.}
\label{tab:lsrw_uhdll_with_params}
\setlength{\tabcolsep}{4pt}
\renewcommand{\arraystretch}{1.15}
\footnotesize
\resizebox{\textwidth}{!}{%
\begin{tabular}{c c c ccc ccc ccc}
\toprule
\multirow{2}{*}{Method} & 
\multirow{2}{*}{\begin{tabular}[c]{@{}c@{}}FPS$\uparrow$\\ (600$\times$400)\end{tabular}} & 
\multirow{2}{*}{\begin{tabular}[c]{@{}c@{}}Params$\downarrow$\\ (M)\end{tabular}} &
\multicolumn{3}{c}{LSRW-Nikon} &
\multicolumn{3}{c}{LSRW-Nikon-Reg} &
\multicolumn{3}{c}{UHD-LL} \\
\cmidrule(lr){4-6}\cmidrule(lr){7-9}\cmidrule(lr){10-12}
& & &
PSNR$\uparrow$ & SSIM$\uparrow$ & LPIPS$\downarrow$ &
PSNR$\uparrow$ & SSIM$\uparrow$ & LPIPS$\downarrow$ &
PSNR$\uparrow$ & SSIM$\uparrow$ & LPIPS$\downarrow$ \\
\midrule
ZeroDCE\cite{guo2020zero}        & 629.79 & 0.075 & 15.028 & 0.4183 & 0.2438 & 17.236 & 0.6966 & 0.2104 & 17.071 & 0.6391 & 0.5136 \\
MIRNet\cite{zamir2020learning}         & 6.50 & 31.79 & 17.377 & 0.5015 & 0.5388 & 18.966 & 0.6105 & 0.5271 & 20.550 & 0.8067 & 0.4351 \\
UFormer\cite{wang2022uformer}        & 8.12 & 50.88 & 17.086 & 0.5003 & 0.4627 & 18.788 & 0.6165 & 0.4680 & 19.306 & 0.8053 & 0.4360 \\
Retformer\cite{cai2023retinexformer}  & 28.45  & 1.53 & 17.303 & 0.5063 & 0.4897 & 19.716 & 0.6201 & 0.4734 & 22.217 & 0.8350 & 0.2815 \\
CIDNet\cite{yan2025hvi}         & 25.20  & 1.88 & 17.148 & 0.5038 & 0.2984 & 19.444 & 0.6599 & 0.3062 & 23.799 & 0.8746 & 0.2094 \\
ICLR\cite{xu2025iclr}           & 14.75 & 4.27 & \underline{17.487} & \underline{0.5067} & 0.2747 & 19.525 & 0.6700 & 0.2833 & \underline{24.229} & \underline{0.8844} & \underline{0.1984} \\
\midrule
\textbf{Ours-GEAM}     & 532.65 & 1.25 & 17.244 & 0.4634 & \textbf{0.2244} & \underline{20.548} & \textbf{0.7161} & \textbf{0.1944} & 22.975 & 0.6644 & 0.4690 \\
\textbf{Ours-Full}  & 127.18 & 1.83 &
\textbf{17.824} & \textbf{0.5085} & \underline{0.2382} &
\textbf{20.778} & \underline{0.7090} & \underline{0.2386} &
\textbf{24.895} & \textbf{0.8913} & \textbf{0.1911} \\
\bottomrule
\end{tabular}%
}
\end{table*}

\begin{figure}[tb]
  \centering

  \setlength{\tabcolsep}{0.5pt}
  \renewcommand{\arraystretch}{0.3}

  \begin{tabular}{ccccccc}

    \includegraphics[width=0.14\linewidth]{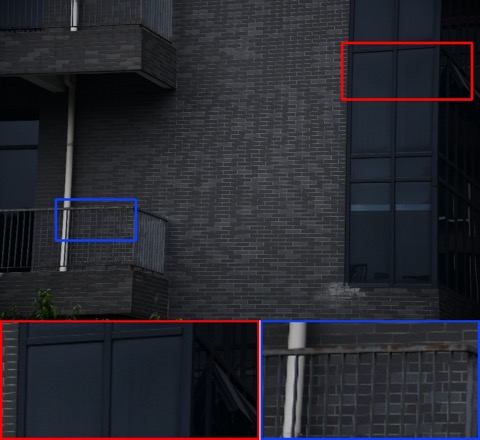} &
    \includegraphics[width=0.14\linewidth]{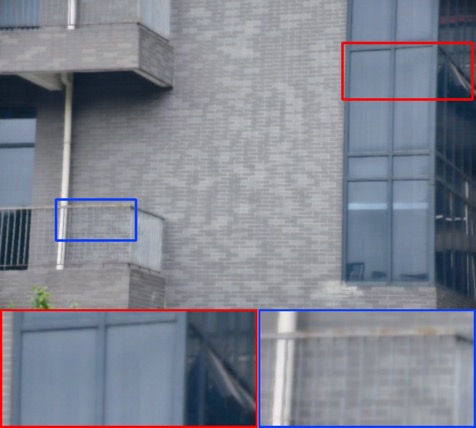} &
    \includegraphics[width=0.14\linewidth]{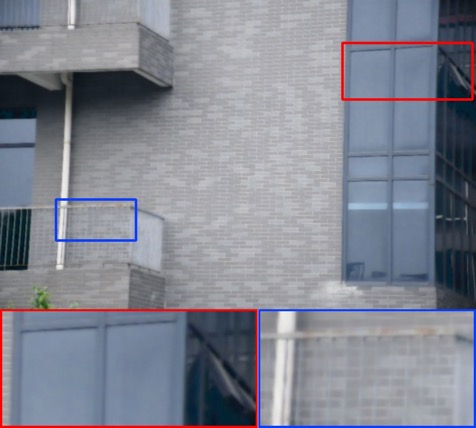} &
    \includegraphics[width=0.14\linewidth]{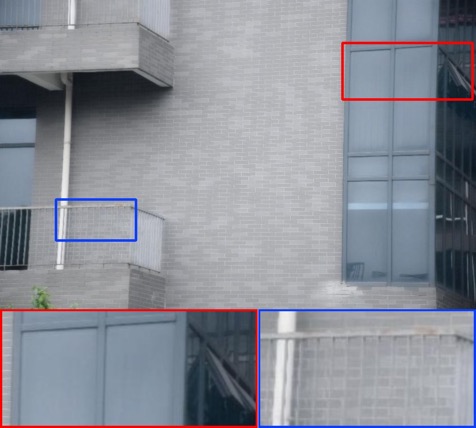} &
    \includegraphics[width=0.14\linewidth]{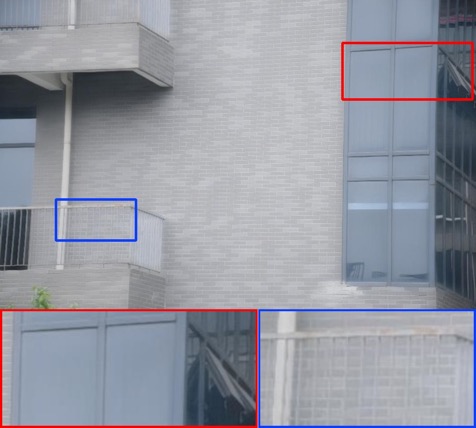} &
    \includegraphics[width=0.14\linewidth]{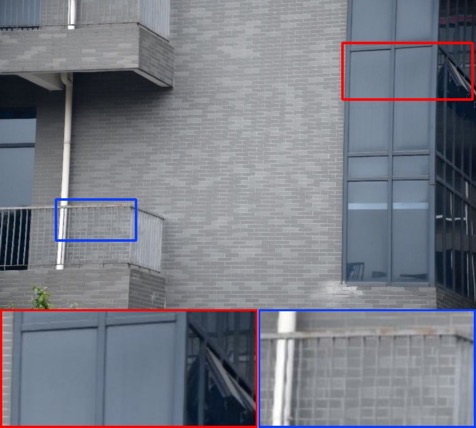} &
    \includegraphics[width=0.14\linewidth]{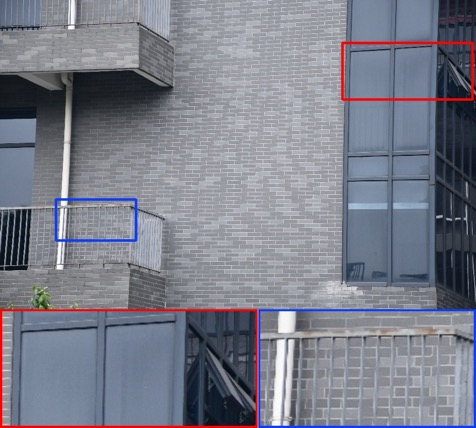} \\

    \includegraphics[width=0.14\linewidth]{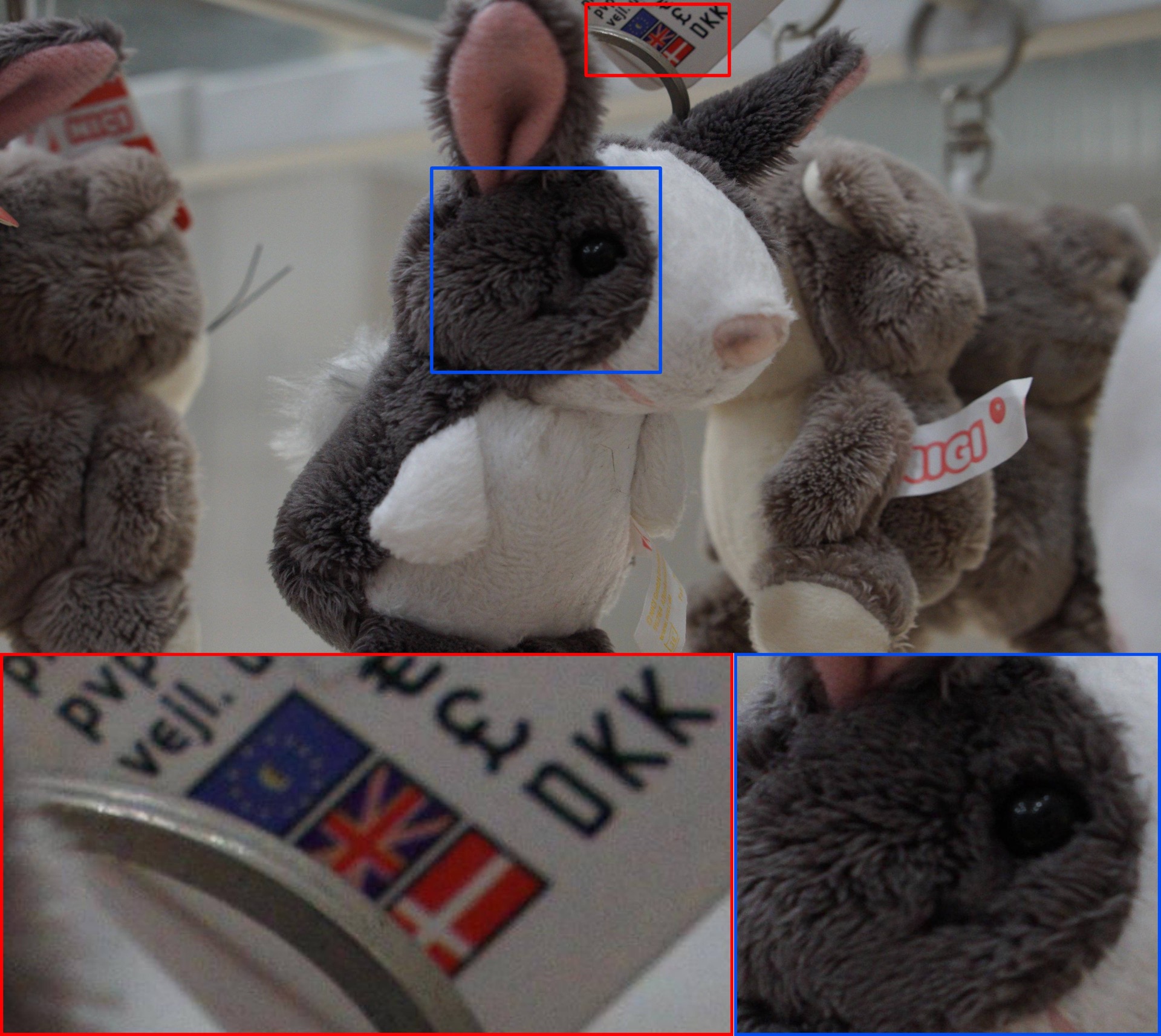} &
    \includegraphics[width=0.14\linewidth]{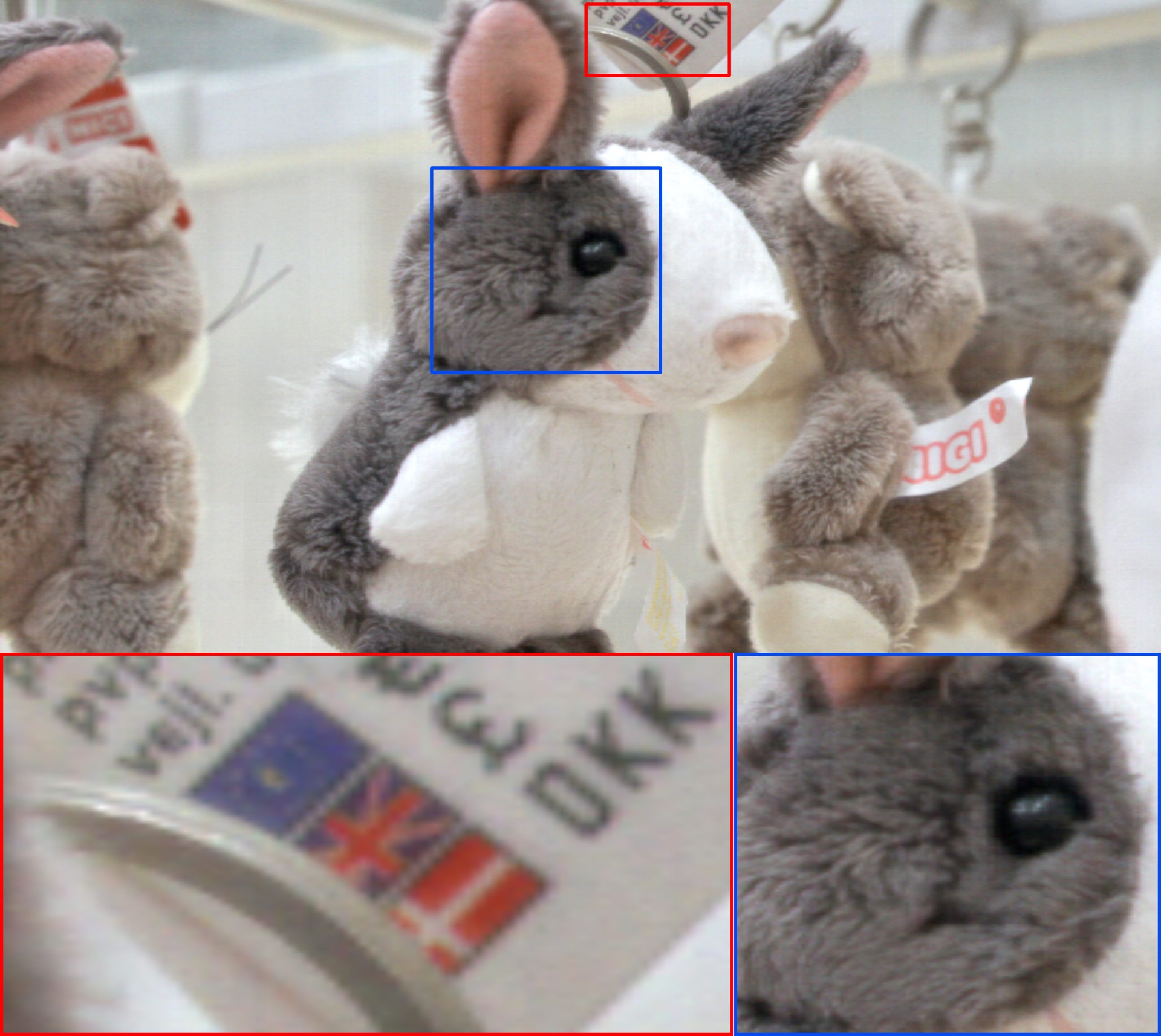} &
    \includegraphics[width=0.14\linewidth]{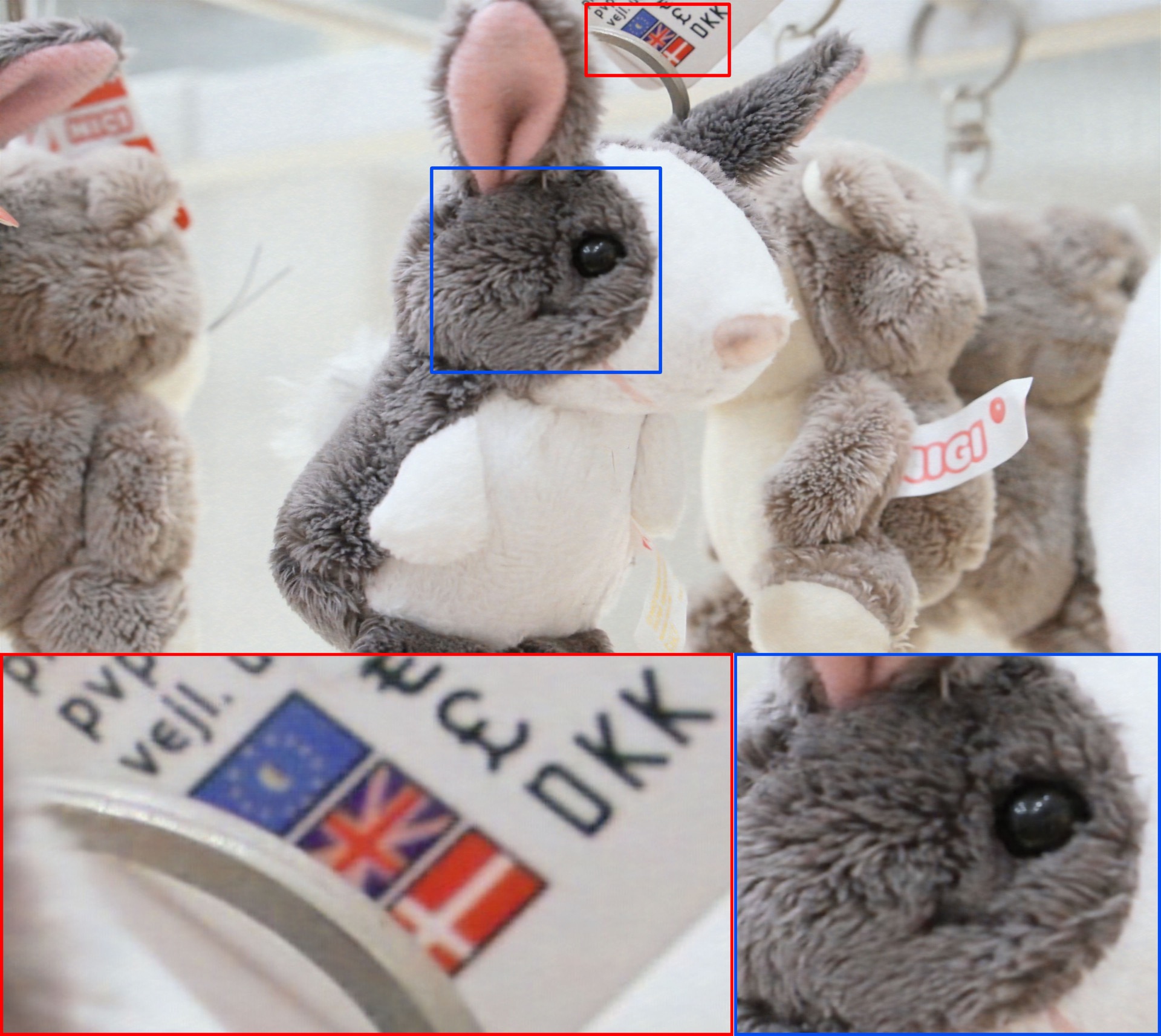} &
    \includegraphics[width=0.14\linewidth]{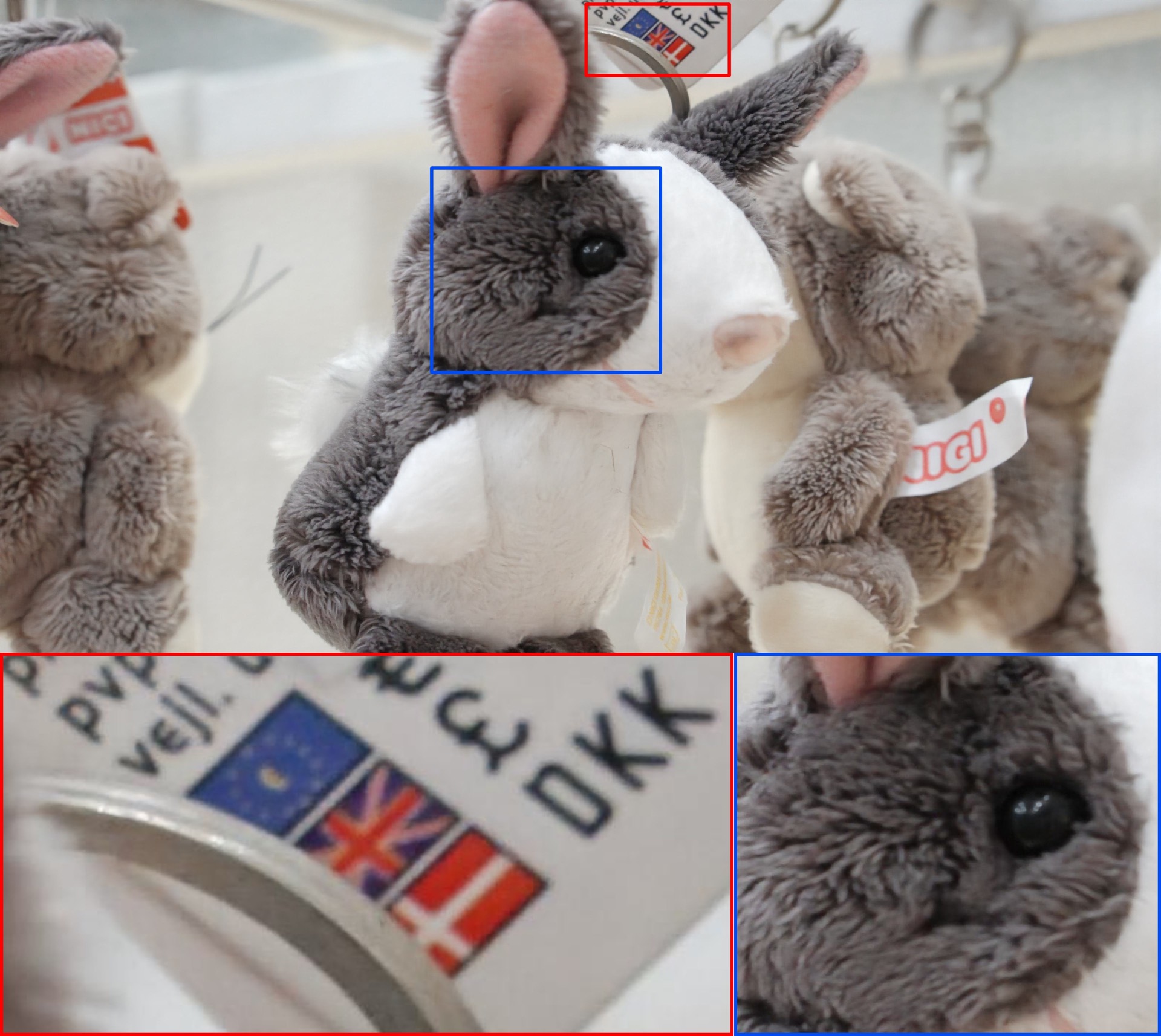} &
    \includegraphics[width=0.14\linewidth]{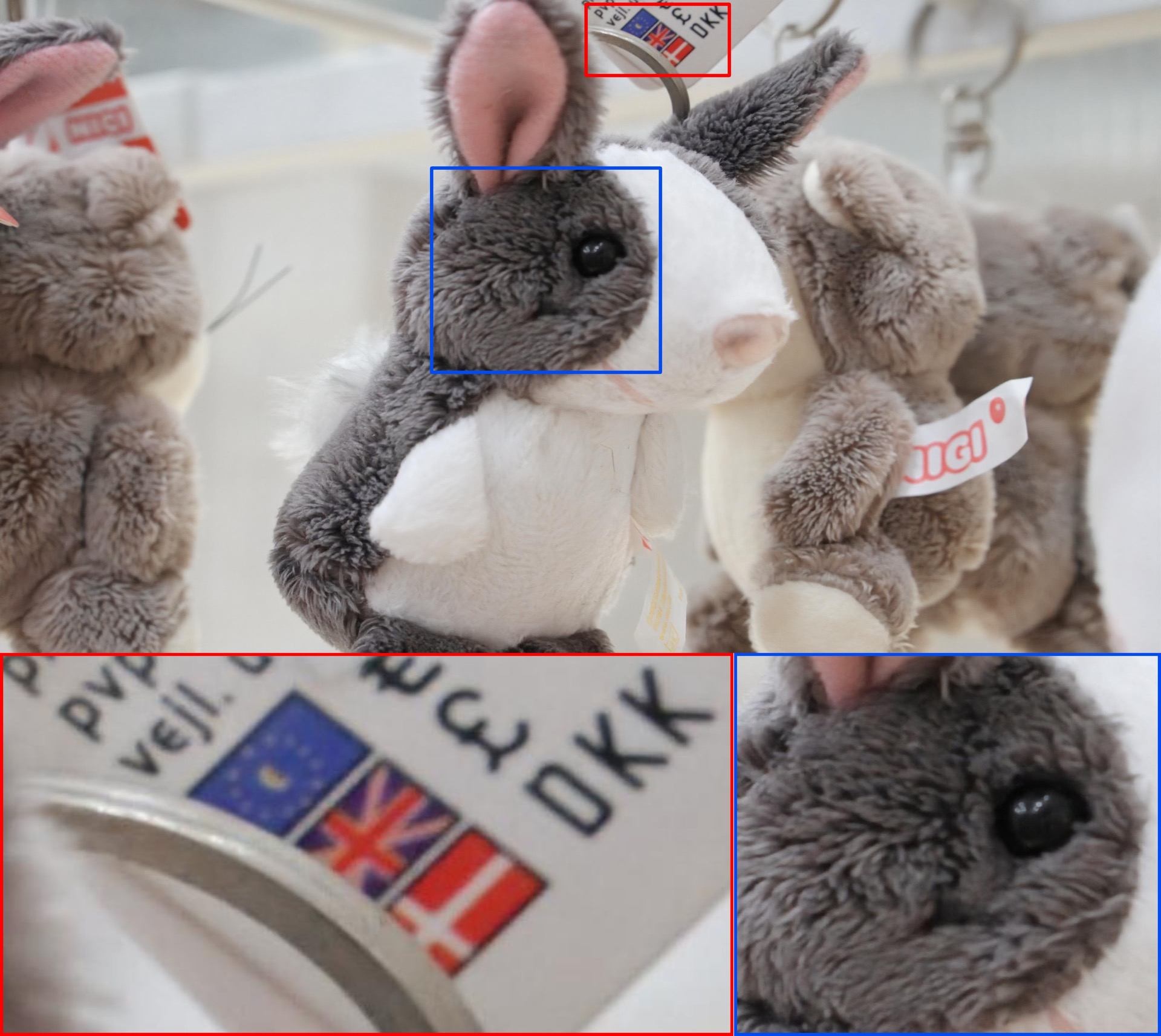} &
    \includegraphics[width=0.14\linewidth]{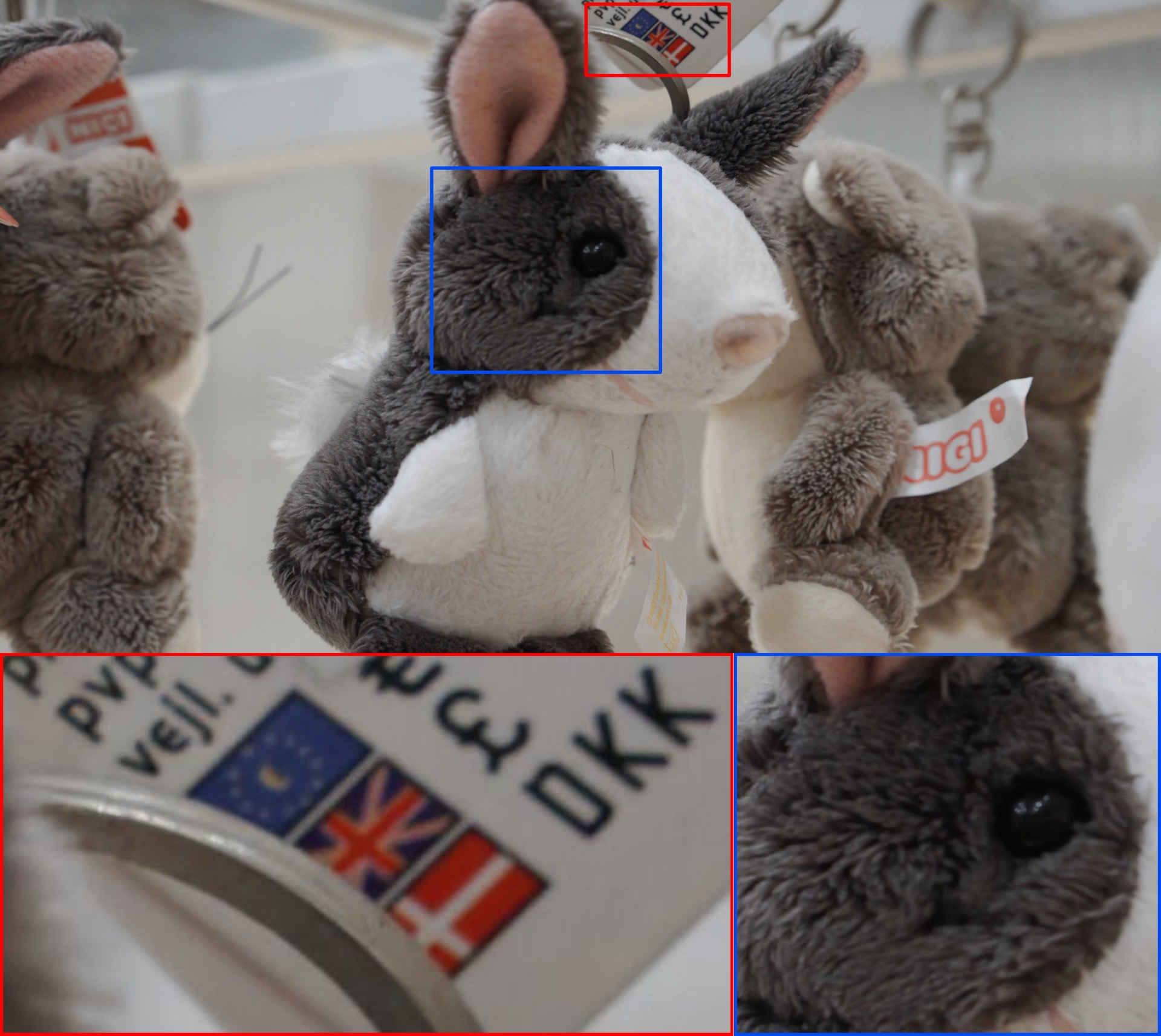} &
    \includegraphics[width=0.14\linewidth]{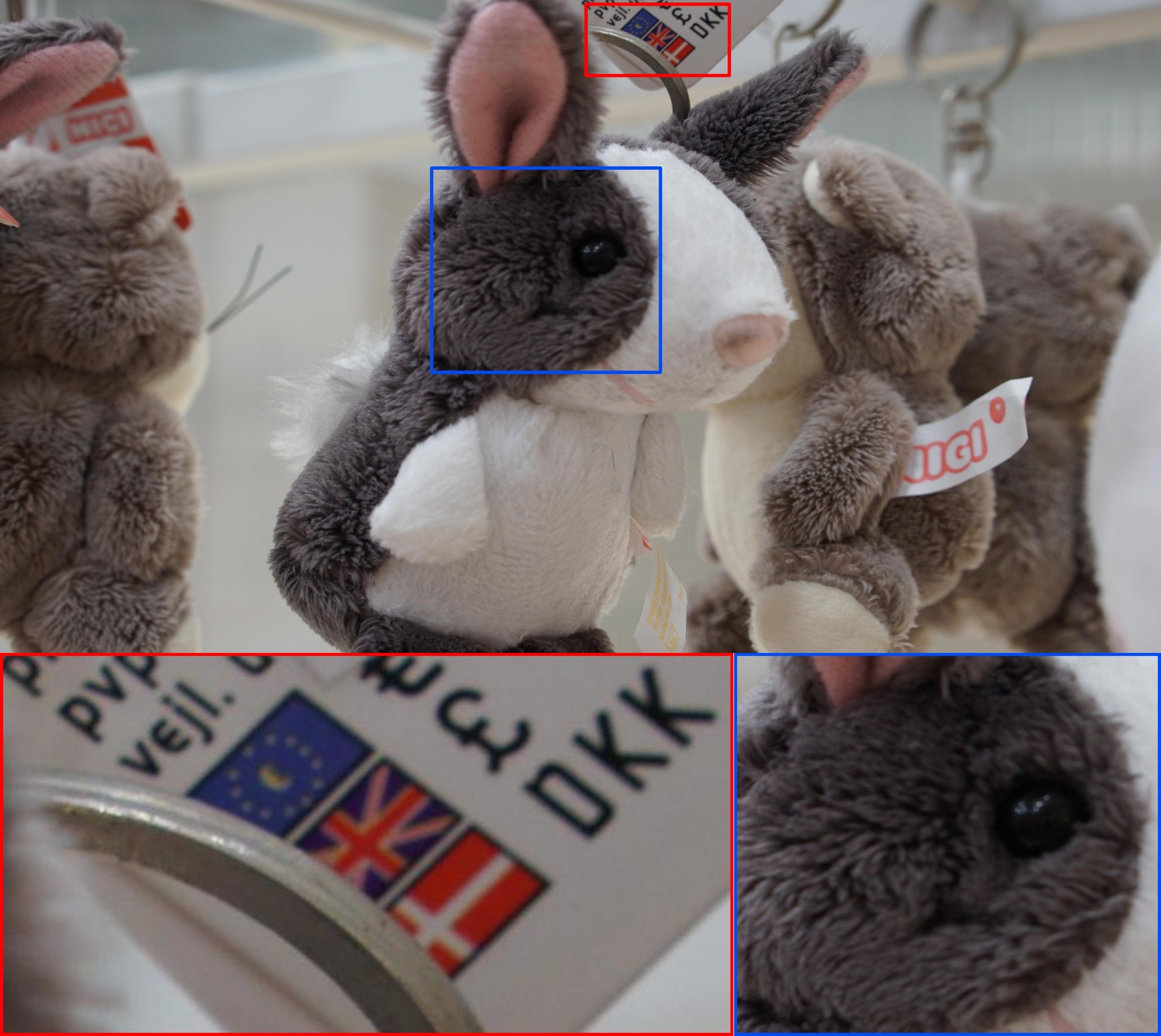} \\

    \\[-1pt]
    {\scriptsize{Input}} &
    {\scriptsize{UFormer}\cite{wang2022uformer}} &
    {\scriptsize{Retformer}\cite{cai2023retinexformer}} &
    {\scriptsize{CIDNet}\cite{yan2025hvi}} &
    {\scriptsize{ICLR}\cite{xu2025iclr}} &
    {\scriptsize{Ours}} &
    {\scriptsize{GT}} \\
  \end{tabular}

  \caption{Visual comparisons of the enhanced results by different methods on LSRW-Nikon\cite{hai2023r2rnet} and UHD-LL\cite{li2023embedding}. Zoom in for the best view.}
  \label{fig:lsrw_uhdll_visual}
\end{figure} 

\noindent\textbf{Metrics.}
For reference-based assessment, we report Peak Signal-to-Noise Ratio~(PSNR), Structural Similarity~(SSIM)~\cite{wang2004image}, and Learned Perceptual Image Patch Similarity~(LPIPS)~\cite{zhang2018unreasonable} with AlexNet~\cite{krizhevsky2012imagenet}. For no-reference evaluation, we employ the  Image Quality Assessment~(IQA) score from Q-Align~\cite{wu2023q}, a perceptual quality evaluator based on large multimodal models, and using Natural Image Quality Evaluator~(NIQE)~\cite{mittal2012making} as a complementary naturalness measure. Furthermore, we provide the number of model parameters and report the inference speed in Frames Per Second (FPS) measured at $600 \times 400$ resolution.

\subsection{Experimental Settings}
The proposed ATP framework is implemented in PyTorch and trained on a single NVIDIA 4090 GPU with 48~GB of memory. We adopt the AdamW~\cite{loshchilov2017decoupled} optimizer with $\beta_1 = 0.9$ and $\beta_2 = 0.999$. The learning rate follows a cosine annealing schedule~\cite{loshchilov2016sgdr}, decaying from the initial value to $1 \times 10^{-6}$. GEAM is trained for 300 epochs with a learning rate of $2 \times 10^{-4}$ and a batch size of 16, while DPM is trained for 1000 epochs with a learning rate of $1 \times 10^{-3}$ and a batch size of 8. Training images are randomly cropped to $256 \times 256$ patches and augmented with random flips and rotations.

\begin{figure*}[tb]
  \centering
  \setlength{\tabcolsep}{1.0pt}
  \renewcommand{\arraystretch}{0.3}

  \begin{tabular}{cccccc}

    \raisebox{0.065\linewidth}{\rotatebox{90}{\scriptsize DICM}} &
    \includegraphics[width=0.19\linewidth]{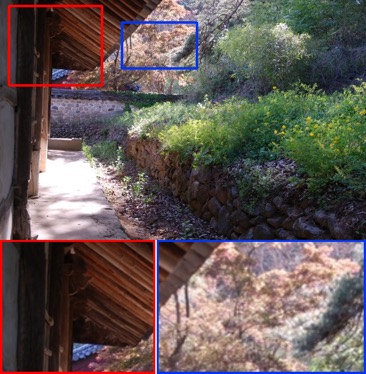} &
    \includegraphics[width=0.19\linewidth]{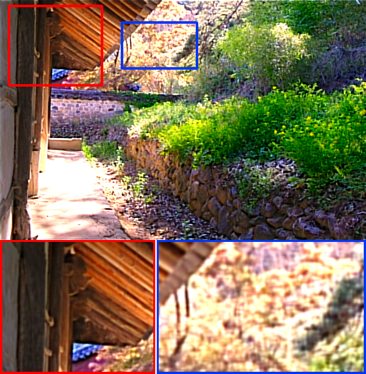} &
    \includegraphics[width=0.19\linewidth]{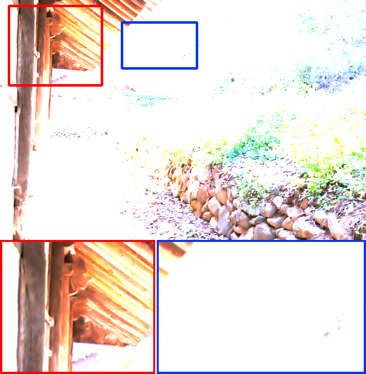} &
    \includegraphics[width=0.19\linewidth]{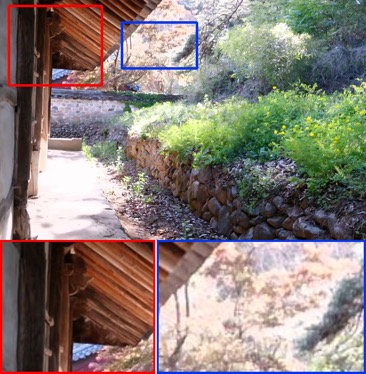} &
    \includegraphics[width=0.19\linewidth]{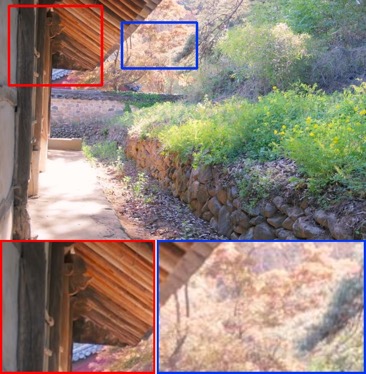}  \\

    \raisebox{0.075\linewidth}{\rotatebox{90}{\scriptsize LIME}} &
    \includegraphics[width=0.19\linewidth]{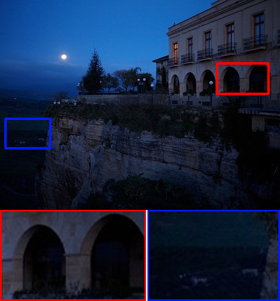} &
    \includegraphics[width=0.19\linewidth]{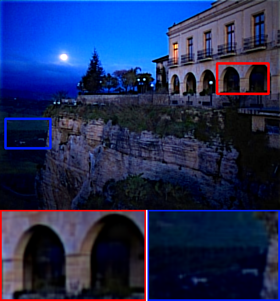} &
    \includegraphics[width=0.19\linewidth]{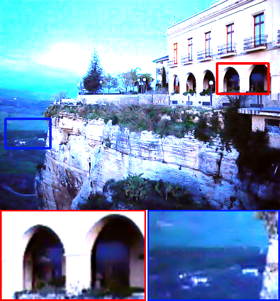} &
    \includegraphics[width=0.19\linewidth]{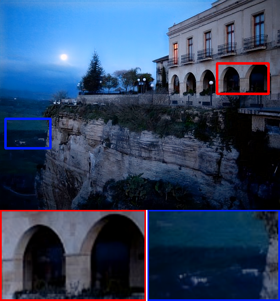} &
    \includegraphics[width=0.19\linewidth]{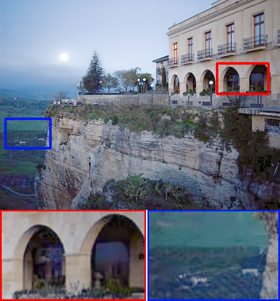}  \\

    \raisebox{0.075\linewidth}{\rotatebox{90}{\scriptsize MEF}} &
    \includegraphics[width=0.19\linewidth]{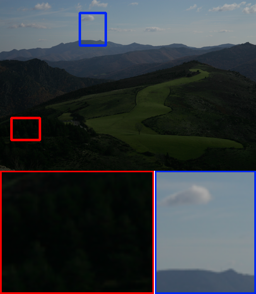} &
    \includegraphics[width=0.19\linewidth]{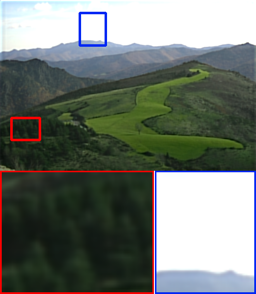} &
    \includegraphics[width=0.19\linewidth]{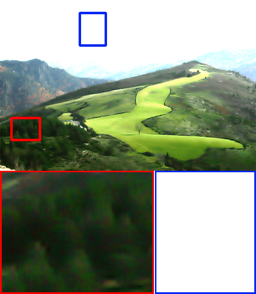} &
    \includegraphics[width=0.19\linewidth]{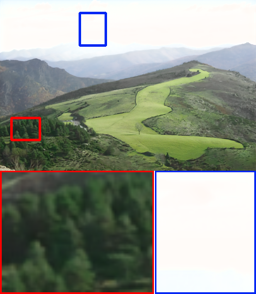} &
    \includegraphics[width=0.19\linewidth]{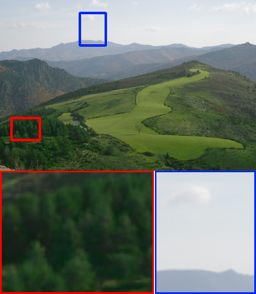}  \\

    \raisebox{0.050\linewidth}{\rotatebox{90}{\scriptsize NPE}} &
    \includegraphics[width=0.19\linewidth]{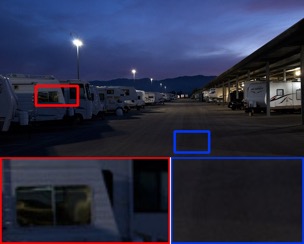} &
    \includegraphics[width=0.19\linewidth]{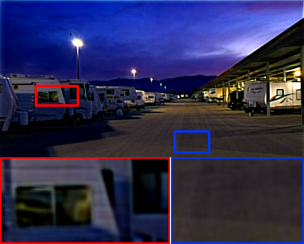} &
    \includegraphics[width=0.19\linewidth]{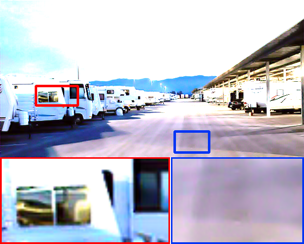} &
    \includegraphics[width=0.19\linewidth]{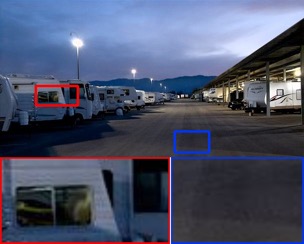} &
    \includegraphics[width=0.19\linewidth]{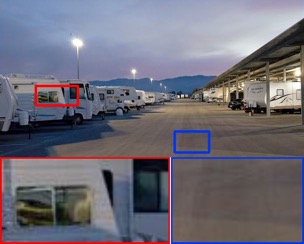}  \\

    \raisebox{0.085\linewidth}{\rotatebox{90}{\scriptsize VV}} &
    \includegraphics[width=0.19\linewidth]{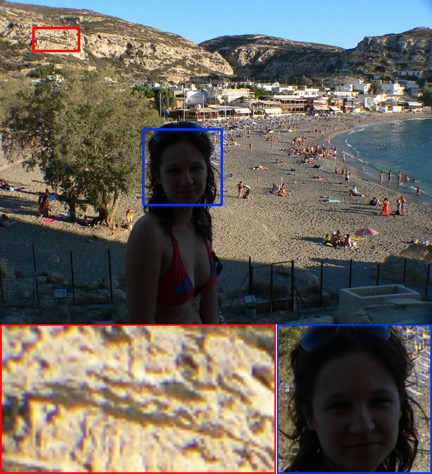} &
    \includegraphics[width=0.19\linewidth]{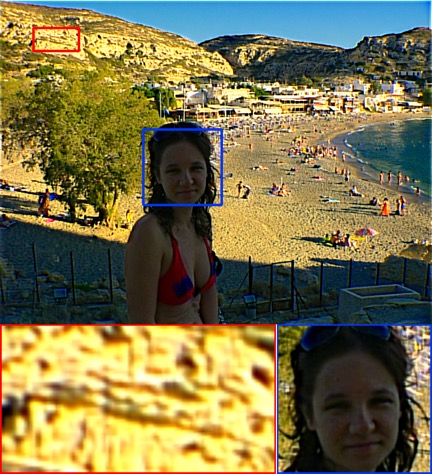} &
    \includegraphics[width=0.19\linewidth]{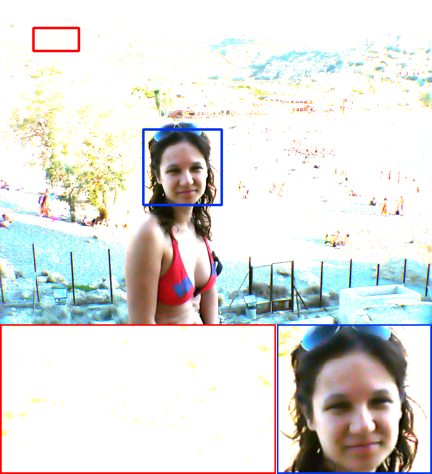} &
    \includegraphics[width=0.19\linewidth]{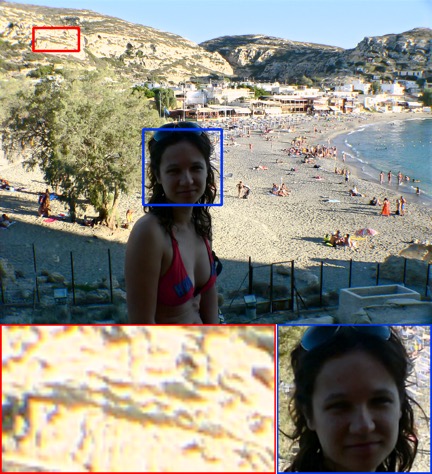} &
    \includegraphics[width=0.19\linewidth]{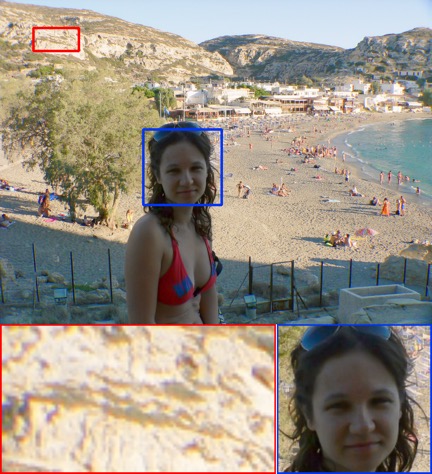}  \\

    \\[-1pt]
    & {\scriptsize{Input}} &
      {\scriptsize{MIRNet}\cite{zamir2020learning}} &
      {\scriptsize{RUAS}\cite{liu2021retinex}} &
      {\scriptsize{CIDNet}\cite{yan2025hvi}} &
      {\scriptsize{Ours}} \\
  \end{tabular}

  \caption{Visual comparison on the five unpaired datasets. Zoom in for the best view.}
  \label{fig:unpaired_visual}
  \vspace{-10pt}
\end{figure*}

\begin{figure}[tb]
  \centering
  \includegraphics[width=\linewidth,height=7.5cm,keepaspectratio]{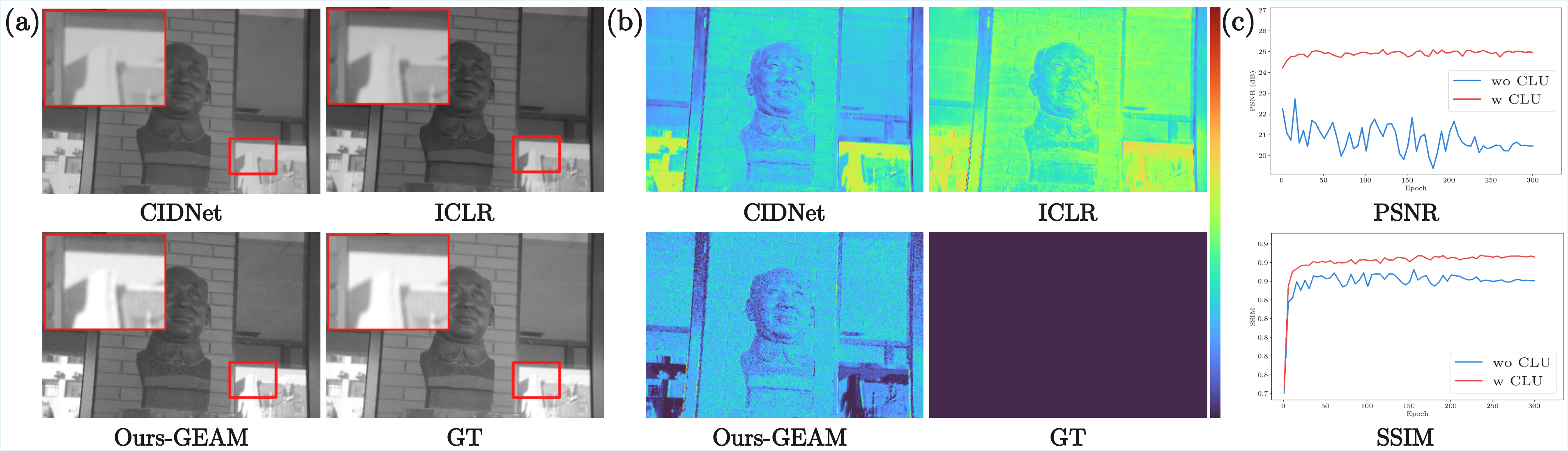}

  \caption{Analysis of global energy alignment and CLU. (a) Reconstructed Y-channel DWT-LL components. (b) Absolute error maps. (c) PSNR and SSIM training curves with and without CLU, given the low global luminance error evidenced in (a) and (b).}
  \label{fig:LL_error}
  \vspace{-5pt}
\end{figure}

\subsection{Comparisons with State-of-the-Art Methods}

\begin{table*}[t]
\centering
\caption{No-reference evaluation on five low-light datasets. Best results are in \textbf{bold} and second best are \underline{underlined}.}
\label{tab:qalign_iqa_niqe}
\setlength{\tabcolsep}{4pt}
\renewcommand{\arraystretch}{1.15}
\footnotesize
\resizebox{\textwidth}{!}{%
\begin{tabular}{l cc cc cc cc cc}
\toprule
\multirow{2}{*}{Method} &
\multicolumn{2}{c}{DICM} &
\multicolumn{2}{c}{LIME} &
\multicolumn{2}{c}{MEF} &
\multicolumn{2}{c}{NPE} &
\multicolumn{2}{c}{VV} \\
\cmidrule(lr){2-3}\cmidrule(lr){4-5}\cmidrule(lr){6-7}\cmidrule(lr){8-9}\cmidrule(lr){10-11}
& IQA$\uparrow$ & NIQE$\downarrow$
& IQA$\uparrow$ & NIQE$\downarrow$
& IQA$\uparrow$ & NIQE$\downarrow$
& IQA$\uparrow$ & NIQE$\downarrow$
& IQA$\uparrow$ & NIQE$\downarrow$ \\
\midrule
RUAS\cite{liu2021retinex}          & 2.04 & 7.05 & 2.11 & 5.37 & 2.69 & 5.41 & 1.63 & 7.09 & 2.27 & 4.75 \\
MIRNet\cite{zamir2020learning}        & 2.69 & 5.05 & 2.40 & 5.55 & 2.17 & 5.50 & 2.21 & 5.53 & 2.58 & 4.47 \\
UFormer\cite{wang2022uformer}       & 2.54 & 5.32 & 2.49 & 5.32 & 2.31 & 5.52 & 2.43 & 5.96 & 2.57 & 5.39 \\
Retinexformer\cite{cai2023retinexformer} & 3.18 & 4.13 & 3.25 & 4.21 & 3.05 & 4.18 & 2.95 & 4.21 & 2.92 & 3.26 \\
GSAD\cite{hou2023global}          & 3.25 & 4.15 & \underline{3.31} & 4.60 & 3.40 & 4.45 & \underline{3.36} & 4.72 & 3.01 & 4.10 \\
CIDNet\cite{yan2025hvi}        & \textbf{3.36} & \underline{3.79} & 3.29 & 4.52 & \textbf{3.50} & \underline{3.96} & 3.06 & \textbf{4.00} & \underline{3.09} & \underline{3.09} \\
\midrule
Ours          & \underline{3.34} & \textbf{3.77} & \textbf{3.35} & \textbf{3.78} & \underline{3.44} & \textbf{3.41} & \textbf{3.41} & \underline{4.12} & \textbf{3.15} & \textbf{3.07} \\
\bottomrule
\end{tabular}%
}
\vspace{-10pt}
\end{table*}

\noindent\textbf{Results on LOL datasets.}
We compare ATP against 13 representative methods on the LOLv1 and LOLv2 benchmarks. As reported in \cref{tab:lol_with_params}, our method achieves the best PSNR on all three datasets, surpassing ICLR by 0.79~dB, 1.47~dB, and 0.64~dB, respectively. ATP also attains the highest SSIM on all three datasets and the best LPIPS on LOLv1 and LOLv2-Real. On LOLv2-Real, ATP surpasses the previous best by 1.14~dB in PSNR, indicating substantial improvement in handling real-world low-light degradations. ATP maintains exceptional inference efficiency, achieving over 500 and 100 FPS on GEAM and the full model, respectively. Qualitative comparisons in \cref{fig:lol_visual} corroborate these findings, our method exhibits superior luminance consistency, accurate color reproduction, and high-fidelity texture restoration. For instance, our method restores finer textures and more accurate colors on the blue chairs~(row~2) and blue sofas~(row~3). In extremely dark scenes containing high-intensity light sources(row~1 and row~4), our method effectively restores the luminance and color of dark regions while preserving the original appearance of the light sources, avoiding over-exposure artifacts or unnatural dimming observed in competing approaches. Additional visual comparisons are provided in the supplementary materials.

\noindent\textbf{Results on LSRW-Nikon and UHD-LL.}
Results on two additional challenging benchmarks are reported in \cref{tab:lsrw_uhdll_with_params}. ATP achieves state-of-the-art performance across all metrics on both datasets. We report additional results on a registered version of the LSRW-Nikon (LSRW-Nikon-Reg), as certain original outdoor image pairs exhibit local offsets~\cite{hai2023r2rnet}. Details of the registration strategy are provided in the supplementary materials. Visual comparisons in \cref{fig:lsrw_uhdll_visual} further highlight the advantages of ATP: even in scenes that are not extremely dark, it preserves intricate textures and maintains faithful luminance. This effectively avoids the halo artifacts prevalent in UFormer, as well as the unnatural over-brightening and color distortion common in other approaches. These consistent gains across diverse resolutions and conditions validate the robust generalization capability of our framework.

\noindent\textbf{Results on unpaired datasets.}
To evaluate generalization without ground truth, we conduct experiments on five unpaired benchmarks using Q-Align IQA and NIQE. As shown in \cref{tab:qalign_iqa_niqe}, ATP achieves superior perceptual quality, obtaining the highest IQA scores on LIME, NPE, and VV, while remaining highly competitive on DICM and MEF. For NIQE, our method attains the lowest scores on four of five datasets, demonstrating its ability to produce natural results with minimal artifacts. Qualitative results in \cref{fig:unpaired_visual} further confirm that ATP yields more visually pleasing outputs with well-balanced contrast. Notably, for challenging unpaired samples with locally under-exposed regions~(row 2), our method demonstrates a superior ability to effectively enhance visibility while recovering high-fidelity textures and vivid colors in extreme darkness.

\noindent\textbf{Effectiveness of global energy anchoring.}
We additionally report the performance of GEAM alone in \cref{tab:lol_with_params} and \cref{tab:lsrw_uhdll_with_params}. By predicting only a $3 \times 4$ affine projection matrix, GEAM already achieves~24.001 dB on LOLv2-Real, outperforming the previous state-of-the-art. The visualizations of Y-channel LL-components and their corresponding absolute error maps in \cref{fig:LL_error} further corroborate this performance. Notably, GEAM maintains lower luminance error, especially in high-light regions such as the sky, it achieves luminance levels highly consistent with the ground truth. This stands in contrast to the significant luminance deviations observed in CIDNet and ICLR, highlighting that employing complex nonlinear networks to approximate inherently linear global luminance adjustments is suboptimal and less robust.

\section{Ablation Study}
In this section, we conduct ablation studies on the LOLv2-Real dataset to validate the contribution of each component and design choice in ATP.

\begin{table}[t]
\centering
\caption{Ablation study of different components.}
\label{tab:ablation_components}
\setlength{\tabcolsep}{4pt}
\renewcommand{\arraystretch}{1.15}
\footnotesize
\begin{tabular}{c c c c c c c c}
\toprule
\multirow{2}{*}{Set} & \multicolumn{3}{c}{Components} & \multirow{2}{*}{MG} & \multirow{2}{*}{PSNR$\uparrow$} & \multirow{2}{*}{SSIM$\uparrow$} & \multirow{2}{*}{LPIPS$\downarrow$} \\
\cmidrule(lr){2-4}
& GEAM & DPM & CLU &  &  &  &  \\
\midrule
A & $\times$ & $\surd$ & $\times$ & -        & 22.309 & 0.8523 & 0.1268 \\
B & $\times$ & $\surd$ & $\surd$  & -        & 15.069 & 0.6652 & 0.3322 \\
C & $\surd$  & $\surd$ & $\times$ & $\surd$  & 22.718 & 0.8567 & 0.1304 \\
D & $\surd$  & $\times$ & $\times$ & $\times$  & 24.001 & 0.7279 & 0.2490 \\
E & $\surd$  & $\surd$ & $\surd$  & $\times$ & 24.839 & 0.8811 & 0.1054 \\
F & $\surd$  & $\surd$ & $\surd$  & $\surd$  & 25.140 & 0.8851 & 0.0959 \\
\bottomrule
\end{tabular}
\end{table}

\begin{table}[t]
\centering
\caption{Ablation study on Matrix shape (Degrees of Freedom, DoF).}
\label{tab:shape_formulation}
\setlength{\tabcolsep}{4pt}
\renewcommand{\arraystretch}{1.15}
\footnotesize
\begin{tabular}{c l l c c c}
\toprule
Set & Matrix shape (DoF) & Formulation & PSNR$\uparrow$ & SSIM$\uparrow$ & LPIPS$\downarrow$ \\
\midrule
A & $1\times1~(1)$                              & $a \cdot \mathbf{I}$                          & 23.451 & 0.8642 & 0.1149 \\
B & Diag$(\mathbf{d})$ + $\mathbf{b}\,(6)$                        & $ \mathbf{A} \cdot \mathbf{I} + \mathbf{b}$   & 23.014 & 0.8764 & 0.1063 \\
C & $3\times3~(9)$                              & $ \mathbf{A} \cdot \mathbf{I}$                & 21.674 & 0.8651 & 0.1265 \\
D & $3\times4~(12)$                             & $\mathbf{A} \cdot \mathbf{I} + \mathbf{b}$    & 25.140 & 0.8851 & 0.0959 \\
\bottomrule
\end{tabular}
\end{table}

\noindent\textbf{Core component analysis. }
As shown in \cref{tab:ablation_components}, GEAM provides a critical global alignment prior whose absence leads to a substantial performance drop in the subsequent refinement stage. This alignment is further stabilized by CLU. As illustrated in \cref{fig:LL_error}(c), removing CLU triggers significant oscillations in both PSNR and SSIM during training. This occurs because minor residual luminance errors can disrupt the optimization process, causing the network to oscillate between adjusting luminance and restoring details, thereby preventing it from focusing on texture recovery. By suppressing these unnecessary luminance modifications, CLU ensures a consistent optimization objective for the refinement stage. Furthermore, Matrix Guidance (MG) acts as a structural regulator that modulates refinement intensity based on the magnitude of global alignment. This enables adaptive denoising strength and allows the full configuration to achieve the best overall performance.

\begin{table}[t]
\centering
\caption{Ablation of applying DWT on different channels.}
\label{tab:dwt_color_ablation}
\setlength{\tabcolsep}{6pt}
\renewcommand{\arraystretch}{1.15}
\footnotesize
\begin{tabular}{c c c c}
\toprule
DWT applied on & PSNR$\uparrow$ & SSIM$\uparrow$ & LPIPS$\downarrow$ \\
\midrule
YUV              & 24.318 & 0.8768 & 0.1028 \\
RGB              & 24.047 & 0.8777 & 0.1049 \\
Y  & 25.140 & 0.8851 & 0.0959 \\
\bottomrule
\end{tabular}
\end{table}

\begin{table}[t]
\centering
\caption{Ablation study on loss weights.}
\label{tab:loss_weight_ablation}
\setlength{\tabcolsep}{4pt}
\renewcommand{\arraystretch}{1.15}
\scriptsize
\resizebox{\linewidth}{!}{%
\begin{tabular}{l c c c c c c c c c c c}
\toprule
\multirow{2}{*}{Loss} &
\multicolumn{3}{c}{GEAM} &
\multicolumn{5}{c}{DPM} &
\multirow{2}{*}{PSNR$\uparrow$} &
\multirow{2}{*}{SSIM$\uparrow$} &
\multirow{2}{*}{LPIPS$\downarrow$} \\
\cmidrule(lr){2-4}\cmidrule(lr){5-9}
& $\lambda_\text{rec}$ & $\lambda_{\text{mat}}$ & $\lambda_{\text{diag}}$
& $\lambda_{\text{color}}$ & $\lambda_{\text{freq}}$ & $\lambda_{\text{lum}}$ & $\lambda_{\text{ssim}}$ & $\lambda_{\text{vgg}}$
& & & \\
\midrule
A & 1   & 1   & 0.01 & 1 & 1   & 0.01 & 0.5 & 0.01 & 24.912 & 0.8795 & 0.1021 \\
B & 1   & 1   & 0.1  & 1 & 1   & 0.1  & 0.5 & 0.01 & 24.433 & 0.8802 & 0.1090 \\
C & 1   & 1   & 0.1  & 1 & 1   & 0.01 & 0.3 & 0.01 & 24.355 & 0.8818 & 0.0971 \\
D & 1   & 1   & 0.1  & 1 & 1.5 & 0.01 & 0.5 & 0.01 & 24.674 & 0.8827 & 0.1002 \\
E & 1   & 0.5 & 0.2  & 1 & 1   & 0.01 & 0.5 & 0.01 & 23.651 & 0.8800 & 0.0978 \\
F & 0.5 & 1   & 0.1  & 1 & 1   & 0.01 & 0.5 & 0.01 & 24.412 & 0.8825 & 0.0977 \\
G & 1   & 1   & 0.1  & 1 & 1   & 0.01 & 0.5 & 0.01 & 25.140 & 0.8851 & 0.0959 \\
\bottomrule
\end{tabular}%
}
\end{table}

\noindent\textbf{Matrix shape analysis.}
\cref{tab:shape_formulation} provides a detailed comparison of different matrix parameterizations. The results demonstrate that the full $3\times4$ affine transformation with 12-DoF substantially outperforms lower-rank alternatives, primarily due to its superior capacity to capture complex cross-channel interactions alongside per-channel intensity offsets. Notably, the $3\times3$ formulation lacking a bias term (SetC) surprisingly underperforms the simpler diagonal form (SetB). This performance gap highlights that the bias term is indispensable for accurately modeling the additive component of global photometric shifts, suggesting that purely multiplicative adjustments are insufficient for mapping real-world lighting variations.

\noindent\textbf{DWT scope analysis.}
As shown in \cref{tab:dwt_color_ablation}, restricting DWT to the Y channel alone yields the best performance. Extending wavelet decomposition to chrominance channels degrades the results, as the inherently lower SNR of these channels under low-light conditions makes frequency-domain manipulation prone to amplifying color noise and inducing hue drift.

\noindent\textbf{Loss weights analysis.}
We further examine the sensitivity to loss weight configurations in \cref{tab:loss_weight_ablation}. Among all terms, reducing $\lambda_\text{mat}$ causes the largest performance drop, highlighting the importance of matrix supervision in GEAM. The adopted configuration (SetG) achieves the best balance across all metrics.

\section{Conclusion}
In this paper, we find that a simple linear operator with only 12 degrees of freedom can correct the majority of the global energy shifts in LLIE. Based on this, we propose the ATP framework to explicitly decouple global energy alignment from local detail refinement. By shifting the global alignment task to a global linear operator, ATP relieves the high-capacity refinement network from modeling global variations, thereby eliminating overfitting on linear physical constraints. Specifically, the macro anchoring stage utilizes GEAM to learn a scene-adaptive $3\times4$ projection matrix, which stabilizes luminance distribution and corrects color shifts. Under matrix-guided modulation, DPM performs wavelet-based high-frequency denoising and spatial-domain chrominance refinement. Concurrently, the CLU strategy maintains global consistency by restricting unnecessary luminance fluctuations, effectively directing the network to concentrate on fine-grained texture recovery.

\par\vfill\par

\clearpage

\bibliographystyle{splncs04}
\bibliography{main}

\clearpage
\appendix
\begin{center}
\Large\textbf{Supplementary Material for \\ Anchor then Polish for Low-light Enhancement }
\end{center}

\setcounter{page}{1}
\setcounter{figure}{0}
\setcounter{table}{0}
\setcounter{equation}{0}

\section{Other Details of Global Energy Anchoring}

\subsection{Observations on More Datasets}
In Low-Light Image Enhancement~(LLIE) tasks, the energy difference between low-light images and normal images can be decomposed into luminance, chrominance(color), and texture/noise components. The luminance and chrominance components can be directly computed in the YUV color space:
\begin{equation}
E_{\text{lum}}=\|\mathbf{Y}_{\text{img}}-\mathbf{Y}_{\text{gt}}\|_2^2,
E_{\text{chr}}=\|\mathbf{U}_{\text{img}}-\mathbf{U}_{\text{gt}}\|_2^2+\|\mathbf{V}_{\text{img}}-\mathbf{V}_{\text{gt}}\|_2^2,
\label{eq:lum_color}
\end{equation}
where $\text{img}\in\{\text{low},\text{GEA}\}$.The texture and noise components primarily represent variations in edges and high-frequency details. We extract these features using the Laplacian operator, which is a second-order derivative operator that measures grayscale transitions within an image. This operator intuitively enhances edges and textural details to provide an effective approximation of the texture component:
\begin{equation}
    \label{eq:texture}
    \mathbf{Tex}_\text{img}= \text{Laplacian}(\text{Gray}(\mathbf{I}_\text{img})/255),
    E_\text{tex} = \|\mathbf{Tex}_\text{img}-\mathbf{Tex}_\text{gt}\|^{2}_{2},
\end{equation}
where the $\text{Gray}(\cdot)$ denotes conversion to grayscale.

\begin{figure}[tb]
  \centering
  \includegraphics[width=\linewidth]{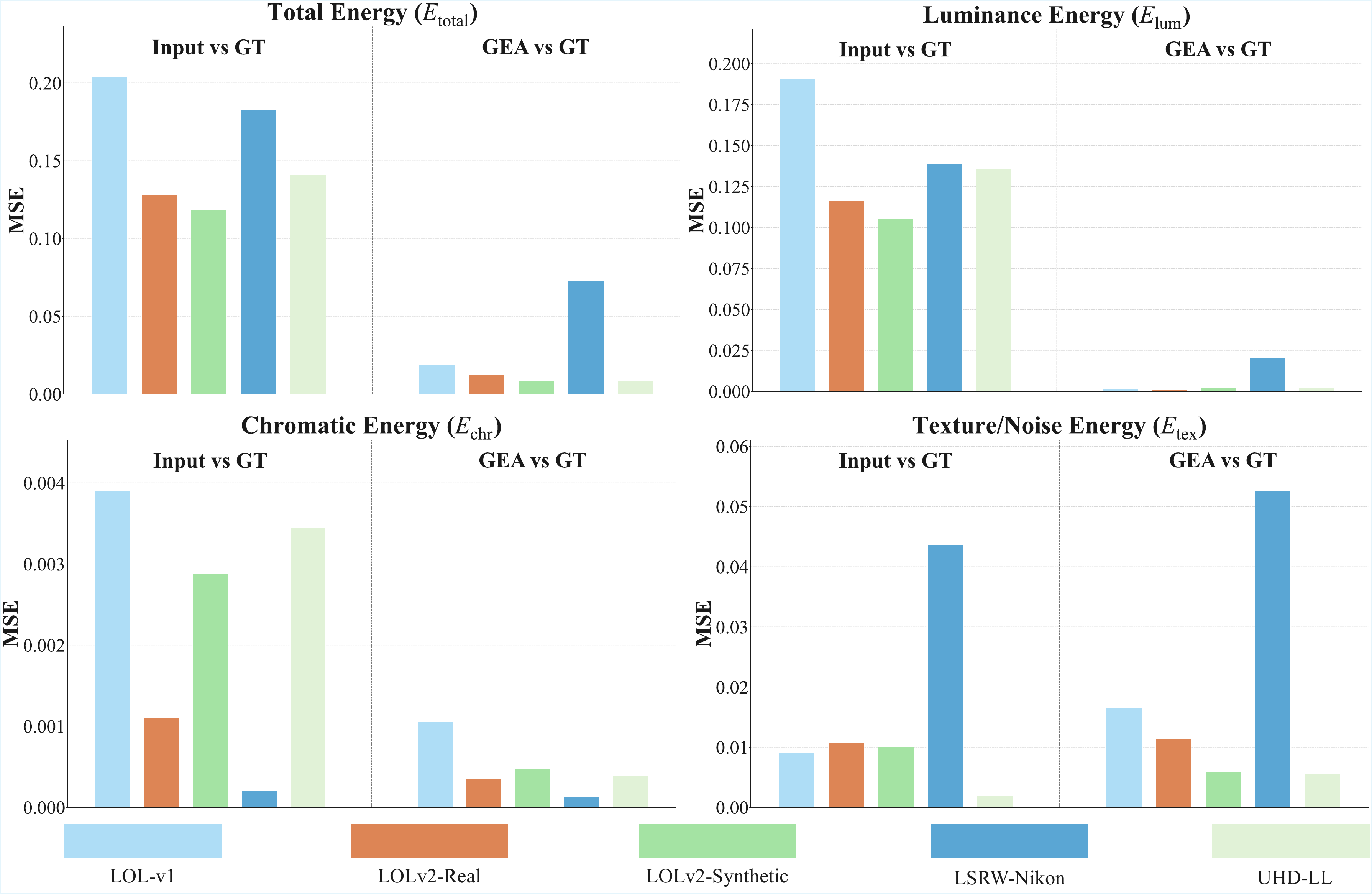}
  \caption{Observations after applying GEA to more datasets.}
  \label{fig:more_observation}
\end{figure}

Based on~\cref{eq:lum_color} and~\cref{eq:texture}, we provide observations across more datasets to demonstrate the effectiveness of GEA on the LLIE task. As shown in~\cref{fig:more_observation}, GEA effectively anchors luminance shifts while correcting chromatic aberrations through channel blending across all observed datasets. However, because GEA anchors global energy through linear operators, non-linear texture and noise variations tend to be amplified during the anchoring process. This observation motivates the use of the global matrix as a conditional signal in subsequent refinement processes, enabling the network to adaptively adjust denoising intensity based on the degree of global energy anchoring.

\subsection{Matrix Structure}
In standard Image Signal Processing (ISP) frameworks, the luminance of a pixel is defined as a weighted sum of the color channels, whereas the chromaticity is determined by the relative ratios between these channels. Following this principle, GEAM predicts an ISP-style affine mapping decomposed into diagonal gains and off-diagonal channel mixing:
\begin{equation}
\mathbf{\hat{I}} = 
\left(
\underbrace{
\begin{bmatrix}
a_\text{rr} & 0 & 0\\
0 & a_\text{gg} & 0\\
0 & 0 & a_\text{bb}
\end{bmatrix}}_{\mathbf{D}\text{ (diagonal gain)}}
+
\underbrace{
\begin{bmatrix}
0 & a_\text{rg} & a_\text{rb}\\
a_\text{gr} & 0 & a_\text{gb}\\
a_\text{br} & a_\text{bg} & 0
\end{bmatrix}}_{\mathbf{O}\text{ (off-diagonal mixing)}}
\right)
\mathbf{I}_\text{low}
+
\underbrace{
\begin{bmatrix}
b_\text{r}\\b_\text{g}\\b_\text{b}
\end{bmatrix}}_{\mathbf{b}},
\label{eq:app_rgb_affine}
\end{equation}
where $\mathbf{I}_\text{low}=[R,G,B]^{\top}$ and $\mathbf{\hat{I}}=[\hat{R}
,\hat{G},\hat{B}]^{\top}$ are the input and output pixels.

This structure allows the model to decouple brightness and color adjustments:
\begin{itemize}
    \item \textbf{Luminance primarily via $\mathbf{D}$:} The diagonal entries directly scale the intensity of each channel. Since luminance is a weighted sum of RGB, these gains provide the primary control for exposure.
    \item \textbf{Color primarily via $\mathbf{O}$:} The off-diagonal entries explicitly mix information between channels (\emph{e.g.}, adding a fraction of Green to Red). This alters the RGB ratios, which is the fundamental mechanism for color correction and saturation adjustment.
\end{itemize}

\begin{figure}[tb]
  \centering
  \includegraphics[width=\linewidth]{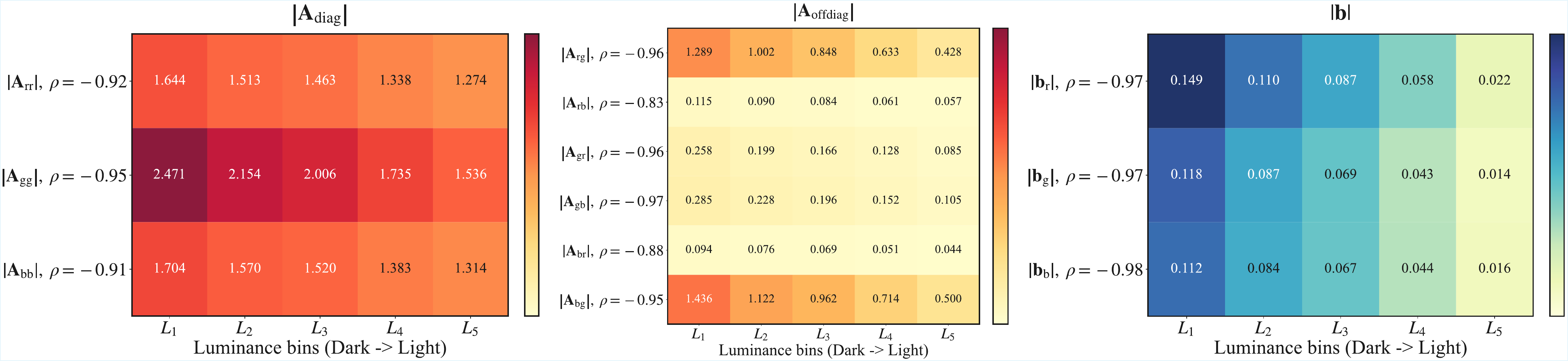}
  \caption{Matrix structure of GEAM prediction on LOLv2-Real. Where the $\rho$ indicates a monotonic relationship between matrix elements and input luminance. The closer $\left | \rho \right | $ is to 1, the stronger the relationship.}
  \label{fig:matrix_structure}
\end{figure}

In LLIE, the most significant transformation is the restoration of exposure, which is primarily handled by the diagonal components. Consequently, an ISP-consistent matrix is expected to be diagonal-dominant, where $\mathbf{D}$ provides the necessary exposure lift and $\mathbf{O}$ performs fine-grained color calibration. Furthermore, the magnitudes of these coefficients should adapt to the luminance level of the input image. Specifically, when the input image becomes brighter, the required gains for exposure in $\mathbf{D}$ and the additive offsets in $\mathbf{b}$ should decrease accordingly. Similarly, the values in the off-diagonal mixing matrix $\mathbf{O}$ should also scale down to prevent over-saturation and maintain natural color transitions. The statistical analysis of prediction parameters across different brightness intervals in~\cref{fig:matrix_structure} further validates that the matrix generated by GEAM strictly adheres to this physical prior. The pronounced trend of coefficients decreasing with increasing input brightness indicates that GEAM does not perform arbitrary enhancement. Moreover, the mixing intensity of the G channel in the R~($\mathbf{A}_\text{rg}$) and B~($\mathbf{A}_\text{bg}$) channels is significantly higher than other cross-channel interactions. This phenomenon aligns perfectly with the physical characteristics of the original image sensor, where the G channel has twice the sampling density of the R and B channels and a higher signal-to-noise ratio. This demonstrates that GEAM successfully learns and simulates the physical laws inherent in traditional ISP workflows, further proving the interpretability of GEA.

\begin{figure}[tb]
  \centering
  \includegraphics[width=0.8\linewidth]{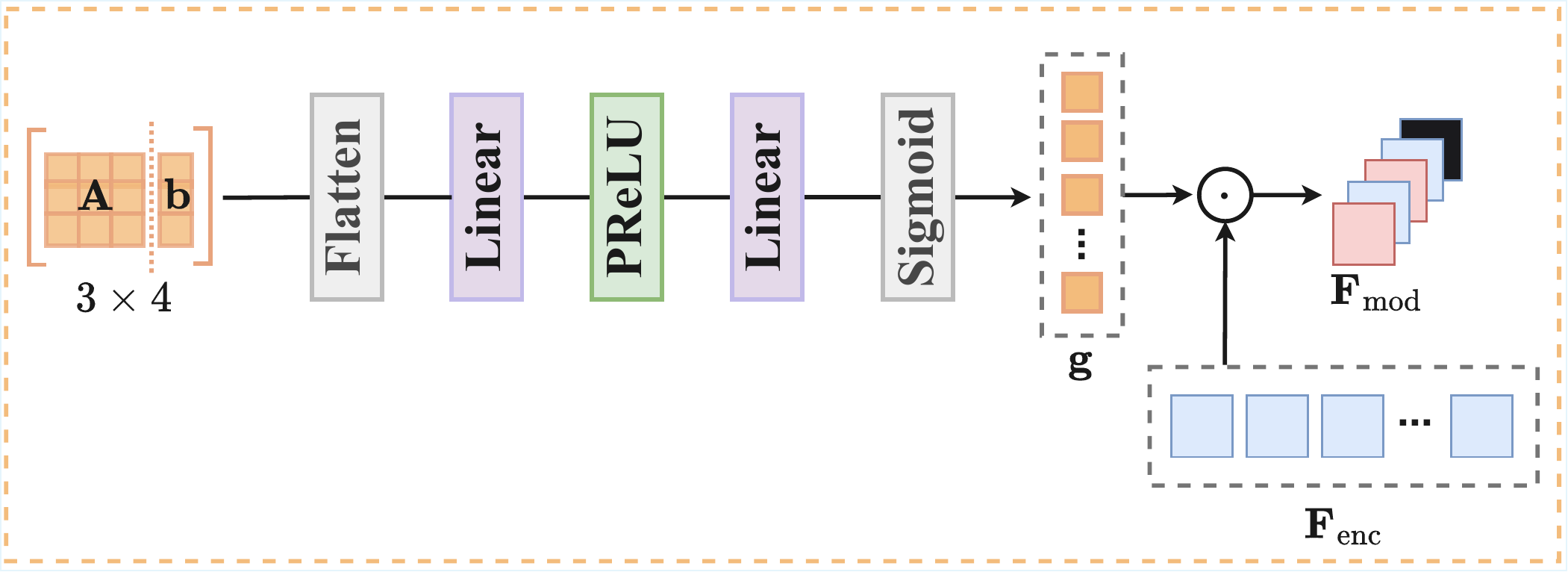}
  \caption{Parameter Embedding Module. Red indicates that the feature is amplified, black indicates that the feature is suppressed, and blue indicates that the feature remains unchanged.}
  \label{fig:pem}
\end{figure}

\section{Details of Matrix Guidance}
In this section, we detail the specific method of parameter embedding and the guiding mechanism of the global energy alignment matrix predicted during the anchoring stage for the texture refinement process in the polishing stage.

\subsection{Parameter Embedding}
Intuitively, the global energy alignment matrix theoretically amplifies low-light images, as demonstrated in~\cref{fig:matrix_structure}. During global energy alignment, residual noise in dark regions is also amplified to varying degrees depending on the matrix's strength. This led us to the following analysis: \textit{Could the intensity of global energy alignment serve as prior information to guide DPM in denoising and texture restoration?} This would enable DPM to distinguish whether the current high-frequency feature components stem from amplified noise due to excessive matrix energy or from the inherent edge textures of objects within the image.

Based on this analysis, we designed the feature modulation method illustrated in~\cref{fig:pem}. Specifically, we first flatten the matrix~$\mathbf{\hat{M}}$ by GEAM predicted into a 12-dimensional vector. This vector is then passed through a lightweight MLP to map it into sample-specific channel gating coefficients~$\mathbf{g}\in(0,2)^{C \times 1}$, which recalibrate the features output by the encoder:
\begin{equation}
    \mathbf{F}_{\text{mod}} = \text{Enc}([\mathbf{I}_{\text{align}};\mathcal{Y}_{\text{HF}}]) \odot \text{PE}(\mathbf{\hat{M}}) = \text{Enc}([\mathbf{I}_{\text{align}};\mathcal{Y}_{\text{HF}}]) \odot \mathbf{g}.
\end{equation}
This mechanism enables the DPM to learn whether each channel should be enhanced or suppressed under the current global energy intensity state. This preserves features more conducive to denoising and texture recovery, thereby achieving dynamic feature selection.

\begin{figure}[tb]
  \centering
  \includegraphics[width=\linewidth]{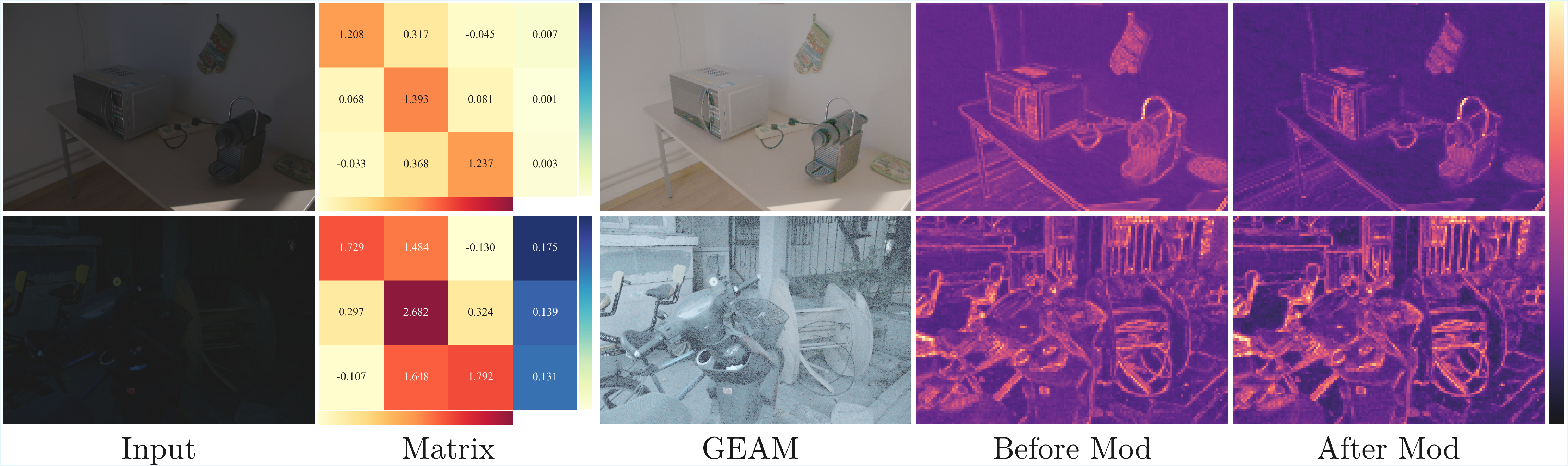}
  \caption{Visual analysis of the proposed Matrix Guidance mechanism.}
  \label{fig:matrix_feature}
\end{figure}

\subsection{Feature Modulation}
To empirically validate the efficacy of the proposed matrix-guided modulation, we conduct a visual analysis of the feature responses before and after the application of gating coefficients $\mathbf{g}$, as illustrated in~\cref{fig:matrix_feature}. It can be observed that the behavior of the proposed Matrix Guidance mechanism adapts dynamically to the degradation level of the input. For moderately low-light inputs where object visibility is largely retained (row 1), the requisite magnitude of the energy anchoring matrix remains comparatively low. In this scenario, the GEAM applies only subtle global illumination adjustments, thereby avoiding the over-amplification of latent noise. Consequently, the subsequent DPM refinement stage can prioritize the preservation of fine textures; redundant features are effectively suppressed to isolate edge information and facilitate superior detail recovery. Conversely, under severe low-light conditions where scene content is practically indistinguishable (row 2), the system necessitates a substantially stronger energy anchoring matrix. This requires GEAM to execute aggressive global adjustments, which inherently amplifies latent degradation and yields a noise-dominated intermediate representation. Consequently, the subsequent DPM refinement stage not only actively preserves robust structural features to maintain its denoising capability, but also attenuates a small fraction of less informative features to prevent severe noise from being misinterpreted as valid details, thereby achieving superior denoising performance.

\section{Registration Methods on the LSRW-Nikon}
To mitigate evaluation biases caused by pixel offsets, which occur in the LSRW-Nikon~\cite{hai2023r2rnet} because outdoor image pairs are often not perfectly aligned, we adopt the Enhanced Correlation Coefficient~\cite{evangelidis2008parametric}(ECC) method for image registration. By aligning the predicted and ground truth images at a subpixel level, we aim to ensure that the quantitative results reflect the restoration quality as faithfully as possible.

The ECC algorithm achieves this high-precision alignment by maximizing the zero-mean normalized cross-correlation (ZNCC) between the reference and the warped images. Let $\mathbf{I}_\text{gt}$ denote the reference GT image and $\mathbf{I}_\text{in}$ denote the input image. We seek an optimal geometric transformation $\mathbf{W}(\mathbf{x}; \mathbf{p})$, parameterized by a vector $\mathbf{p}$ (\emph{e.g.}, an affine transformation matrix), that maps pixel coordinates $\mathbf{x}$ from the input space to the GT space. The objective function to be maximized is defined as:
\begin{equation}
\max_{\mathbf{p}} \text{E}_\text{ECC}(\mathbf{p}) = \frac{\sum_{\mathbf{x}} \mathbf{\tilde{I}}_\text{gt}(\mathbf{x}) \mathbf{\tilde{I}}_\text{in}(\mathbf{W}(\mathbf{x}; \mathbf{p}))}{\sqrt{\sum_{\mathbf{x}} \mathbf{\tilde{I}}_\text{gt}(\mathbf{x})^2} \sqrt{\sum_{\mathbf{x}} \mathbf{\tilde{I}}_\text{in}(\mathbf{W}(\mathbf{x}; \mathbf{p}))^2}},
\end{equation}
where $\mathbf{\tilde{I}}_\text{gt}$ and $\mathbf{\tilde{I}}_\text{in}$ represent the zero-mean centered image intensities. This zero-mean formulation makes the registration process robust to global illumination changes and photometric distortions, which are inherently severe in low-light image pairs.

Once the optimal transformation matrix is determined, it is universally applied to warp the input, the GT, and the outputs of all evaluated enhancement methods. Following the warping operation, the boundary regions inevitably contain zero-padding or distorted pixels due to spatial shifting. To eliminate these border artifacts, we compute a shared valid bounding box by extracting the intersection of the non-empty warped regions and further cropping a fixed margin (\emph{e.g.}, $4$ pixels) from the borders. This strictly uniform alignment and cropping protocol guarantees that all competitive models are evaluated on the exact same valid spatial region, thereby preventing any metric discrepancies caused by individual misalignment or border artifacts.

\section{More Visualization Results}
In this section, we provide extensive qualitative comparisons to further validate the effectiveness of our proposed ATP.

\begin{figure*}[tb]
  \centering

  \setlength{\tabcolsep}{0.5pt}
  \renewcommand{\arraystretch}{0.3}

  \begin{tabular}{ccccccc}

    \includegraphics[width=0.14\linewidth]{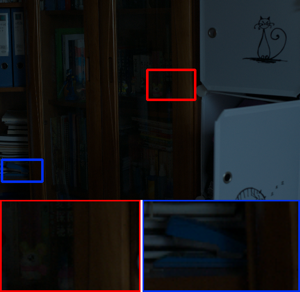} &
    \includegraphics[width=0.14\linewidth]{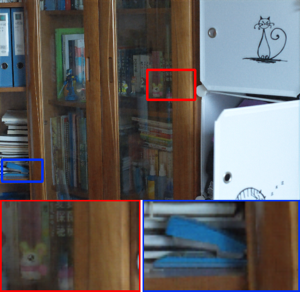} &
    \includegraphics[width=0.14\linewidth]{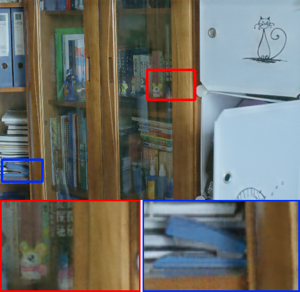} &
    \includegraphics[width=0.14\linewidth]{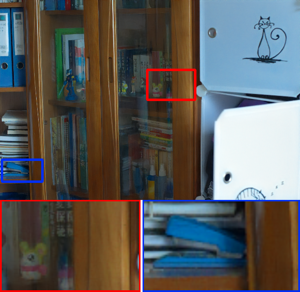} &
    \includegraphics[width=0.14\linewidth]{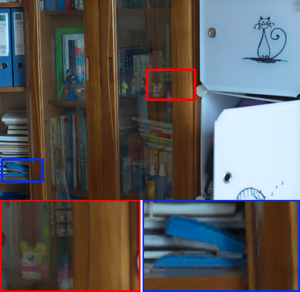} &
    \includegraphics[width=0.14\linewidth]{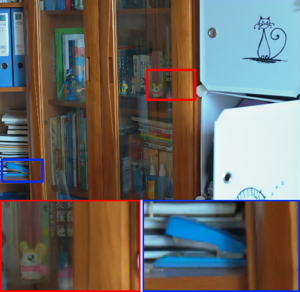} &
    \includegraphics[width=0.14\linewidth]{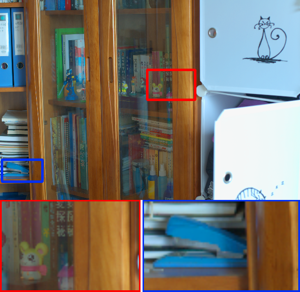} \\

    \includegraphics[width=0.14\linewidth]{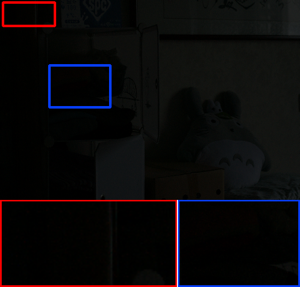} &
    \includegraphics[width=0.14\linewidth]{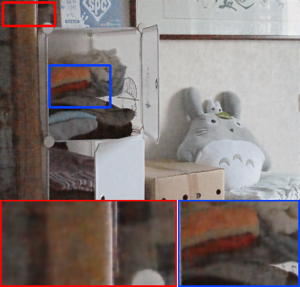} &
    \includegraphics[width=0.14\linewidth]{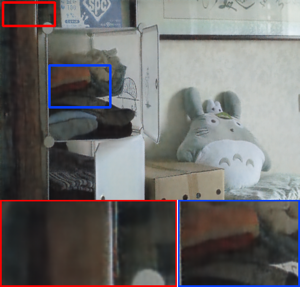} &
    \includegraphics[width=0.14\linewidth]{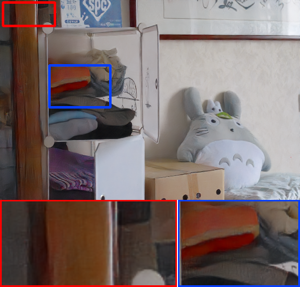} &
    \includegraphics[width=0.14\linewidth]{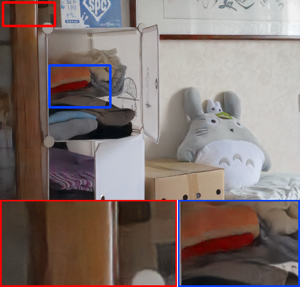} &
    \includegraphics[width=0.14\linewidth]{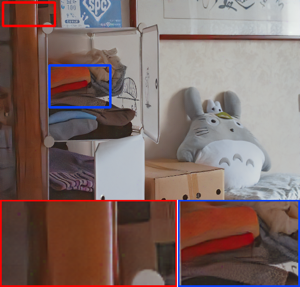} &
    \includegraphics[width=0.14\linewidth]{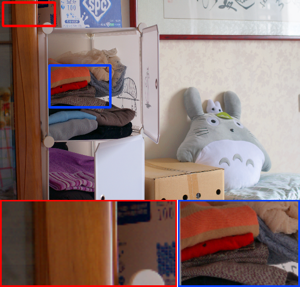} \\

    \includegraphics[width=0.14\linewidth]{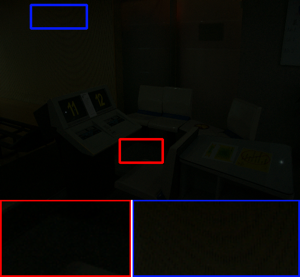} &
    \includegraphics[width=0.14\linewidth]{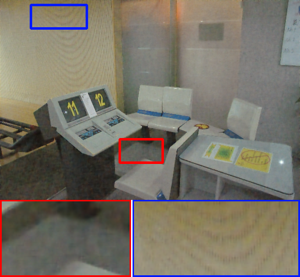} &
    \includegraphics[width=0.14\linewidth]{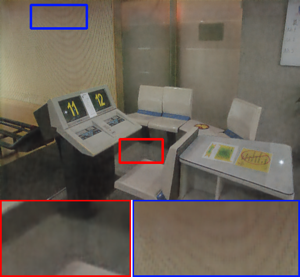} &
    \includegraphics[width=0.14\linewidth]{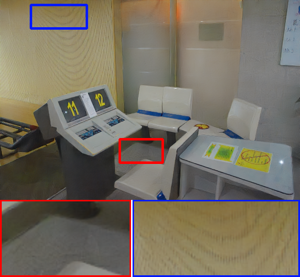} &
    \includegraphics[width=0.14\linewidth]{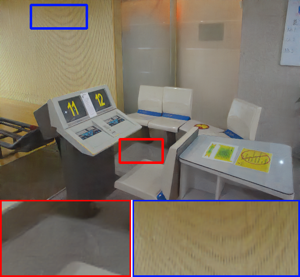} &
    \includegraphics[width=0.14\linewidth]{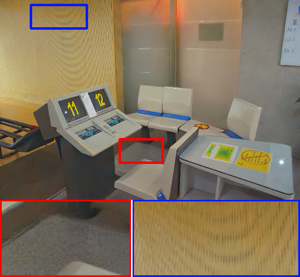} &
    \includegraphics[width=0.14\linewidth]{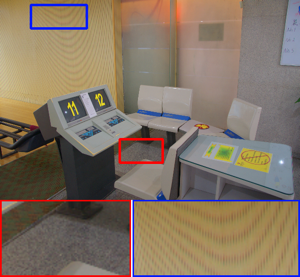} \\

    \includegraphics[width=0.14\linewidth]{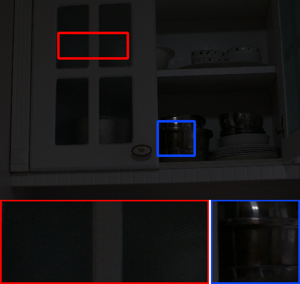} &
    \includegraphics[width=0.14\linewidth]{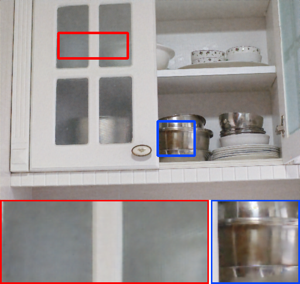} &
    \includegraphics[width=0.14\linewidth]{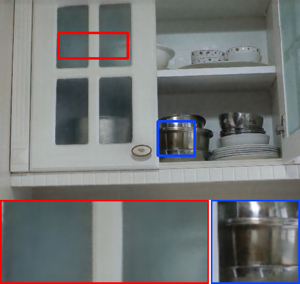} &
    \includegraphics[width=0.14\linewidth]{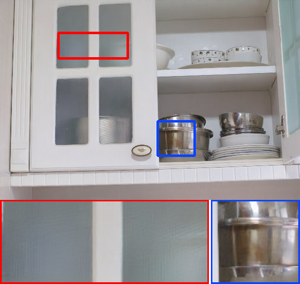} &
    \includegraphics[width=0.14\linewidth]{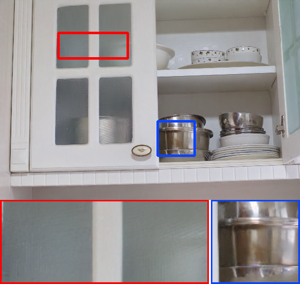} &
    \includegraphics[width=0.14\linewidth]{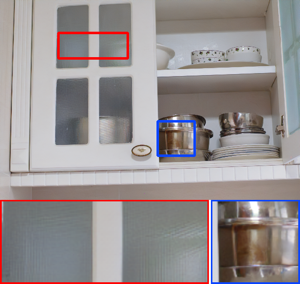} &
    \includegraphics[width=0.14\linewidth]{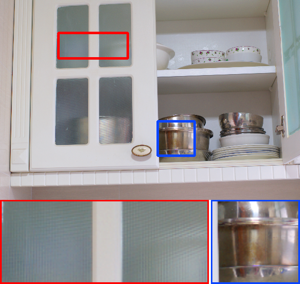} \\

    \\[-1pt]
    {\scriptsize{Input}} &
    {\scriptsize{Retformer}\cite{cai2023retinexformer}} &
    {\scriptsize{LDiff}\cite{jiang2024lightendiffusion}} &
    {\scriptsize{CIDNet}\cite{yan2025hvi}} &
    {\scriptsize{ICLR}\cite{xu2025iclr}} &
    {\scriptsize{Ours}} &
    {\scriptsize{GT}} \\
  \end{tabular}

  \caption{More visual comparisons of the enhanced results by different methods on LOLv1~\cite{wei2018deep}. Zoom in for the best view.}
  \label{fig:more_lolv1_visual}
\end{figure*}

\begin{figure*}[tb]
  \centering

  \setlength{\tabcolsep}{0.5pt}
  \renewcommand{\arraystretch}{0.3}

  \begin{tabular}{ccccccc}

    \includegraphics[width=0.14\linewidth]{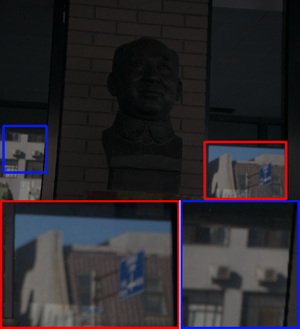} &
    \includegraphics[width=0.14\linewidth]{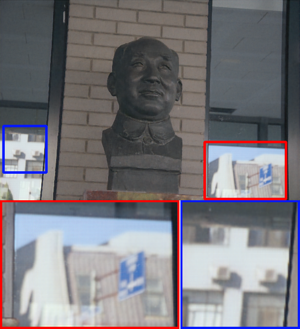} &
    \includegraphics[width=0.14\linewidth]{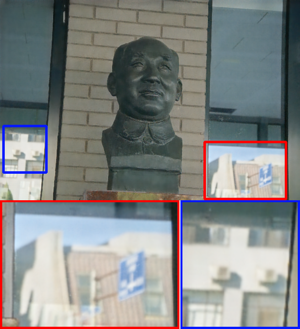} &
    \includegraphics[width=0.14\linewidth]{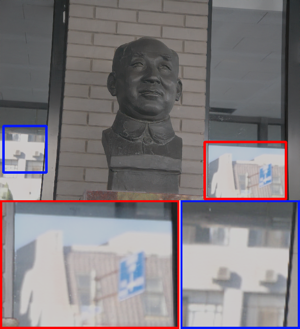} &
    \includegraphics[width=0.14\linewidth]{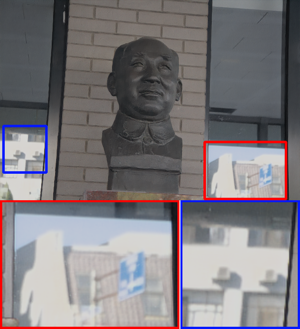} &
    \includegraphics[width=0.14\linewidth]{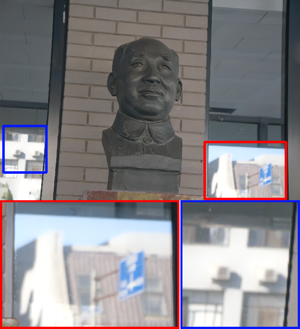} &
    \includegraphics[width=0.14\linewidth]{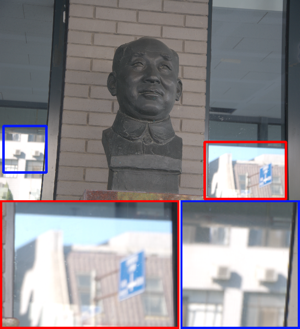} \\

    \includegraphics[width=0.14\linewidth]{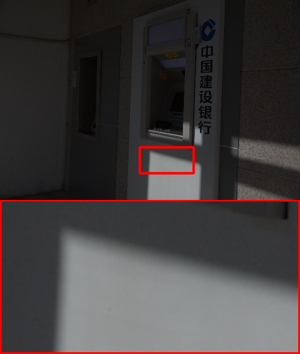} &
    \includegraphics[width=0.14\linewidth]{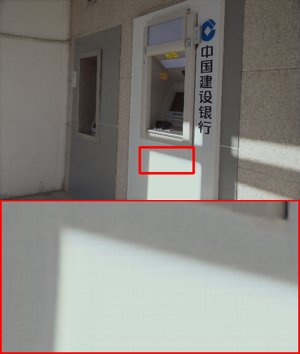} &
    \includegraphics[width=0.14\linewidth]{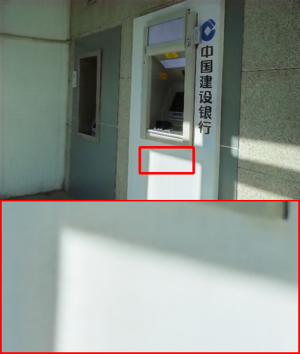} &
    \includegraphics[width=0.14\linewidth]{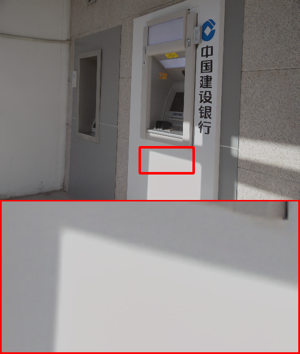} &
    \includegraphics[width=0.14\linewidth]{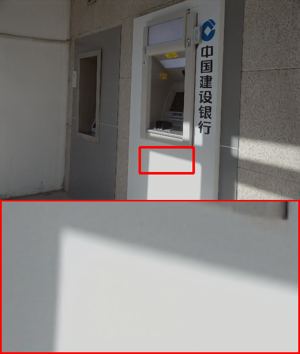} &
    \includegraphics[width=0.14\linewidth]{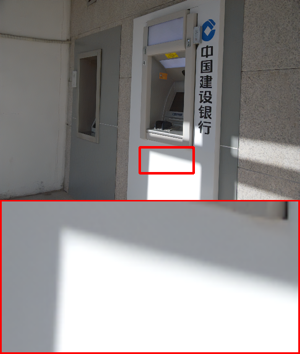} &
    \includegraphics[width=0.14\linewidth]{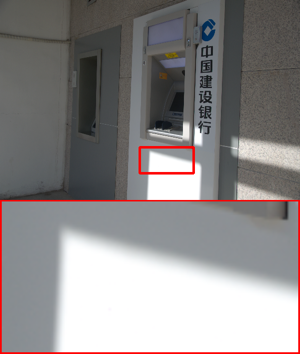} \\

    \includegraphics[width=0.14\linewidth]{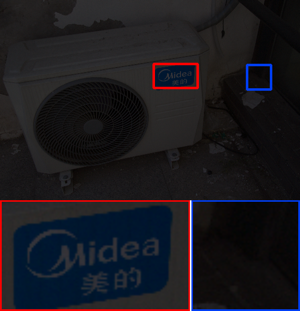} &
    \includegraphics[width=0.14\linewidth]{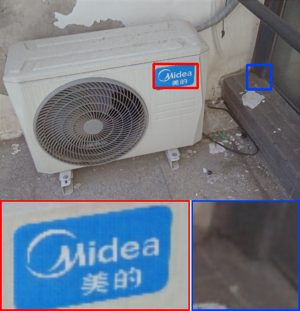} &
    \includegraphics[width=0.14\linewidth]{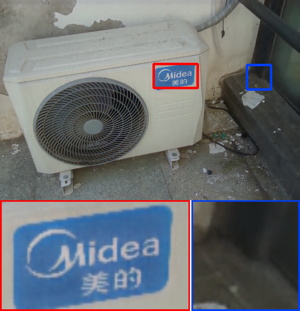} &
    \includegraphics[width=0.14\linewidth]{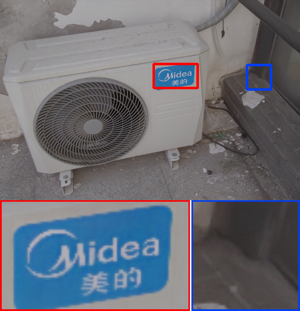} &
    \includegraphics[width=0.14\linewidth]{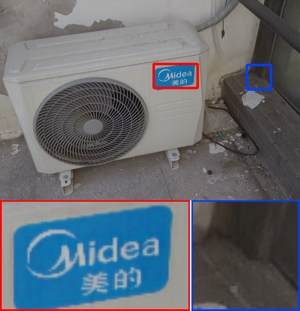} &
    \includegraphics[width=0.14\linewidth]{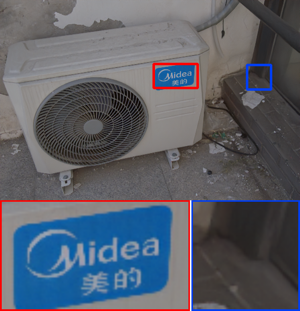} &
    \includegraphics[width=0.14\linewidth]{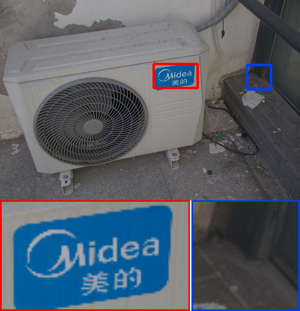} \\

    \includegraphics[width=0.14\linewidth]{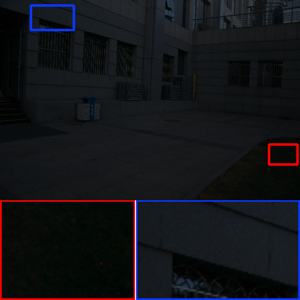} &
    \includegraphics[width=0.14\linewidth]{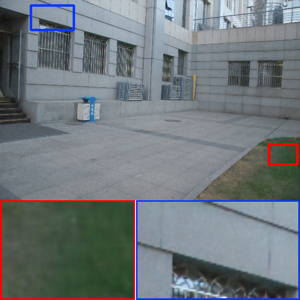} &
    \includegraphics[width=0.14\linewidth]{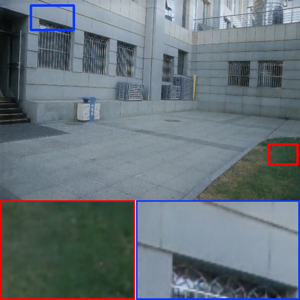} &
    \includegraphics[width=0.14\linewidth]{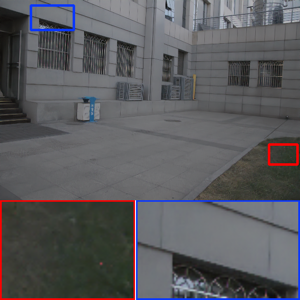} &
    \includegraphics[width=0.14\linewidth]{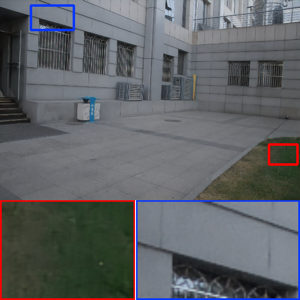} &
    \includegraphics[width=0.14\linewidth]{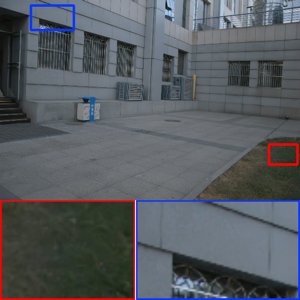} &
    \includegraphics[width=0.14\linewidth]{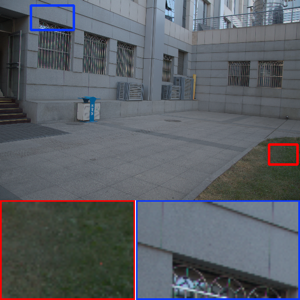} \\

    \\[-1pt]
    {\scriptsize{Input}} &
    {\scriptsize{Retformer}\cite{cai2023retinexformer}} &
    {\scriptsize{LDiff}\cite{jiang2024lightendiffusion}} &
    {\scriptsize{CIDNet}\cite{yan2025hvi}} &
    {\scriptsize{ICLR}\cite{xu2025iclr}} &
    {\scriptsize{Ours}} &
    {\scriptsize{GT}} \\
  \end{tabular}

  \caption{More visual comparisons of the enhanced results by different methods on LOLv2-Real~\cite{yang2021sparse}. Zoom in for the best view.}
  \label{fig:more_lolv2_real_visual}
\end{figure*}

\begin{figure*}[tb]
  \centering

  \setlength{\tabcolsep}{0.5pt}
  \renewcommand{\arraystretch}{0.3}

  \begin{tabular}{ccccccc}

    \includegraphics[width=0.14\linewidth]{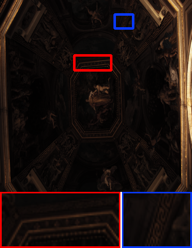} &
    \includegraphics[width=0.14\linewidth]{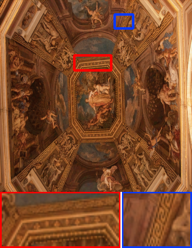} &
    \includegraphics[width=0.14\linewidth]{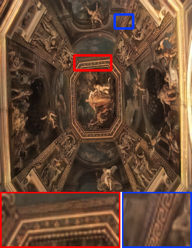} &
    \includegraphics[width=0.14\linewidth]{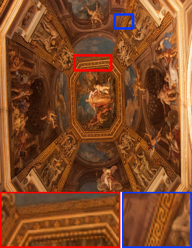} &
    \includegraphics[width=0.14\linewidth]{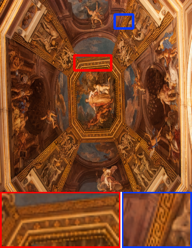} &
    \includegraphics[width=0.14\linewidth]{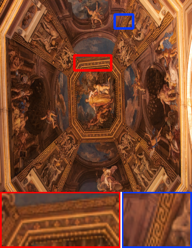} &
    \includegraphics[width=0.14\linewidth]{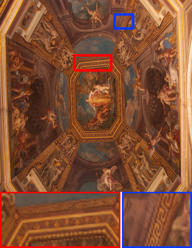} \\

    \includegraphics[width=0.14\linewidth]{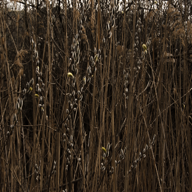} &
    \includegraphics[width=0.14\linewidth]{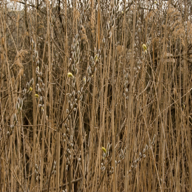} &
    \includegraphics[width=0.14\linewidth]{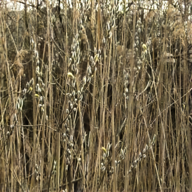} &
    \includegraphics[width=0.14\linewidth]{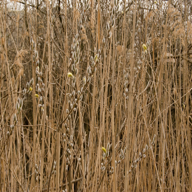} &
    \includegraphics[width=0.14\linewidth]{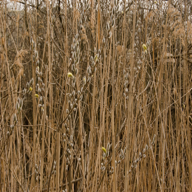} &
    \includegraphics[width=0.14\linewidth]{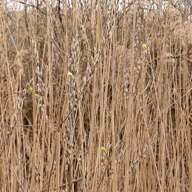} &
    \includegraphics[width=0.14\linewidth]{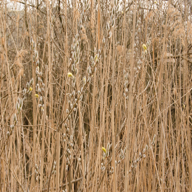} \\

    \includegraphics[width=0.14\linewidth]{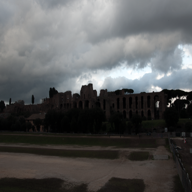} &
    \includegraphics[width=0.14\linewidth]{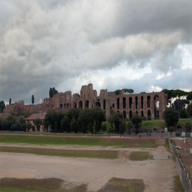} &
    \includegraphics[width=0.14\linewidth]{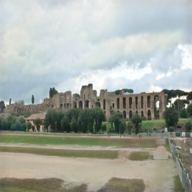} &
    \includegraphics[width=0.14\linewidth]{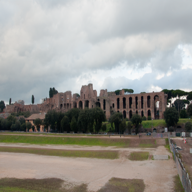} &
    \includegraphics[width=0.14\linewidth]{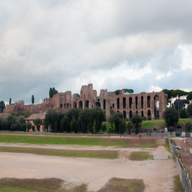} &
    \includegraphics[width=0.14\linewidth]{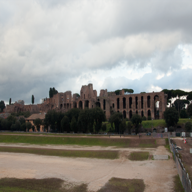} &
    \includegraphics[width=0.14\linewidth]{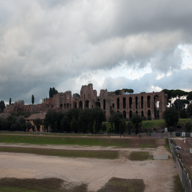} \\

    \\[-1pt]
    {\scriptsize{Input}} &
    {\scriptsize{Retformer}\cite{cai2023retinexformer}} &
    {\scriptsize{LDiff}\cite{jiang2024lightendiffusion}} &
    {\scriptsize{CIDNet}\cite{yan2025hvi}} &
    {\scriptsize{ICLR}\cite{xu2025iclr}} &
    {\scriptsize{Ours}} &
    {\scriptsize{GT}} \\
  \end{tabular}

  \caption{More visual comparisons of the enhanced results by different methods on LOLv2-Synthetic~\cite{yang2021sparse}. Zoom in for the best view.}
  \label{fig:more_lolv2_syn_visual}
\end{figure*}

\begin{figure*}[tb]
  \centering

  \setlength{\tabcolsep}{0.5pt}
  \renewcommand{\arraystretch}{0.3}

  \begin{tabular}{ccccccc}

    \includegraphics[width=0.14\linewidth]{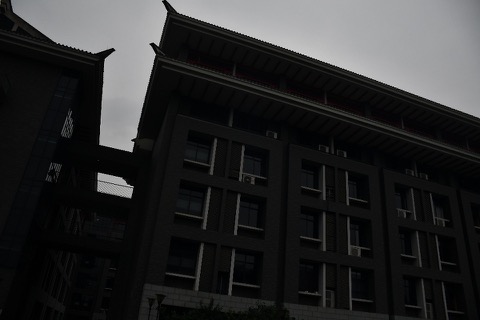} &
    \includegraphics[width=0.14\linewidth]{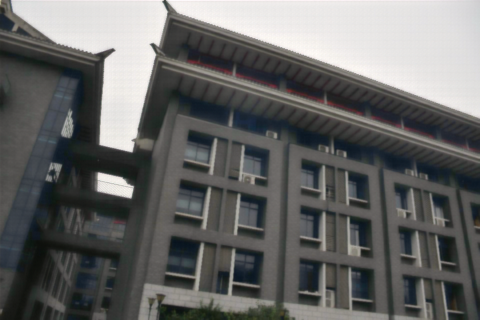} &
    \includegraphics[width=0.14\linewidth]{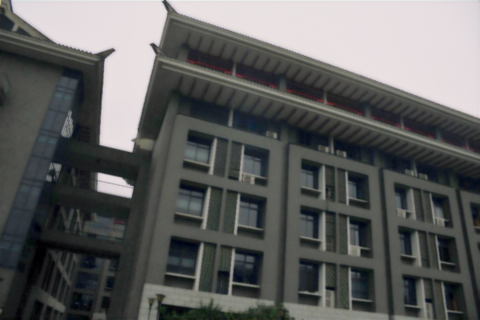} &
    \includegraphics[width=0.14\linewidth]{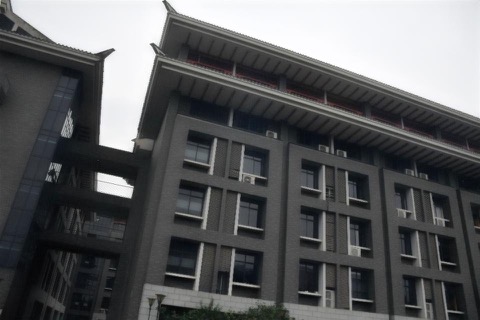} &
    \includegraphics[width=0.14\linewidth]{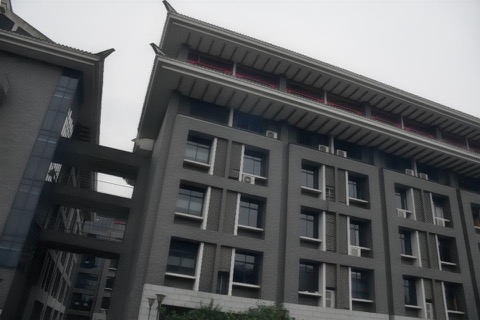} &
    \includegraphics[width=0.14\linewidth]{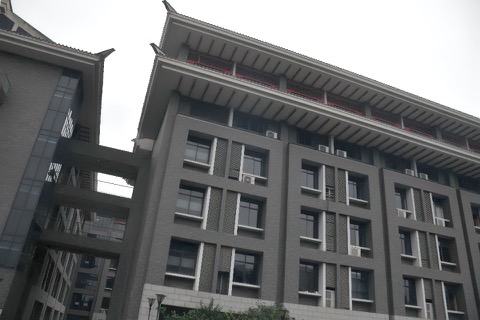} &
    \includegraphics[width=0.14\linewidth]{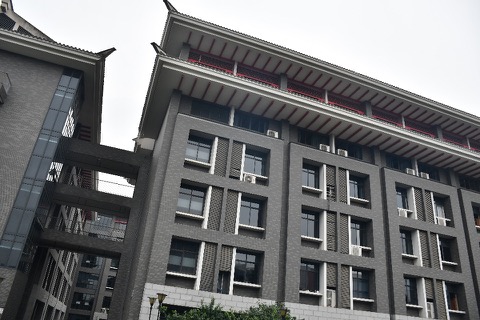} \\

    \includegraphics[width=0.14\linewidth]{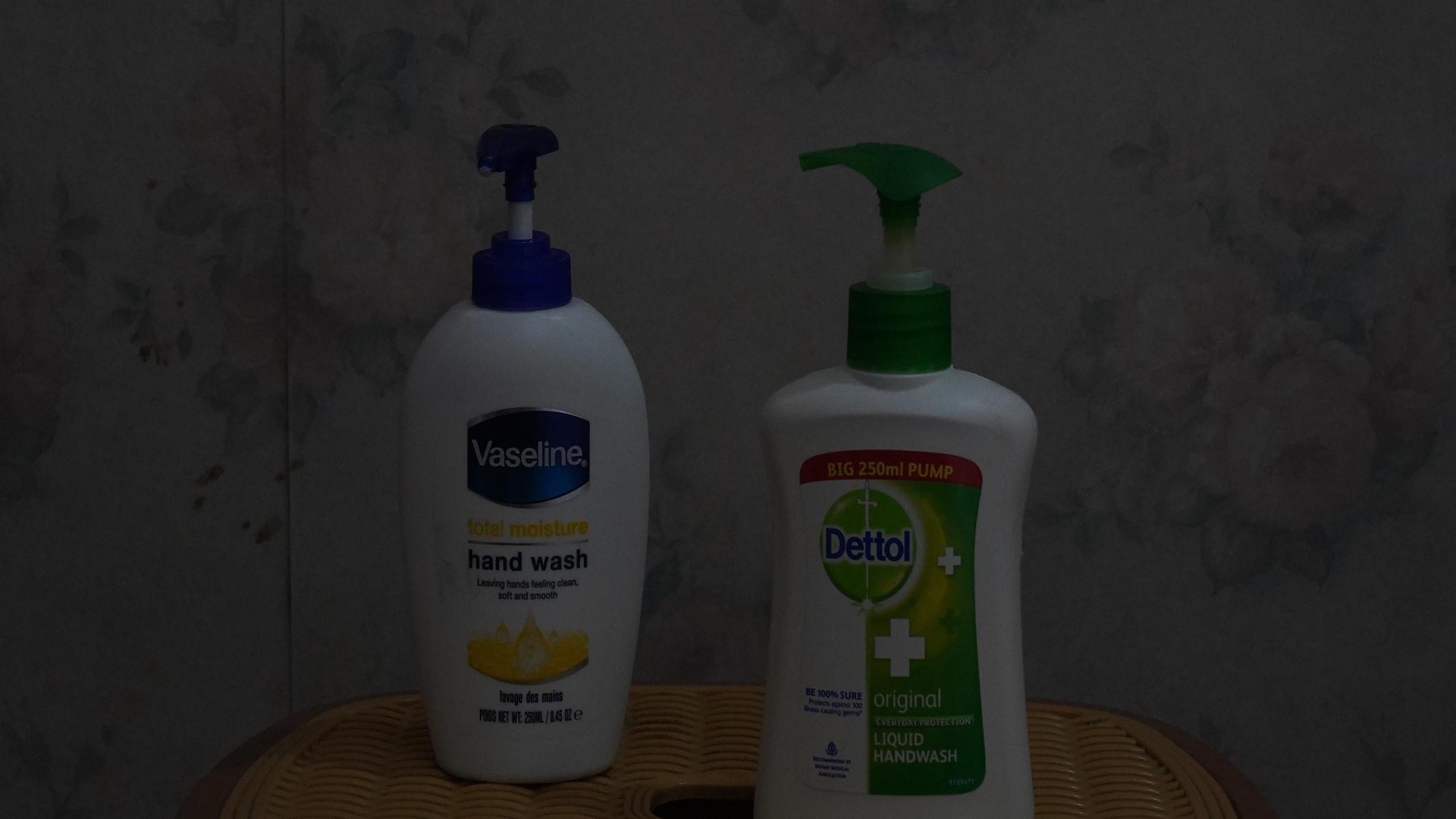} &
    \includegraphics[width=0.14\linewidth]{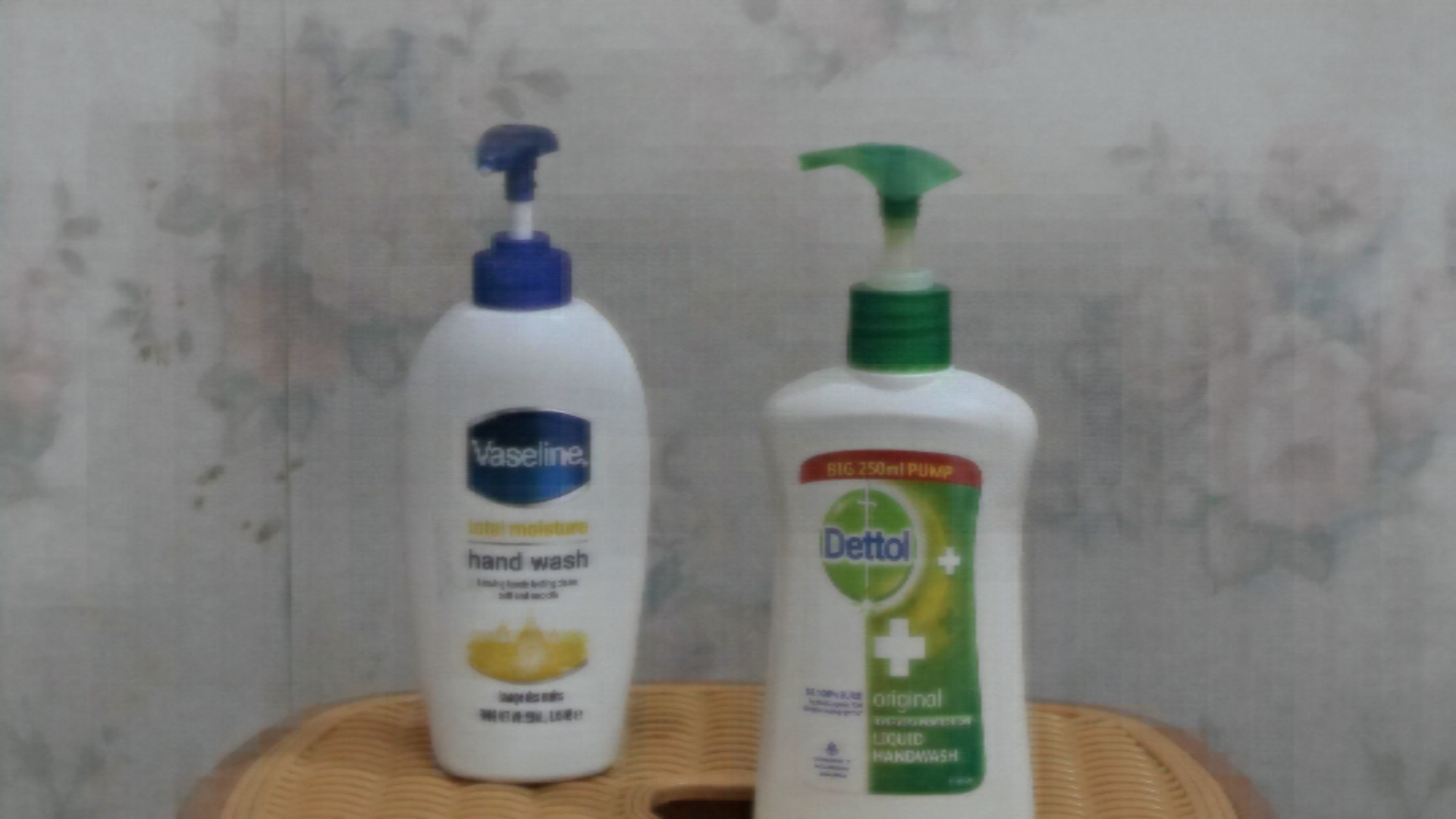} &
    \includegraphics[width=0.14\linewidth]{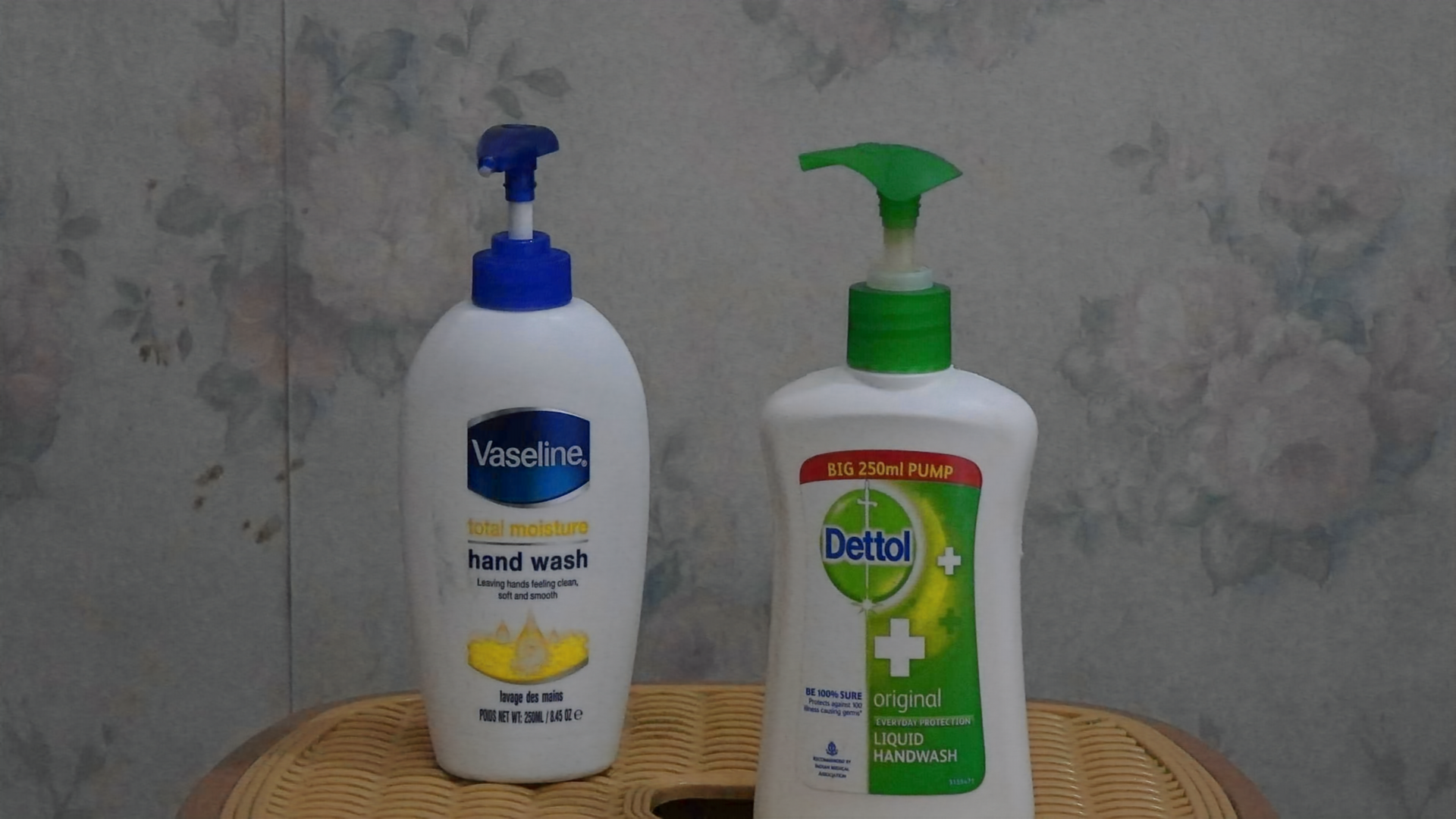} &
    \includegraphics[width=0.14\linewidth]{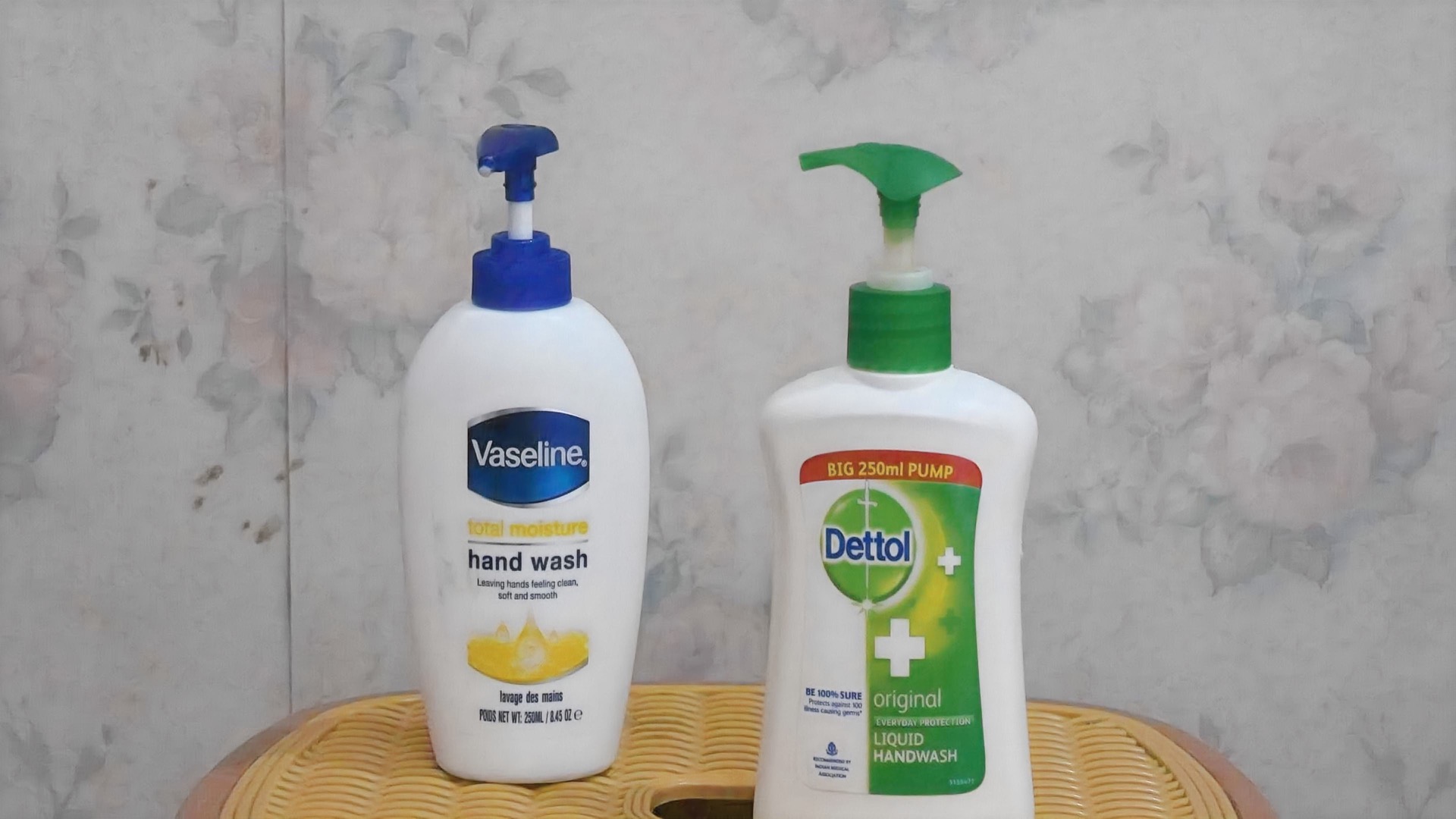} &
    \includegraphics[width=0.14\linewidth]{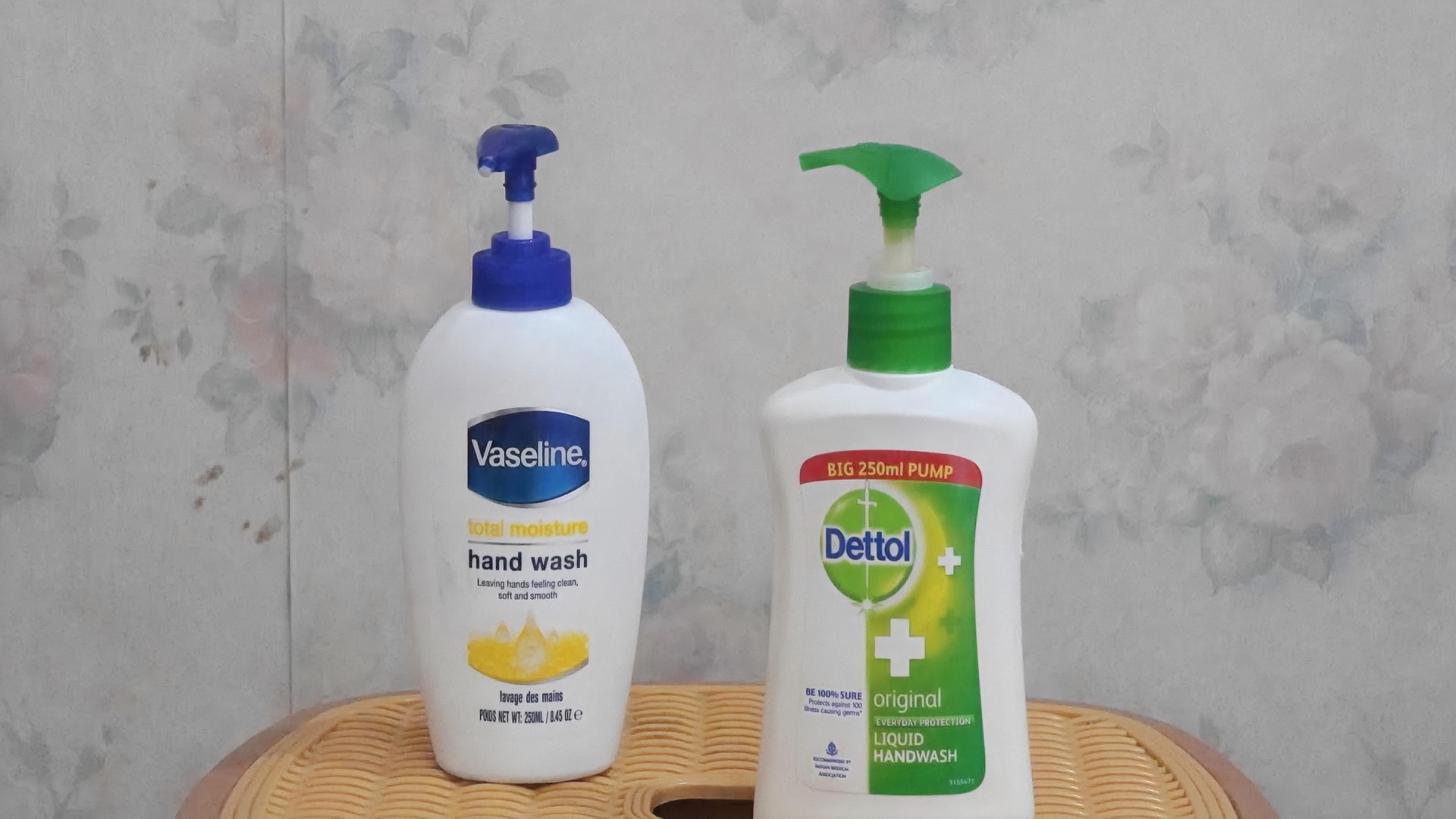} &
    \includegraphics[width=0.14\linewidth]{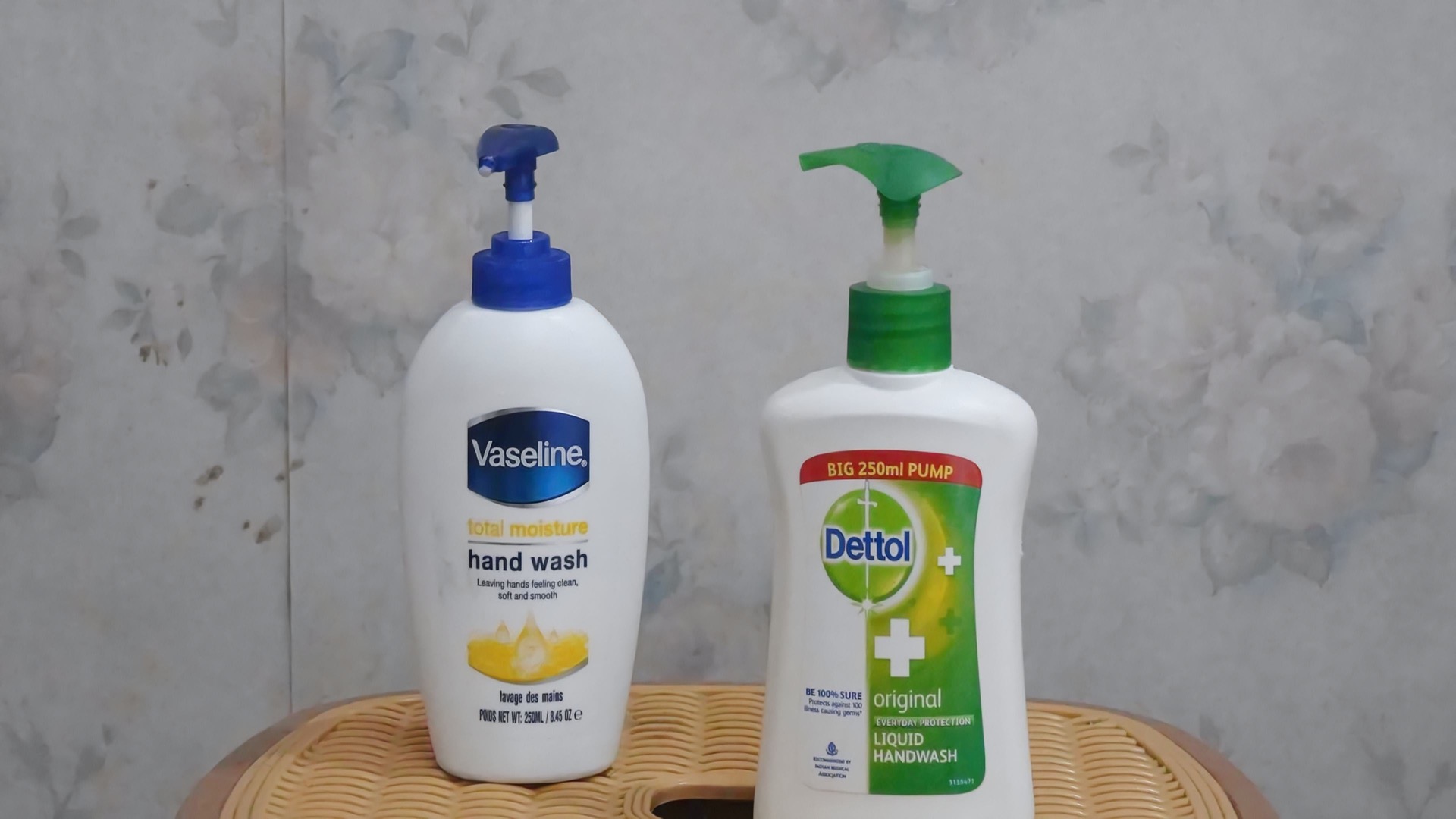} &
    \includegraphics[width=0.14\linewidth]{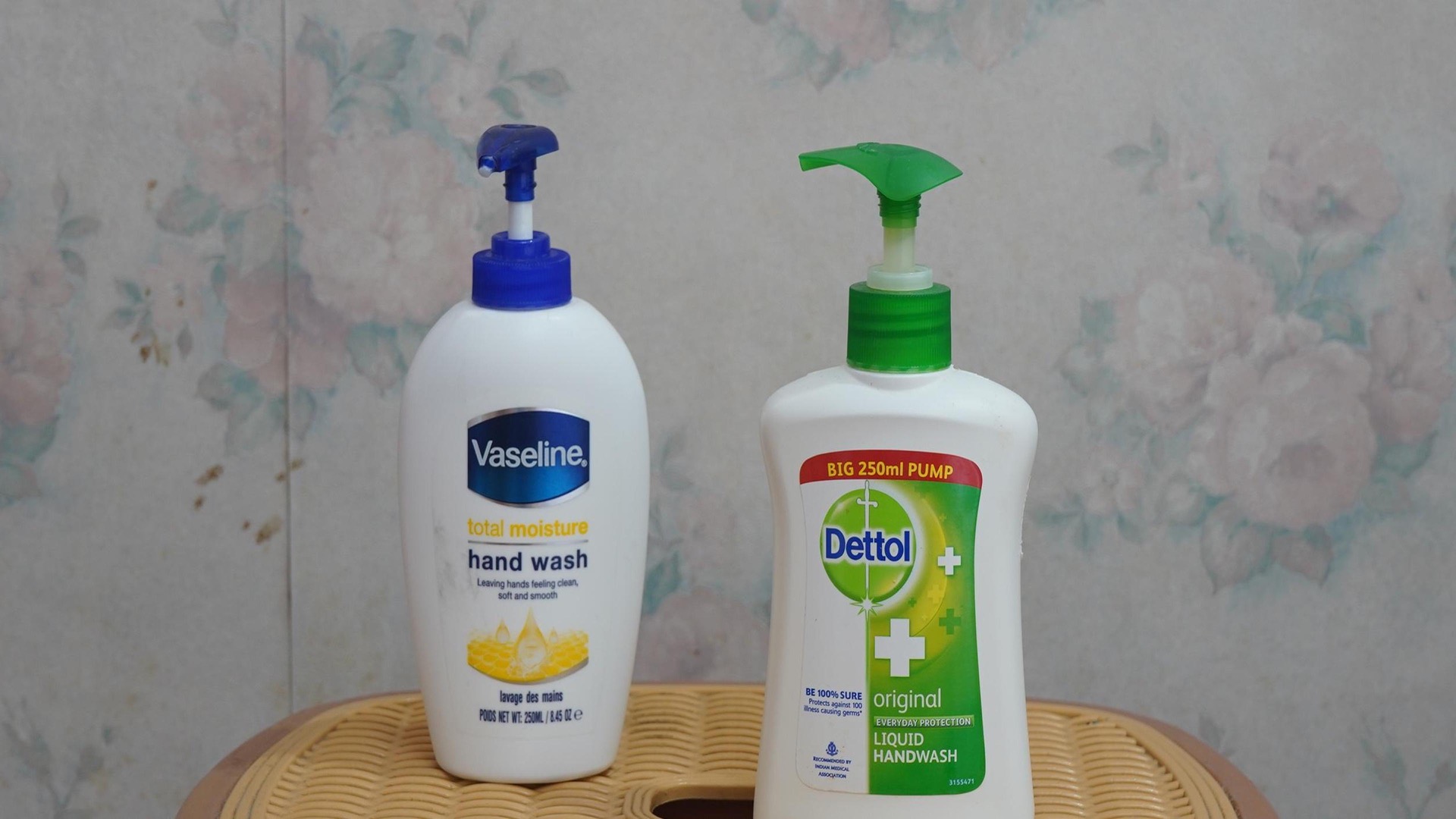} \\

    \includegraphics[width=0.14\linewidth]{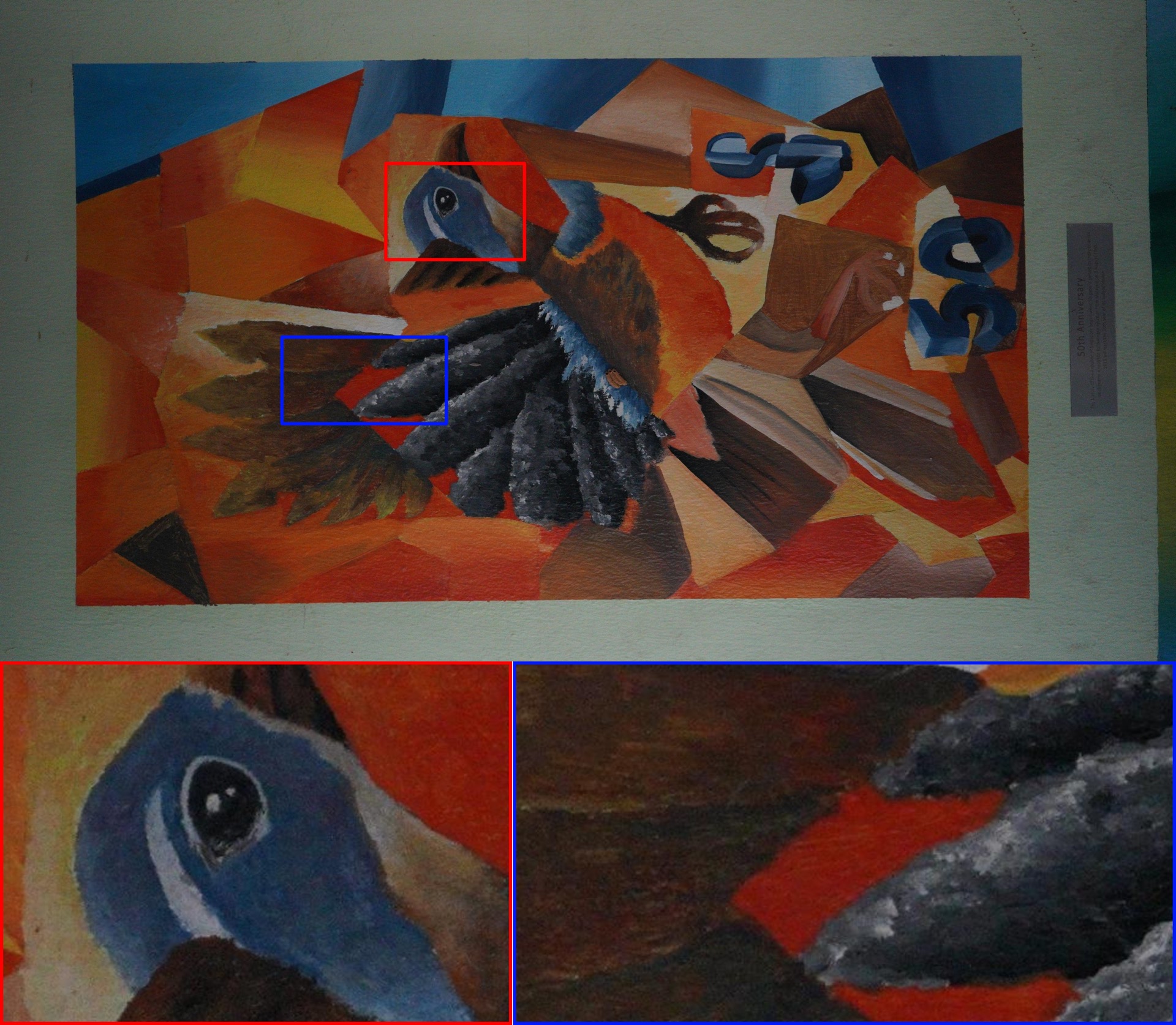} &
    \includegraphics[width=0.14\linewidth]{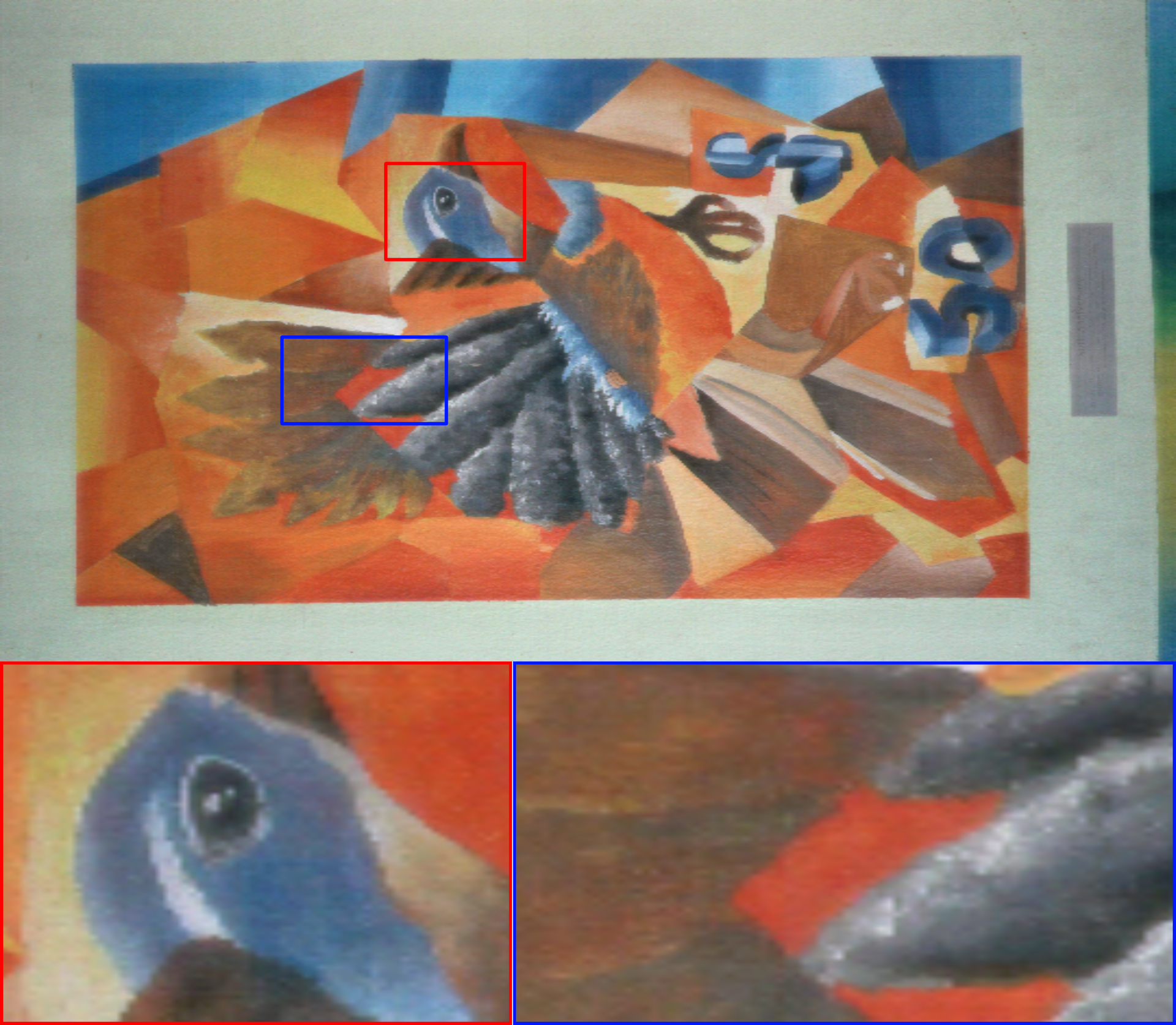} &
    \includegraphics[width=0.14\linewidth]{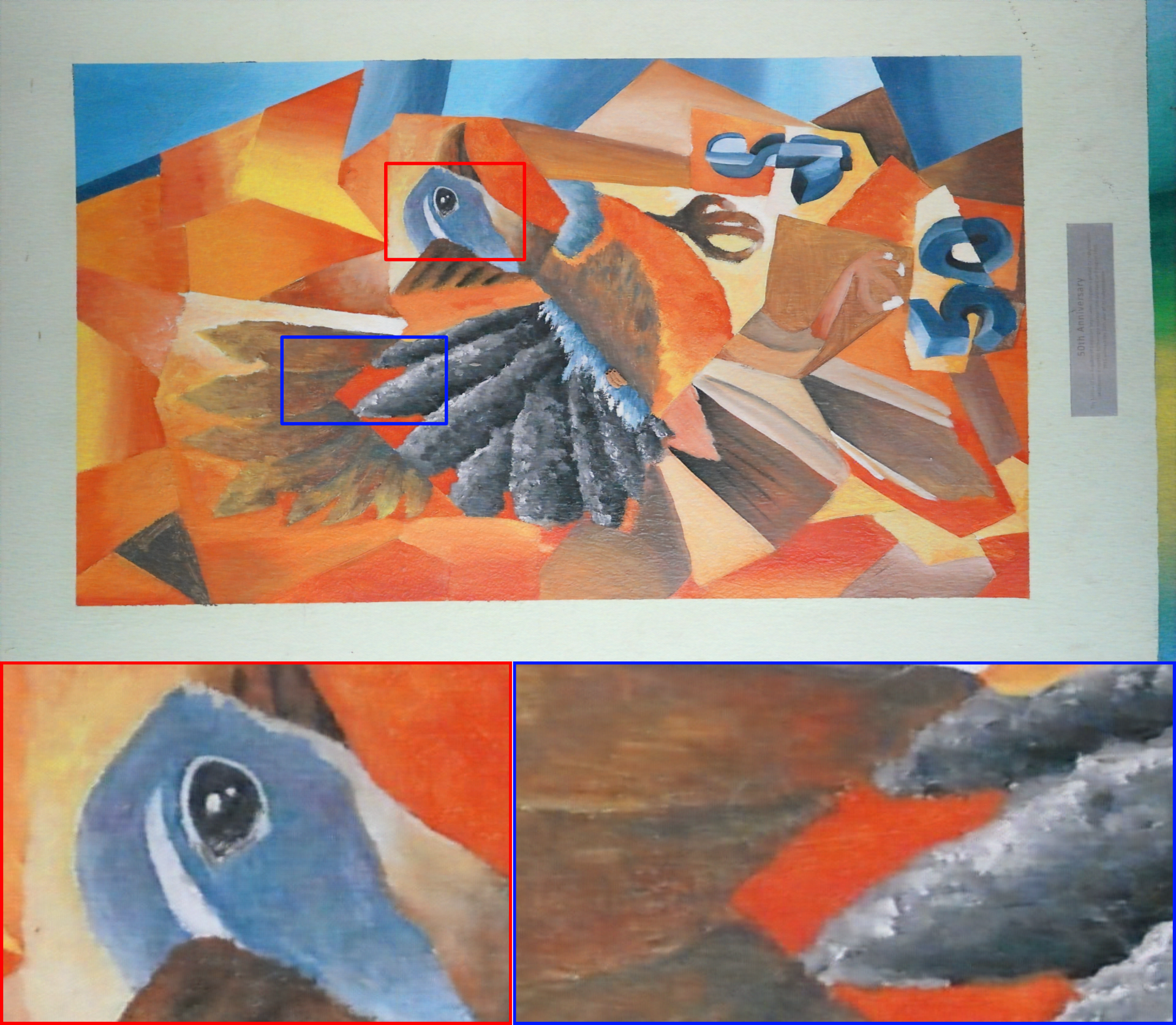} &
    \includegraphics[width=0.14\linewidth]{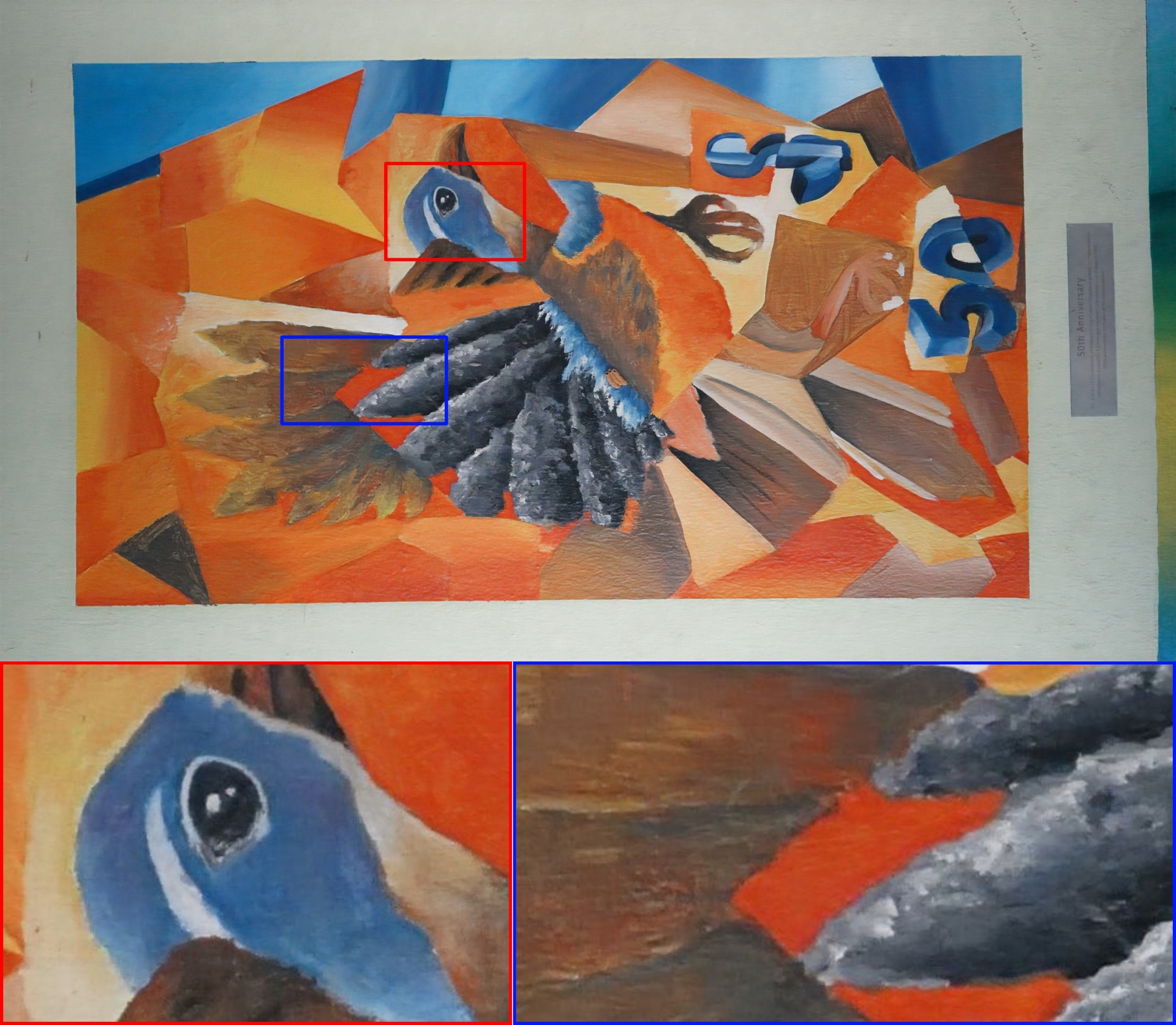} &
    \includegraphics[width=0.14\linewidth]{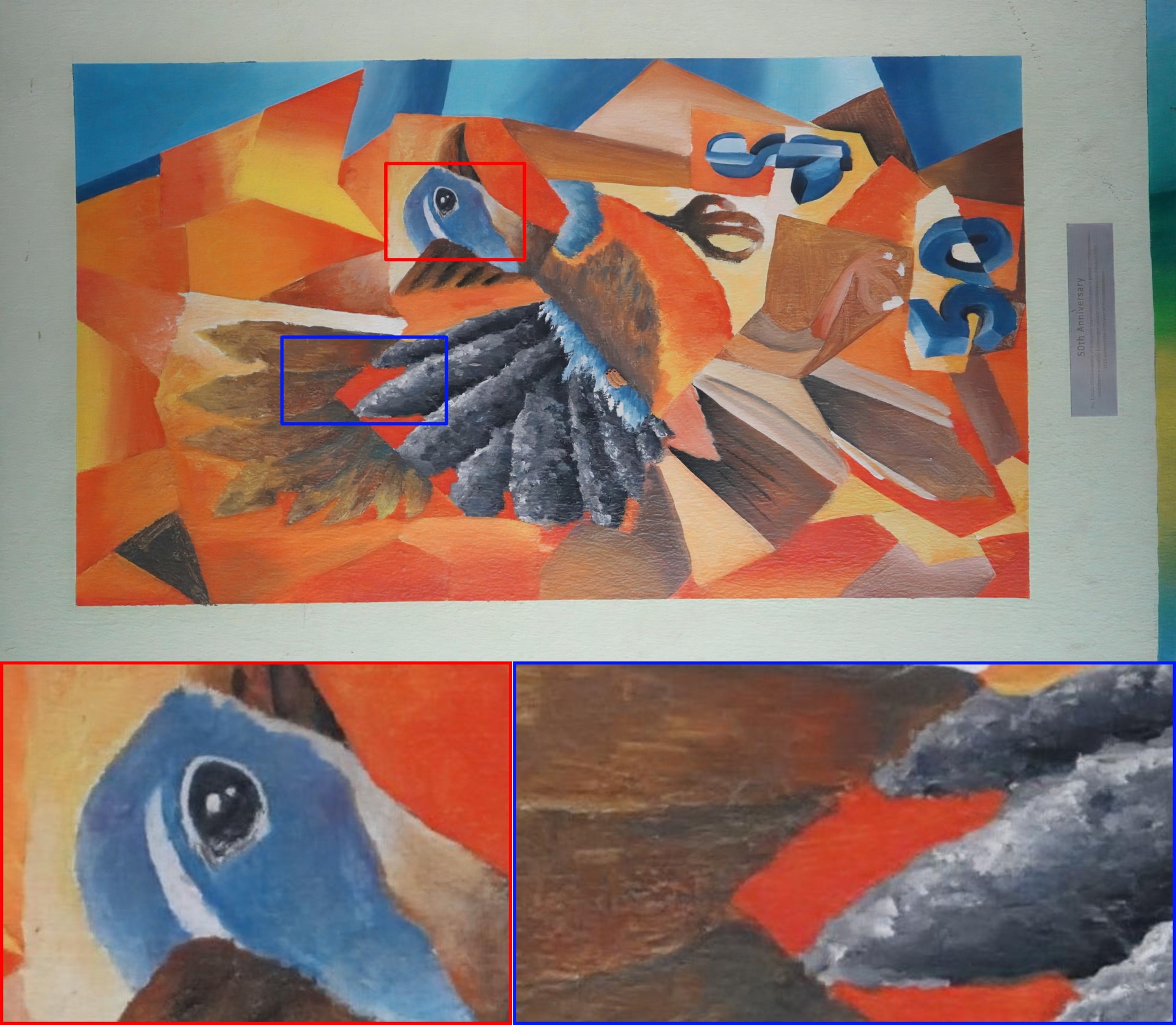} &
    \includegraphics[width=0.14\linewidth]{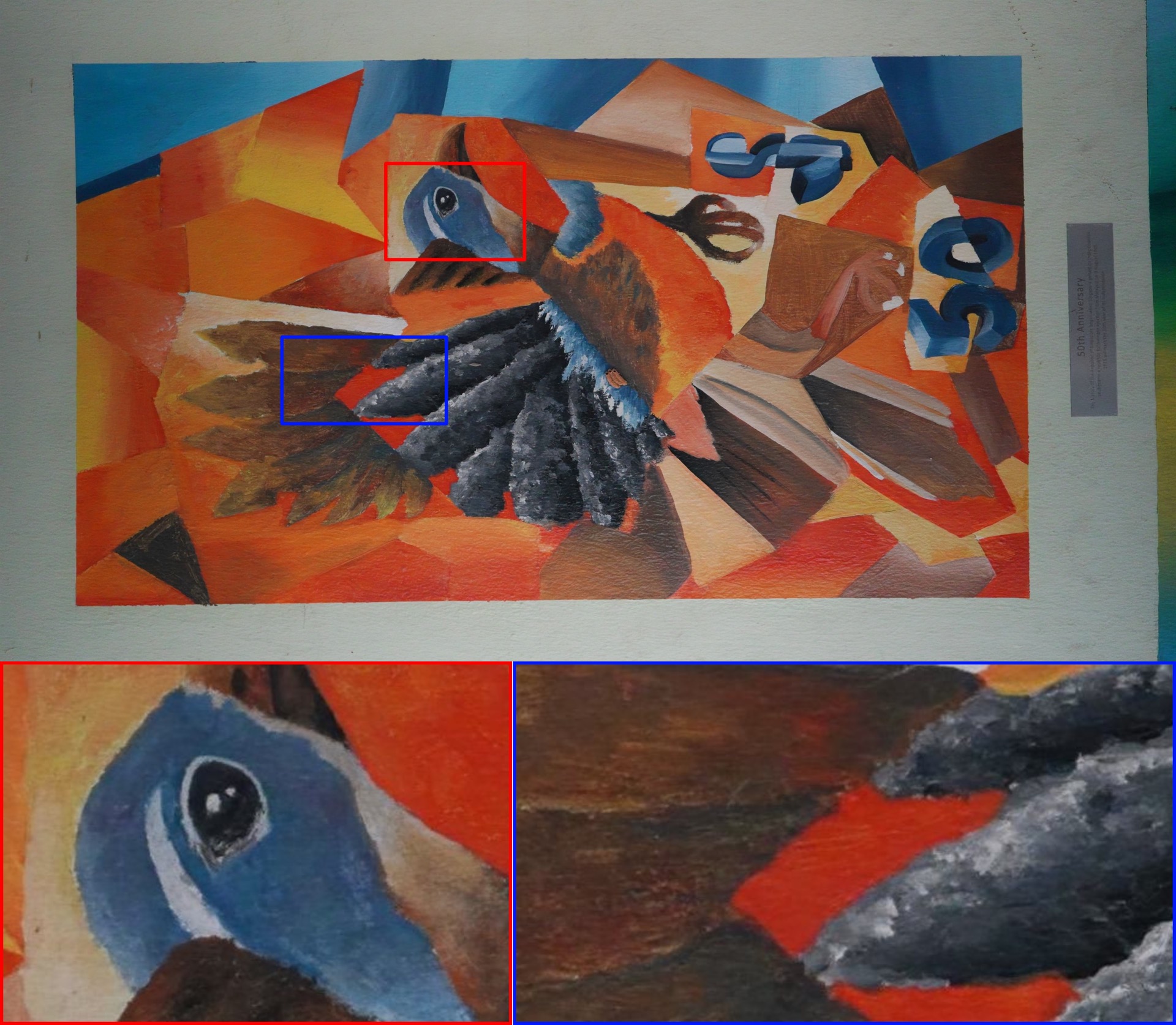} &
    \includegraphics[width=0.14\linewidth]{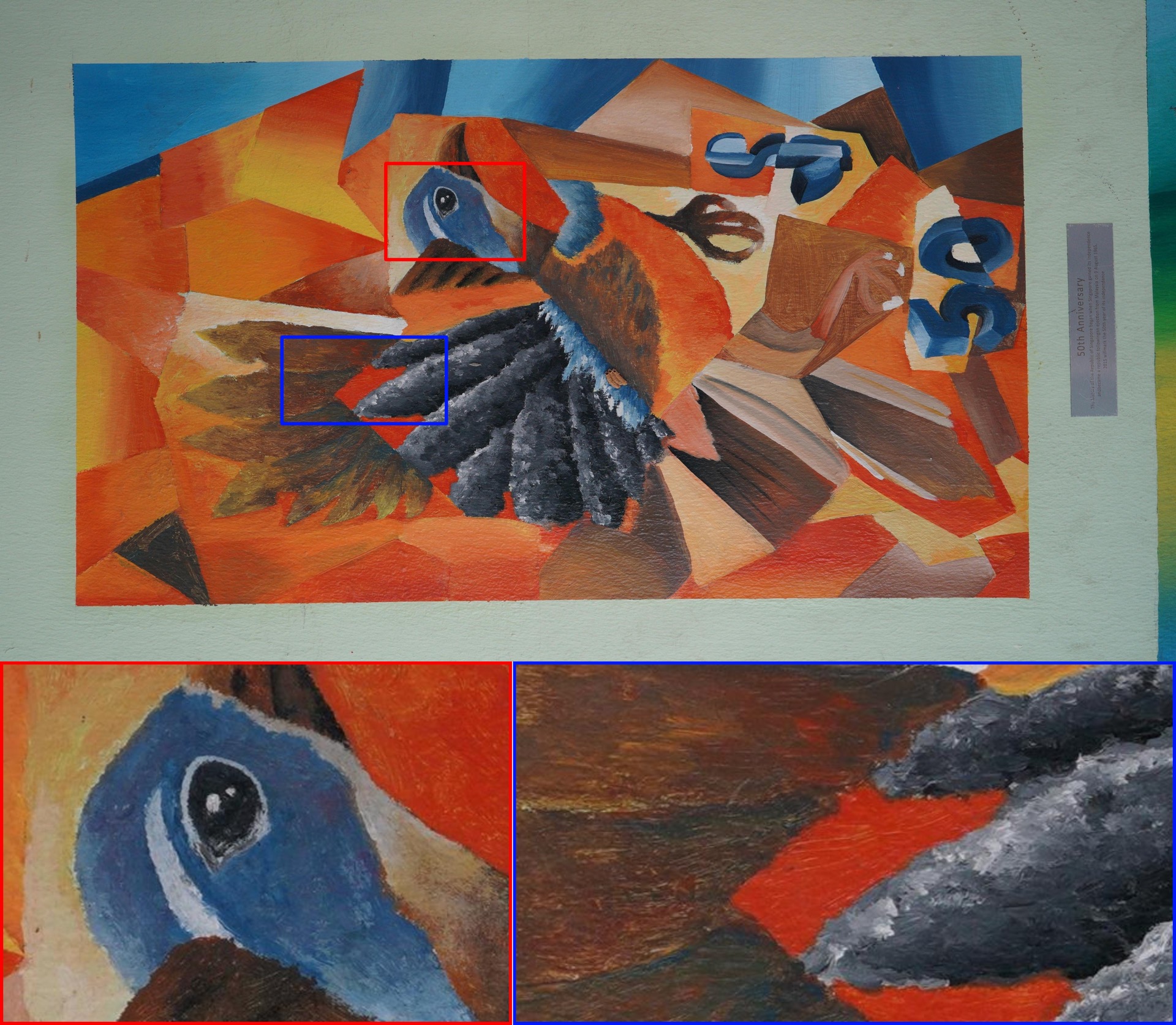} \\

    \\[-1pt]
    {\scriptsize{Input}} &
    {\scriptsize{UFormer}\cite{wang2022uformer}} &
    {\scriptsize{Retformer}\cite{cai2023retinexformer}} &
    {\scriptsize{CIDNet}\cite{yan2025hvi}} &
    {\scriptsize{ICLR}\cite{xu2025iclr}} &
    {\scriptsize{Ours}} &
    {\scriptsize{GT}} \\
  \end{tabular}

  \caption{More visual comparisons of the enhanced results by different methods on LSRW-Nikon~\cite{hai2023r2rnet} and UHD-LL~\cite{li2023embedding}. Zoom in for the best view.}
  \label{fig:more_lsrw_uhdll_visual}
\end{figure*}

\clearpage
\section{Limitations}
\begin{figure*}[tb]
  \centering

  \setlength{\tabcolsep}{0.5pt}
  \renewcommand{\arraystretch}{0.3}

  \begin{tabular}{ccccccc}

    \includegraphics[width=0.14\linewidth]{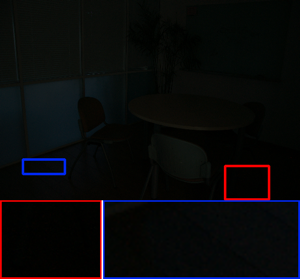} &
    \includegraphics[width=0.14\linewidth]{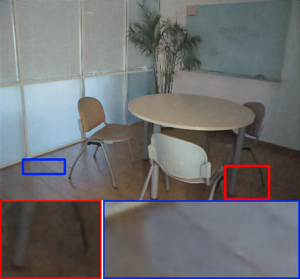} &
    \includegraphics[width=0.14\linewidth]{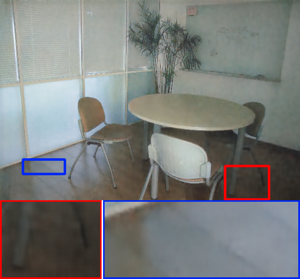} &
    \includegraphics[width=0.14\linewidth]{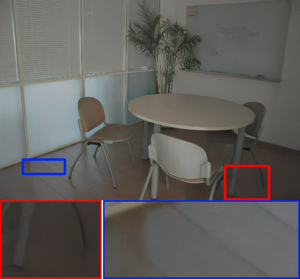} &
    \includegraphics[width=0.14\linewidth]{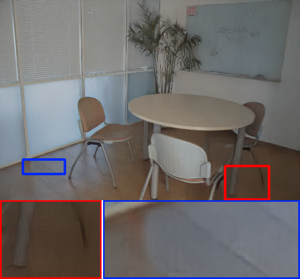} &
    \includegraphics[width=0.14\linewidth]{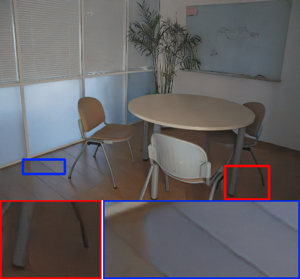} &
    \includegraphics[width=0.14\linewidth]{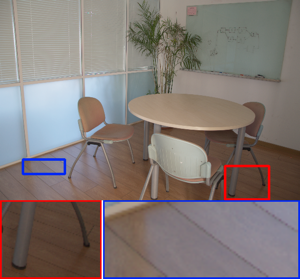} \\

    \\[-1pt]
    {\scriptsize{Input}} &
    {\scriptsize{Retformer}\cite{cai2023retinexformer}} &
    {\scriptsize{LDiff}\cite{jiang2024lightendiffusion}} &
    {\scriptsize{CIDNet}\cite{yan2025hvi}} &
    {\scriptsize{ICLR}\cite{xu2025iclr}} &
    {\scriptsize{Ours}} &
    {\scriptsize{GT}} \\
  \end{tabular}

\caption{Qualitative comparison in challenging real-world low-light scenarios. In scenes characterized by significant spatial variations in illumination and extreme contrast, ATP and existing methods both encounter difficulties in perfect color restoration. However, ATP demonstrates superior robustness in preserving structural details and suppressing artifacts.}
\label{fig:limitation}
\end{figure*}

Although our proposed ATP method achieves leading performance across various resolutions and lighting conditions, we acknowledge its limitations. As illustrated in~\cref{fig:limitation}, in challenging real world scenarios with low illumination, ATP struggles to achieve perfect lighting and color restoration, similar to other existing approaches. Such scenarios become particularly challenging if these specific illumination distribution patterns (characterized by strongly spatially varying lighting and the simultaneous presence of bright and extremely dark regions) are underrepresented or entirely absent in the training data distribution. Nevertheless, by leveraging its robust global energy estimation and decoupled restoration capabilities, ATP still obtains comparatively better results in these extreme cases. It effectively suppresses artifacts arising from insufficient illumination enhancement and preserves fine structural details. This highlights the ongoing challenges in achieving perfect generalization to all unseen severe degradation conditions caused by low illumination, while simultaneously emphasizing the significant advantages of introducing explicit global energy alignment and decoupled restoration.

\end{document}